\documentclass[twoside,british,5p]{elsarticle}
\usepackage[T1]{fontenc}
\usepackage{textcomp}
\usepackage[utf8]{inputenc}
\usepackage{xcolor}
\usepackage{prettyref}
\usepackage{booktabs}
\usepackage{calc}
\usepackage{mathrsfs}
\usepackage{amsmath}
\usepackage{amssymb}
\usepackage{graphicx}
\usepackage{xargs}[2008/03/08]

\makeatletter

\newcommand*\LyXZeroWidthSpace{\hspace{0pt}}
\providecommand{\tabularnewline}{\\}

\journal{Robotics and Autonomous Systems}

\usepackage{lipsum}

\makeatother

\usepackage{babel}
\begin{document}
\begin{frontmatter}
\title{A receding-horizon multi-contact motion planner for legged robots
in challenging environments}
\author[UoM]{D.S.J.~Derwent}
\ead{daniel.johnson-2@manchester.ac.uk}
\author[UoM]{S.~Watson}
\ead{simon.watson@manchester.ac.uk}
\author[UoM]{B.V.~Adorno\corref{cor1}}
\ead{bruno.adorno@manchester.ac.uk}
\cortext[cor1]{Corresponding author}
\address[UoM]{Manchester Centre for Robotics and AI, University of Manchester, Oxford
Rd, Manchester M13 9PL, UK}
\address{}
\begin{abstract}
We present a novel receding-horizon multi-contact motion planner for
legged robots in challenging scenarios, able to plan motions such
as chimney climbing, navigating very narrow passages or crossing large
gaps. Our approach adds new capabilities to the state of the art,
including the ability to reactively re-plan in response to new information,
and planning contact locations and whole-body trajectories simultaneously,
simplifying the implementation and removing the need for post-processing
or complex multi-stage approaches. Our method is more resistant to
local minima problems than other potential field based approaches,
and our quadratic-program-based posture generator returns nodes more
quickly than those of existing algorithms. Rigorous statistical analysis
shows that, with short planning horizons (\emph{e.g.}, one step ahead),
our planner is faster than the state-of-the-art across all scenarios
tested (between 45\% and 98\% faster on average, depending on the
scenario), while planning less efficient motions (requiring 5\% fewer
to 700\% more stance changes on average). In all but one scenario
(Chimney Walking), longer planning horizons (\emph{e.g}., four steps
ahead) extended the average planning times (between 73\% faster and
400\% slower than the state-of-the-art) but resulted in higher quality
motion plans (between 8\% more and 47\% fewer stance changes than
the state-of-the-art).
\end{abstract}
\end{frontmatter}

\selectlanguage{english}%

\global\long\def\imi{\hat{\imath}}%

\global\long\def\imj{\hat{\jmath}}%

\global\long\def\imk{\hat{k}}%

\global\long\def\dual{\varepsilon}%

\global\long\def\dq#1{\underline{\boldsymbol{#1}}}%

\global\long\def\quat#1{\boldsymbol{#1}}%

\global\long\def\mydual#1{\underline{#1}}%

\global\long\def\spin{\text{Spin}(3)}%

\global\long\def\spinr{\text{Spin}(3){\ltimes}\mathbb{R}^{3}}%

\global\long\def\mymatrix#1{\boldsymbol{#1}}%

\global\long\def\myvec#1{\boldsymbol{#1}}%

\global\long\def\hamidq#1#2{\overset{#1}{\operatorname{\mymatrix H}}_{8}\left(#2\right)}%

\global\long\def\hamiquat#1#2{\overset{#1}{\operatorname{\mymatrix H}}_{4}\left(#2\right)}%

\global\long\def\getp#1{\operatorname{\mathcal{P}}\left(#1\right)}%

\global\long\def\getd#1{\operatorname{\mathcal{D}}\left(#1\right)}%

\global\long\def\real#1{\operatorname{\mathrm{Re}}\left(#1\right)}%

\global\long\def\imag#1{\operatorname{\mathrm{Im}}\left(#1\right)}%

\global\long\def\vector{\operatorname{vec}}%

\global\long\def\veceight#1{\operatorname{vec}_{8}\left(#1\right)}%

\global\long\def\vecfour#1{\operatorname{vec}_{4}\left(#1\right)}%

\global\long\def\vecsix#1{\operatorname{vec}_{6}\left(#1\right)}%

\global\long\def\vecthree#1{\operatorname{vec}_{3}\left(#1\right)}%

\global\long\def\swap#1{\text{swap}\{#1\}}%

\global\long\def\tplus{\dq{{\cal T}}}%

\global\long\def\ad#1#2{\text{Ad}\left(#1\right)#2}%

\global\long\def\adsharp#1#2{\text{Ad}_{\sharp}\left(#1\right)#2}%

\global\long\def\adflat#1#2{\text{Ad}_{\flat}\left(#1\right)#2}%

\global\long\def\dotproduct#1{\langle#1\rangle}%

\global\long\def\norm#1{\left\Vert #1\right\Vert }%

\global\long\def\diag#1{\operatorname{diag}\left(#1\right)}%

\selectlanguage{british}%

\selectlanguage{english}%
\global\long\def\argminimone#1#2#3#4{\begin{aligned}#1\:  &  \underset{#2}{\arg\!\min}  &   &  #3\\
  &  \text{subject to}  &   &  #4 
\end{aligned}
 }%

\global\long\def\minimtwo#1#2#3#4{ \begin{aligned} &  \underset{#1}{\min}  &   &  #2 \\
  &  \text{subject to}  &   &  #3 \\
  &   &   &  #4 
\end{aligned}
 }%

\global\long\def\minimone#1#2#3{ \begin{aligned} &  \underset{#1}{\min}  &   &  #2 \\
  &  \text{subject to}  &   &  #3 
\end{aligned}
 }%

\global\long\def\argminimtwo#1#2#3#4#5{ \begin{aligned}#1\:  &  \underset{#2}{\arg\!\min}  &   &  #3 \\
  &  \text{subject to}  &   &  #4\\
  &   &   &  #5 
\end{aligned}
 }%

\global\long\def\argmaximtwo#1#2#3#4#5{ \begin{aligned}#1\:  &  \underset{#2}{\arg\!\max}  &   &  #3 \\
  &  \text{subject to}  &   &  #4 \\
  &   &   &  #5 
\end{aligned}
 }%
\selectlanguage{british}%

\selectlanguage{english}%

\global\long\def\frame#1{\mathscr{F}_{#1}}%

\global\long\def\config{\myvec q}%

\global\long\def\transition{\myvec q_{t}}%

\global\long\def\point{\quat p}%

\global\long\def\rot{\quat r}%

\global\long\def\line{\dq l}%

\global\long\def\direction{\quat l}%

\global\long\def\moment{\quat m}%

\global\long\def\plane{\dq{\pi}}%

\global\long\def\unitnormal{\quat n}%

\global\long\def\contact{c}%

\global\long\def\patch{\mathcal{P}}%

\global\long\def\stance{s}%

\global\long\def\pose{\dq h}%

\global\long\def\jointangles{\myvec{\theta}}%

\global\long\def\jointangle{\theta}%

\global\long\def\identity{\mymatrix I}%

\global\long\def\node{\mathfrak{n}}%

\global\long\def\algorithmname{\text{{RHCP}}}%

\global\long\def\configguidepath{\mathfrak{q}}%

\global\long\def\guidepath{\configguidepath}%

\global\long\def\patchguidepath{\mathfrak{p}}%

\global\long\def\potfield{U}%

\global\long\def\rootnode{\node_{0}}%

\global\long\def\bestnode{\node_{\text{best}}}%

\global\long\def\nextnode{\node_{\text{next}}}%

\global\long\def\previousroot{\node_{\text{prev}}}%

\global\long\def\mindist{d_{\min}}%

\global\long\def\potfieldweighting{\alpha}%

\global\long\def\force{\myvec f}%

\global\long\def\COM{\quat p_{C}}%

\global\long\def\veczero{\myvec 0}%

\global\long\def\vecslack{\myvec s}%

\global\long\def\closest{\min}%

\global\long\def\selected{\diamond}%

\global\long\def\mean{\mu}%

\global\long\def\scale{\rho}%

\global\long\def\stdev{\sigma}%

\global\long\def\normality{\nu}%

\global\long\def\dist{X}%

\global\long\def\given{\mid}%

\global\long\def\data{\mathcal{D}}%

\global\long\def\effsize{\delta}%

\global\long\def\regweighting{\lambda}%

\global\long\def\slackweighting{\beta}%

\global\long\def\samplevec{\myvec a}%

\global\long\def\covariance{\mymatrix K}%

\global\long\def\axang#1{#1_{\rot}}%

\global\long\def\axpos#1{#1_{\point}}%

\global\long\def\metric{m}%

\global\long\def\samplemean{M}%

\global\long\def\samplestdev{S}%

\global\long\def\deg{\text{°}}%

\global\long\def\weighting{w}%

\global\long\def\interlim{\gamma}%

\global\long\def\indexrobot{i}%

\global\long\def\indexenv{j}%

\global\long\def\indexgeneration{\kappa}%

\global\long\def\horizondepth{\indexgeneration_{\max}}%

\global\long\def\kmax{\horizondepth}%

\global\long\def\isreal{\in\mathbb{R}}%

\global\long\def\isvector#1{\isreal{}^{#1}}%

\global\long\def\ismatrix#1#2{\isvector{#1\times#2}}%

\global\long\def\isnonnegative{\in[0,\infty)}%

\global\long\def\ispositive{\in(0,\infty)}%

\global\long\def\ispositiveint{\in\mathbb{N}}%

\global\long\def\isnonnegativeint{\in\mathbb{N}_{0}}%

\global\long\def\quatset{\mathbb{H}}%

\global\long\def\isquat{\in\quatset}%

\global\long\def\purequatset{\quatset_{p}}%

\global\long\def\ispurequat{\in\purequatset}%

\global\long\def\unitquatset{\mathbb{S}^{3}}%

\global\long\def\isunitquat{\in\unitquatset}%

\global\long\def\isunitpurequat{\ispurequat\cap\unitquatset}%

\global\long\def\dqset{\mathcal{H}}%

\global\long\def\isdq{\in\dqset}%

\global\long\def\ispuredq{\in\dqset_{p}}%

\global\long\def\unitdqset{\dq{\mathcal{S}}}%

\global\long\def\isunitdq{\in\unitdqset}%

\global\long\def\isline{\ispuredq\cap\unitdqset}%

\global\long\def\isplane{\isdq}%

\global\long\def\workspace{\mathcal{W}}%

\global\long\def\robotEXCEPTION{\mathcal{B}}%

\global\long\def\body{\mathcal{B}}%

\global\long\def\obstacle{\mathcal{O}}%

\global\long\def\area{\mathcal{A}}%

\global\long\def\allareas{\mathcal{E}}%

\global\long\def\allcontactpatches{\mathcal{R}}%

\global\long\def\stancespace{\mathcal{S}}%

\global\long\def\jointspace{\Theta}%

\global\long\def\configspace{\mathcal{Q}}%

\global\long\def\baseconfigspace{\mathcal{Q_{\text{base}}}}%

\global\long\def\feasiblespace{\mathcal{F}}%

\global\long\def\freespace{\workspace_{\text{free}}}%

\global\long\def\foliation{\mathcal{M}}%

\global\long\def\nodesequence{\mathcal{N}}%

\global\long\def\generation{\mathcal{G}}%

\global\long\def\cache{\mathcal{C}}%

\global\long\def\min{\text{min}}%

\global\long\def\max{\text{max}}%

\global\long\def\transpose{\intercal}%

\global\long\def\initial{I}%

\global\long\def\goal{G}%

\global\long\def\areaplane#1{\plane_{\area_{#1}}}%

\global\long\def\areanormal#1{\unitnormal_{\area_{#1}}}%

\global\long\def\areaboundmatrix#1{\mymatrix A_{\area_{#1}}}%

\global\long\def\areaboundvector#1{\myvec b_{\area_{#1}}}%

\global\long\def\nearpatches{\allcontactpatches_{\text{near}}}%

\global\long\def\midpatches{\allcontactpatches_{\text{mid}}}%

\global\long\def\farpatches{\allcontactpatches_{\text{far}}}%

\global\long\def\fieldpatchweighting{\beta_{\indexrobot}}%

\global\long\def\fieldpatchoffset{b}%

\global\long\def\horizon{\generation_{\horizondepth}}%

\global\long\def\escapenodes{\mathcal{G}_{\text{esc}}}%

\global\long\def\bodycollisionsurfaces#1{\Pi_{\mathrm{\obstacle}\mathcal{B}_{#1}}}%

\global\long\def\allcollisionsurfaces{\Pi_{\mathrm{\obstacle}}}%

\global\long\def\collisionbodies{\mathcal{H}_{\robotEXCEPTION}}%

\global\long\def\selfcollisionbodies{\mathcal{Y}_{\mathcal{\robotEXCEPTION}}}%

\global\long\def\contactpoint#1{\point_{#1}^{\area_{#1}}}%

\global\long\def\contposeerror{\tilde{\myvec x}}%

\global\long\def\setpoint{\config_{s}}%

\global\long\def\torsopos{\point_{T}}%

\global\long\def\torsorot{\rot_{T}}%

\global\long\def\contfooterror{\tilde{\point_{m}}}%

\global\long\def\movfootpos{\point_{m}}%

\global\long\def\contvariables{\myvec x}%

\global\long\def\conterror{\tilde{\contvariables}}%

\global\long\def\contscaler{\eta_{\contvariables}}%

\global\long\def\continput{\myvec u}%

\global\long\def\jac{\mymatrix J}%

\global\long\def\normforce{f_{\unitnormal_{\indexrobot}}}%

\global\long\def\desnormforce{f_{\unitnormal_{i}d}}%

\global\long\def\normforceerror{\tilde{\normforce}}%

\global\long\def\normforcegain{\eta_{f}}%

\global\long\def\pgain{\eta_{f_{P}}}%

\global\long\def\igain{\eta_{f_{I}}}%

\global\long\def\radius{r}%

\global\long\def\stiffness{k}%

\global\long\def\qvmin{\myvec v_{\text{\ensuremath{\min}}}}%

\global\long\def\qvmax{\myvec v_{\max}}%

\global\long\def\numbodies{n_{\body}}%

\global\long\def\numjoints{n_{\mathcal{J}}}%

\global\long\def\numtheta{n_{\mathcal{\theta}}}%

\global\long\def\numareabounds{n_{\text{bounds}}}%

\global\long\def\numcontacts{n_{\contact}}%

\global\long\def\numdof{n_{\text{DOF}}}%

\global\long\def\numfreefeet{n_{\text{free}}}%

\global\long\def\numpatches{n_{\allcontactpatches}}%

\global\long\def\numareas{n_{\allareas}}%

\global\long\def\numwayps{n_{W}}%

\global\long\def\numstancecontacts{n_{\stance}}%

\global\long\def\numtasks{n_{\tilde{\myvec x}}}%

\global\long\def\numobstacles{n_{\obstacle}}%

\global\long\def\numslackvariables{\ell}%

\global\long\def\crossproduct#1#2{#1\times#2}%

\newcommandx\monodistance[1][usedefault, addprefix=\global, 1=]{d_{#1}}%

\newcommandx\signdistance[3][usedefault, addprefix=\global, 1=]{d_{#2,#3}^{#1}}%

\newcommandx\signederror[3][usedefault, addprefix=\global, 1=]{\tilde{d}_{#2,#3}^{#1}}%

\global\long\def\squaredistance#1#2{D_{#1,#2}}%

\global\long\def\squaredistanceROC#1#2{\dot{D}_{#1,#2}}%

\newcommandx\monosquaredistance[1][usedefault, addprefix=\global, 1=]{D_{#1}}%

\global\long\def\safedistance#1{d_{\text{safe},#1}}%

\global\long\def\vecn#1#2{\operatorname{vec}_{#2}\left(#1\right)}%

\global\long\def\invecn#1#2{\operatorname{\underline{vec}}_{#2}\left(#1\right)}%

\global\long\def\invecthree#1{\invecn{#1}3}%

\global\long\def\invecfour#1{\invecn{#1}4}%

\global\long\def\invecsix#1{\invecn{#1}6}%

\global\long\def\inveceight#1{\invecn{#1}8}%

\global\long\def\innerproduct#1#2{\dotproduct{#1,#2}}%

\global\long\def\potential#1{\potfield\left(#1\right)}%

\global\long\def\trajectory#1#2{\mathcal{T}_{#1,#2}}%

\global\long\def\fullnode#1#2{\left(\node_{#1},\stance_{#2},\config_{#2},\trajectory{#1}{#2}\right)}%

\global\long\def\fullcontact#1#2{\left(\patch_{#1},\area_{#2},\contactpoint{#2}\right)}%

\global\long\def\revnode#1{\node_{#1}^{\text{rev}}}%

\global\long\def\fkm#1#2{\quat p_{#1}\left(#2\right)}%

\global\long\def\tdistsub#1#2{t_{#2}\left(\mean_{#1},\scale_{#1},\normality_{#1}\right)}%

\newcommandx\tdist[1][usedefault, addprefix=\global, 1=]{t\left(\mean_{#1},\scale_{#1},\normality_{#1}\right)}%

\global\long\def\prob#1{P\left(#1\right)}%

\global\long\def\argminimthree#1#2#3#4#5#6{ \begin{aligned}#1\:  &  \underset{#2}{\arg\!\min}  &   &  #3 \\
  &  \text{subject to}  &   &  #4\\
  &   &   &  #5 \\
  &   &   &  #6 
\end{aligned}
 }%

\global\long\def\argminimfour#1#2#3#4#5#6#7{ \begin{aligned}#1\:  &  \underset{#2}{\arg\!\min}  &   &  #3 \\
  &  \text{subject to}  &   &  #4\\
  &   &   &  #5 \\
  &   &   &  #6\\
  &   &   &  #7 
\end{aligned}
 }%

\global\long\def\argminimfive#1#2#3#4#5#6#7#8{ \begin{aligned}#1\:  &  \underset{#2}{\arg\!\min}  &   &  #3 \\
  &  \text{subject to}  &   &  #4\\
  &   &   &  #5 \\
  &   &   &  #6\\
  &   &   &  #7\\
  &   &   &  #8 
\end{aligned}
 }%

\global\long\def\lumppot#1{\monosquaredistance[\patchguidepath_{#1}]}%
\selectlanguage{british}%

\section{Introduction\label{sec:Introduction}}

A unique challenge of legged motion planning is the need to plan how
the robot will make and break contacts with its environment. If the
desired form of motion is known and approximately cyclical (\emph{e.g.},
walking across moderately uneven terrain), it may suffice to use a
pre-specified or adaptive gait. However, if robots are to perform
acyclic motions such as navigating sparse irregular footholds \citep{CBM-Bretl2006JournalPaper}
or rough terrain where the precise contact locations are critical
\citep{CBM-Hauser2008Primatives}, then they must explicitly plan
where, and in what sequence, individual contacts should be made or
broken. This is referred to as multi-contact motion planning.

Since fully enumerating the set of all possible contact combinations
(referred to as \emph{stances} \citep{CBM-Bretl2006JournalPaper})
is usually impractical, many multi-contact planners use a tree-search
method to construct an approximate map of the stance space, which
is then searched to generate the motion plan \citep{CBM-Bretl2006JournalPaper,CBM-Hauser2008ATHLETE,CVBFP-Escande2013MainPaper,CBM-Ferrari2023Multicontact}.
The design of the search process has significant ramifications for
the behaviour of the overall planning algorithm, and often involves
trade-offs between the generality of the planner, the time needed
to plan, and the quality of planning outputs \citep{MBC-TonneausGroup-Tonneau2018Efficient}.

Additionally, to ensure that a given stance is feasible, multi-contact
planners must find a safe whole-body configuration that allows the
robot to realise the intended stance while respecting certain constraints
(\emph{e.g.}, avoiding collisions, maintaining balance, \emph{etc.}).
The robot must also be able to safely transition into the intended
stance from a previous stance, meaning that a safe whole-body trajectory
linking the two stances must also be found. These lower-level elements
of planning are often handled by a dedicated `posture generator',
which finds \emph{transition configurations} through a numerical optimisation
process \citep{CVBFP-Escande2013MainPaper,CBM-Ferrari2023Multicontact}.

\subsection{Related Works}

Multi-contact planners are typically divided into Motion-Before-Contact
(MBC) and Contact-Before-Motion (CBM) approaches \citep{CBM-Bretl2006JournalPaper}.
MBC algorithms plan a collision-free torso trajectory and then use
a posture generator to find stances and transition configurations
that enable each resulting torso pose. For example, the planner of
Tonneau \emph{et al.} generates a large offline dataset of randomly
sampled limb configurations that is searched at runtime to assemble
configurations \citep{MBC-TonneausGroup-Tonneau2018Efficient}. MBC
planners typically execute faster than CBM planners, needing only
to consider the robot's torso pose rather than the full configuration.
However, MBC approaches must assume what kinds of motion the robot
can execute and constrain the trajectory search to regions where such
motions are most likely possible, leading to a loss of generality
(\emph{e.g.}, \citep{MBC-TonneausGroup-Tonneau2018Efficient} does
not consider configurations where all contacts are on vertical surfaces).

Alternatively, CBM algorithms plan a series of stances that are combined
to form an overall motion, using the posture generator to verify their
feasibility. Early CBM planners by Bretl \emph{et al.,} explicitly
enumerated all available stances, and used a posture generator based
on random sampling to verify the connectivity of the stance graph
\citep{CBM-Bretl2006JournalPaper,CBM-Hauser2008Primatives,CBM-Hauser2005Nongaited}.
These approaches assume that only a small set of discrete footholds
is available to the robot, and thus all stance combinations can be
considered. However, if the planner must identify footholds autonomously
on continuous surfaces, then this kind of explicit enumeration becomes
impractical. 

Another approach by Mordatch \emph{et al.} uses numerical optimisation
to simultaneously plan whole body motions and supporting contacts,
but again this approach relies on a pre-surveyed list of candidate
footholds, rather than identifying footholds autonomously \citep{CBM-Mordatch2012AnimatedFigures}.

The Contacts Very-Best First Planning (CVBFP) algorithm \citep{CVBFP-Escande2013MainPaper}
improves upon these approaches with a search process that plans contacts
between a given contact patch $\patch$ (attached to the robot) and
contact area $\area$ (attached to the environment) without specifying
the location of $\patch$ on $\area$. Instead, the posture generator
chooses the optimal contact location on $\area$ by minimising a potential
field, returning a transition configuration that forms the optimal
contact \citep{CVBFP-Escande2013MainPaper}. However, CVBFP assumes
that safe whole-body trajectories between neighbouring transition
configurations exist without verifying this to be the case. 

This is partially addressed by the multi-stage framework presented
in \citep{CBM-Ferrari2023Multicontact}, wherein the first stage uses
a Rapidly-exploring Random Tree method to compute a sequence of stances
and transition configurations, and the second stage generates a whole
body trajectory that executes the sequence. However, because the stance
sequence is planned prior to the trajectory, it remains possible that
a stance sequence may be returned for which a feasible trajectory
cannot be found.

It is notable that all of the approaches discussed are \emph{global}
planning methods, meaning that they return a complete plan from the
initial state to the goal state \citep{CBM-Bretl2006JournalPaper,CBM-Mordatch2012AnimatedFigures,CVBFP-Escande2013MainPaper,CBM-Ferrari2023Multicontact}.
Global approaches can often generate direct and efficient overall
motion plans, but they also typically require more time to plan than
local approaches, since planning the whole motion is generally more
costly than planning only the next few steps. Global approaches are
also less able to update their plans in response to gaining new information
about the environment and instead may need to repeat the entire planning
process from scratch.

Local planning methods for legged robots typically focus on\emph{
}walking over uneven terrains. For example, the technique in \citep{ML-RHP-Kalakrishnan2011Learning}
uses imitation learning to identify desirable footholds and plans
the torso trajectory over the next four steps. Likewise, in \citep{ML-RHP-Wang2022Learning},
whole-body trajectories are optimised over the next step cycle to
achieve intermediate planning goals that combine to form the complete
task. Alternatively, Melon et al. \citep{ML-RHP-Melon2021Receding}
proposes a receding horizon implementation of the popular Trajectory
Optimisation for Walking Robots (TOWR) framework \citep{ML-RHP-Winkler2018TOWR},
using a mix of heuristic methods and imitation learning to find effective
hot-start conditions for TOWR to execute at online rates.

While these approaches are able to plan multi-contact motions in real
time and update these plans easily, they lack the generality and formal
safety guarantees offered by conventional CBM approaches.

\subsection{Statement of Contributions}

We propose a novel multi-contact motion planning framework, including
a receding-horizon tree-search algorithm and a posture generator based
on vector-field inequalities. The proposed framework retains the generality
of other CBM approaches, while also improving upon the state-of-the-art
by:
\begin{itemize}
\item Rapidly re-planning in response to new information.
\item Improving robustness to planning problems caused by local minima in
the potential field.
\item Guaranteeing that kinematically feasible whole-body trajectories between
stances exist \emph{before} they are added to the search tree, verifying
assumptions made in \citep{CBM-Ferrari2023Multicontact,CVBFP-Escande2013MainPaper}.
\item Generating the stance, transition configuration, and whole-body trajectory
\emph{simultaneously}, removing the need for any additional stages
of planning.
\item Improving execution times without compromising on generality.
\end{itemize}
We demonstrate our approach by planning several challenging motions
for the Corin hexapod \citep{Corin-WebsiteEntry-UOM2020Corin}, including
scenarios where some or all of the robot's contacts are on vertical
surfaces (see \prettyref{sec:Results} for details). Finally, we use
Bayesian data analysis techniques to rigorously compare our planner's
performance to that of CVBFP \citep{CVBFP-Escande2013MainPaper}.

This work is an extension of previous work presented at TAROS 2025
\citep{TAROSPaper-2025-Derwent2025Multi}, which focussed on our posture
generator only. This work discusses the complete planning framework,
composed of both the search process \emph{and} the posture generator,
to offer a fuller picture of the planning process. We also present
new results from experiments and data analysis conducted after the
publication of \citep{TAROSPaper-2025-Derwent2025Multi} that include
new insights into the behaviour of the motion planner.

\subsection{Paper Organisation}

The remainder of the paper is organised as follows. Section~\ref{sec:Mathematical-Preliminaries}
provides some preliminary information regarding the mathematical notations
and terminology used throughout the paper. Section~\ref{sec:Receding-Horizon-Planning}
presents a high-level overview of our proposed planning algorithm
and its main components. Section~\ref{sec:Potential-Field} details
the guide path and potential field that are used to guide both the
tree search and the posture generation process. Section~\ref{sec:Posture-Generation}
discusses our novel posture generation approach, beginning with a
high-level introduction and going on to derive the relevant constraints
and objective functions. Section~\ref{sec:Local-Minima} describes
local minima problems that affect CBM planners and mitigation strategies.
Section~\ref{sec:Numerical-Integration} covers the numerical integration
method used to update the configuration model during posture generation
and contact drift correction. Section~\ref{sec:Results} discusses
the simulation experiments performed with the planner in several different
scenarios including a thorough statistical comparison against the
state-of-the-art. Finally, Section~\ref{sec:Conclusions} concludes
the paper and presents some future work.

\section{Preliminaries\label{sec:Mathematical-Preliminaries}}

This section introduces the concepts from dual quaternion algebra
that are used for robot modelling and control throughout this work
and defines several terms and symbols used in later sections.
\global\long\def\frame#1{\mathscr{F}_{#1}}%
\global\long\def\worldframe{\frame{\workspace}}%

\subsection{Dual Quaternion Algebra}

The planning framework we present relies heavily on geometrical primitives,
such as planes, lines, and points, in addition to rigid transformations,
twists, and wrenches, all of which are elegantly represented using
dual quaternion algebra. Additionally, several robot modelling and
control techniques that are particularly useful for solving posture
generation problems have been developed that utilise the strong algebraic
properties of dual quaternion algebra, making this an attractive choice.

Quaternions belong to the set $\mathbb{H}\triangleq\{h_{1}+\imi h_{2}+\imj h_{3}+\imk h_{4}\::\:h_{1},h_{2},h_{3},h_{4}\in\mathbb{R}\}$,\LyXZeroWidthSpace{}
where $\imi$, $\imj$, and $\imk$ denote imaginary units such that
$\imi^{2}=\imj^{2}=\imk^{2}=\imi\imj\imk=-1$. Quaternions can represent
3D orientations and rotations using elements of the subset $\mathbb{S}^{3}\triangleq\left\{ \quat h\in\mathbb{H}\::\:\norm{\quat h}=1\right\} $,
as well as positions using elements of the subset $\mathbb{H}_{p}\triangleq\{\quat h\in\mathbb{H}\::\:\real{\quat h}=0\}$
\citep{DQ-Adorno2017Fundamentals}.\emph{ Dual} quaternions extend
quaternions and belong to the set $\mathbb{\mathcal{H}}\triangleq\{\quat h+\varepsilon\quat h'\::\:\boldsymbol{h},\quat h'\in\mathbb{H},\,\dual^{2}=0,\,\dual\neq0\}$.
The set of \emph{unit} dual quaternions, $\dq{\mathcal{S}}\triangleq\{\dq h\in\mathbb{\mathcal{H}}\::\:\norm{\dq h}=1\}$,
represent rigid transformations in 3D space, and any $\dq x\in\dq{\mathcal{S}}$
can be written as $\dq x=\quat r+\dual\frac{1}{2}\quat p\quat r,$
where $\quat r\in\mathbb{S}^{3}$ and $\quat p\in\mathbb{H}_{p}$
denote the rotation and translation components of the rigid transformation
respectively \citep{DQ-Adorno2017Fundamentals}. Analogously to $\mathbb{H}_{p}$,
we also define the subset $\mathcal{H}_{p}\triangleq\left\{ \quat h+\dual\quat h':\quat h,\quat h'\in\mathbb{H}_{p}\right\} $,
which is useful for modelling certain primitives.

The coefficients of elements of any of the aforementioned sets can
be bijectively mapped into vectors. For example,
\begin{equation}
\begin{aligned}\vecthree{\imi h_{1}\!+\!\imj h_{2}\!+\!\imk h_{3}}\!= & \!\begin{bmatrix}h_{1} & h_{2} & h_{3}\end{bmatrix}^{T}\\
\vecfour{h_{1}\!+\!\imi h_{2}\!+\!\imj h_{3}\!+\!\imk h_{4}}\!= & \!\begin{bmatrix}h_{1} & h_{2} & h_{3} & h_{4}\end{bmatrix}^{T}.
\end{aligned}
\label{eq:vec-operators}
\end{equation}

This property is very useful for performing numerical optimisation,
discussed in more detail in Section~\prettyref{subsec:VFI-Preliminaries}.

\subsection{Terminology}

The \emph{workspace} is written as $\workspace\subset\purequatset$,
with its fixed coordinate frame denoted $\frame{\workspace}$. The
region of $\workspace$ occupied by obstacles that the robot cannot
intersect (\emph{the obstacle region}) is represented by $\obstacle\subset\workspace$.

In this work, \emph{configurations} are represented in the form $\config=\left[\begin{array}{ccc}
\vecfour{\rot}^{\transpose} & \vecthree{\point}^{\transpose} & \jointangles^{\transpose}\end{array}\right]^{\transpose}\in\mathbb{R}^{n}$ such that the robot's pose is given by the unit dual quaternion $\dq x=\quat r+\dual\frac{1}{2}\quat p\quat r$,
the vector $\jointangles\in\mathbb{R}^{n_{\theta}}$ represents the
joint coordinates, and $n=n_{\theta}+7$. Each joint may be either
prismatic or revolute with a bounded range of motion. The set of all
configurations is the configuration space $\configspace$.

A contact area $\area$ is a 2D surface in $\workspace$ on which
a contact may be formed. The set of all contact areas is denoted $\allareas\triangleq\left\{ \area_{1},\area_{2},\ldots,\area_{\numareas}\right\} $.

A contact patch $\patch_{\indexrobot}$ is an area or point on the
robot that is permitted to participate in contacts. For a humanoid
robot, contact patches may include feet and hands, as well as knees,
elbows, or other parts of the robot. For hexapods and quadrupeds,
this is more likely to include only the robot's feet. The set of contact
patches is denoted $\allcontactpatches$.

A \emph{contact} $\contact_{\indexrobot,\indexenv}$ between a contact
patch $\patch_{\indexrobot}\in\allcontactpatches$ and a contact area
$\area_{\indexenv}\in\mathcal{E}$ is denoted $\contact_{\indexrobot,\indexenv}=\left(\patch_{\indexrobot},\area_{\indexenv},\contactpoint{\indexenv}\right)$,
where $\contactpoint{\indexenv}\in\area_{\indexenv}$ represent the
position of the contact in the reference frame of area $\area_{\indexenv}$.

A \emph{stance} $\stance$ refers to a set of multiple contacts maintained
simultaneously, written as $\stance\triangleq\left\{ \contact_{1},\contact_{2},\ldots,\contact_{\numcontacts}\right\} $
\citep{CBM-Bretl2006JournalPaper}. The \emph{stance space} is the
set of all stances, denoted $\stancespace$. The set of configurations
that satisfy a given stance $\stance$ is written as $\configspace_{\stance}\subset\configspace$.

A configuration is said to be \emph{feasible} if it respects the constraints
on the robot (\emph{e.g.}, collision avoidance, maintaining quasi-static
balance, maintaining contacts, kinematic constraints, \emph{etc.}).
The set of all feasible configurations is written as $\feasiblespace\subset\configspace$.
Different constraints are associated with different stances (\emph{e.g.},
which configurations are balanced depends on the stance), and as such
it is often useful to consider $\feasiblespace_{\stance}\subset\configspace_{\stance}$,
the set of all feasible configurations in stance $\stance$. Stance
$\stance$ is said to be feasible if $\feasiblespace_{\stance}\neq\emptyset$.

If two neighbouring stances $\stance_{1}$ and $\stance_{2}$ are
linked by a configuration $\transition\in\mathcal{\feasiblespace}_{\stance_{1}}\cap\feasiblespace_{\stance_{2}}$,
then $\transition$ is said to be a \emph{transition configuration}
between $\stance_{1}$ and $\stance_{2}$.

\subsection{Vector-Field Inequalities\label{subsec:VFI-Preliminaries}}

Using dual quaternion algebra, one can straightforwardly formulate
task-space constrained controllers with geometrical constraints in
the form of quadratic programs with linear constraints in the control
inputs. As constrained controllers based on vector-field inequalities
play a crucial role in our multi-contact motion planner, let us briefly
recapitulate their basic formulation. Given a task vector $\myvec x\triangleq\myvec x\left(\config\right)\in\mathbb{R}^{m}$,
which might represent, for instance, the robot pose, position, or
orientation, and a constant desired task vector $\myvec x_{d}\in\mathbb{R}^{m}$,
we can minimise the error vector $\tilde{\myvec x}\triangleq\myvec x-\myvec x_{d}$
using the control law
\begin{equation}
\argminimtwo{\continput\in}{\dot{\config}}{\norm{\mymatrix J_{\tilde{\myvec x}}\dot{\myvec q}+\eta_{o}\tilde{\myvec x}}_{2}^{2}+\lambda^{2}\norm{\dot{\myvec q}}_{2}^{2}}{\boldsymbol{W}_{1}\left(\myvec q\right)\dot{\boldsymbol{q}}\preceq\boldsymbol{w}_{1}\left(\myvec q\right)}{\boldsymbol{W}_{2}\left(\myvec q\right)\dot{\boldsymbol{q}}=\boldsymbol{w}_{2}\left(\myvec q\right),}\label{eq:standard-control-problem}
\end{equation}
where $\mymatrix J_{\tilde{\myvec x}}\triangleq\mymatrix J_{\tilde{\myvec x}}\left(\config\right)=\partial\tilde{\myvec x}/\partial\myvec q$,
$\eta_{o}\in\left(0,\infty\right)$ and $\lambda\in(0,\infty)$. The
matrix/vector pair $\boldsymbol{W}_{1}\left(\myvec q\right)\in\mathbb{R}^{\ell_{I}\times n}$
and $\boldsymbol{w}_{1}\left(\myvec q\right)\in\mathbb{R}^{\ell_{I}}$
impose $\ell_{I}$ linear \emph{inequality} constraints on $\dot{\myvec q}$,
whereas $\boldsymbol{W}_{2}\left(\myvec q\right)\in\mathbb{R}^{\ell_{E}\times n}$
and $\boldsymbol{w}_{2}\left(\myvec q\right)\in\mathbb{R}^{\ell_{E}}$
impose $\ell_{E}$ linear \emph{equality} constraints on $\dot{\myvec q}$
\citep{DQ-Marinho2019VFIs}.

The vector-field inequality (VFI) framework allows the transformation
of non-linear geometric constraints on $\myvec q$ into linear constraints
on $\dot{\myvec q}$ that ensure that the original constraints on
$\myvec q$ are met \citep{DQ-Marinho2019VFIs}. This requires only
a differentiable signed distance function $d\triangleq d(\myvec q)$
between a geometrical entity kinematically coupled to the robot and
another geometrical primitive in the task space, and the Jacobian
matrix $\mymatrix J_{d}\triangleq\mymatrix J_{d}\left(\config\right)\in\mathbb{R}^{1\times n}$
such that $\dot{d}=\mymatrix J_{d}\dot{\myvec q}$ \citep{DQ-Marinho2019VFIs}.

\begin{figure}
\centering{}\includegraphics[width=1\columnwidth]{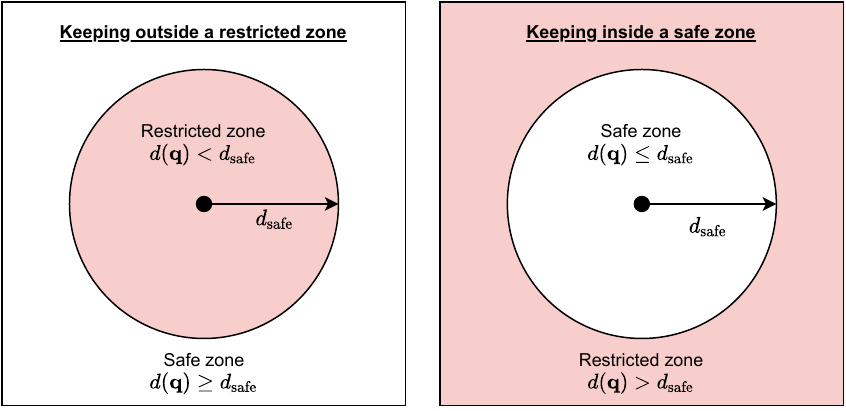}
\caption{Illustrations of keeping the robot outside a restricted zone (\emph{left}),
in which case constraint \eqref{eq:outside-safe-zone} is applied,
and inside a safe zone (\emph{right}), in which case constraint \eqref{eq:inside-safe-zone}
is applied. \label{fig:safe-regions}}
\end{figure}

To keep a robot entity outside of a given zone, \emph{e.g.,} for collision
avoidance, a signed distance $d(\myvec q)$ is defined between the
robot body and the restricted zone and a minimum safe distance requirement
$d(\myvec q)\geq d_{\text{safe}}$, shown on the \emph{left} in Figure~\ref{fig:safe-regions},
is imposed. By defining an error term $\tilde{d}(\myvec q)\triangleq d(\myvec q)-d_{\textrm{safe}}$
the requirement is rewritten as $\tilde{d}(\myvec q)\geq0$. Hence,
using the VFI framework, this requirement is enforced using the constraint
\begin{equation}
\dot{\tilde{d}}(\myvec q)\geq-\eta_{d}\tilde{d}(\myvec q),\label{eq:outside-VFI-baby-ver}
\end{equation}
where $\eta_{d}\in\left(0,\infty\right)$ determines the maximum approach
velocity of the robot body towards the restricted zone as a function
of the distance to the minimum safe distance \citep{DQ-Marinho2019VFIs}.
By applying Gronwall's Lemma, it can be shown that if the robot body
lies outside of the restricted zone at time $t=0$, meaning $\tilde{d}\geq0$,
then $\tilde{d}(\config)\geq0$ for all times $t>0$ \citep{DQ-Marinho2019VFIs}.

Given that $\dot{d}_{\textrm{safe}}=0$ for all $t$, then $\dot{\tilde{d}}(\myvec q)=\mymatrix J_{d}\dot{\myvec q}$.
Hence, \eqref{eq:outside-VFI-baby-ver} is rewritten as

\begin{equation}
-\mymatrix J_{d}\dot{\myvec q}\leq\eta_{d}\tilde{d}(\myvec q).\label{eq:outside-safe-zone}
\end{equation}
Similarly, to keep a robot entity inside a given zone, \emph{e.g.},
keeping the centre of mass inside a support region, the requirement
is restated as $\tilde{d}(\myvec q)\leq0$ (shown on the \emph{right}
in Figure~\ref{fig:safe-regions}) to obtain the constraint

\begin{equation}
\dot{\tilde{d}}(\myvec q)\leq-\eta_{d}\tilde{d}(\myvec q)\iff\mymatrix J_{d}\dot{\myvec q}\leq-\eta_{d}\tilde{d}(\myvec q).\label{eq:inside-safe-zone}
\end{equation}
Robot entities and zones of interest (\emph{i.e.}, restricted zones
and safe zones) can take on a wide variety of forms, with distance
functions and corresponding Jacobians defined in the literature for
combinations of points, lines, planes and many others \citep{DQ-Marinho2019VFIs,DQ-Pereira2022CollisionAvoidanceAndCones,DQ-Quiroz-Omana2019AngularConstraint}.

\section{An Overview of Receding Horizon Planning\label{sec:Receding-Horizon-Planning}}

This section offers a high level overview of our algorithm, focussing
on the receding-horizon, breadth-first tree-search algorithm that
improves upon the state-of-the-art by:
\begin{itemize}
\item Quickly re-planning in response to new information;
\item Having only a single stage and a consistent policy for expanding nodes,
which reduces the complexity of the implementation;
\item Allowing the user to choose how much to prioritise performance versus
quality of planning outputs by adjusting the horizon depth parameter
$\horizondepth$.
\end{itemize}

\subsection{Algorithm Description\label{subsec:High-Level-Summary}}

As discussed in \prettyref{sec:Introduction}, many CBM algorithms
are \emph{global} planners, meaning that they generate a single, complete
motion plan from the robot's starting state to its goal state \citep{CBM-Bretl2006JournalPaper,CBM-Mordatch2012AnimatedFigures,CVBFP-Escande2013MainPaper,CBM-Ferrari2023Multicontact}.
By contrast, given the robot’s \emph{current} state and the goal state,
a\emph{ receding horizon planner }considers possible motions until
a planning horizon is reached, at which point the most promising step
from the current state is executed. This process repeats until the
robot arrives at the goal state. The advantage of doing so is that
the plan can be continuously adapted according to the best current
information the robot has about the environment and its own state,
providing reactivity and adaptability.

In this work, robot states are represented by nodes, written as $\node_{i}=\fullnode{i-1}i$,
where $\stance_{i}\in\stancespace$ and $\config_{i}\in\feasiblespace_{\stance_{i}}$
denote the robot's stance and feasible configuration, whereas $\node_{i-1}$
denotes the parent node of $\node_{i}$ (\emph{i.e.}, a previous state)
and $\trajectory{i-1}i\subset\configspace$ denotes a feasible, whole-body,
quasi-static, configuration space trajectory connecting node $\node_{i-1}$
to node $\node_{i}$. Note that since the trajectory $\trajectory{i-1}i$
is quasi-static, the reverse trajectory $\trajectory i{i-1}$ can
be obtained by reversing the order of the elements in $\trajectory{i-1}i$. 

\begin{figure*}[t]
\centering{}\includegraphics[width=1\textwidth]{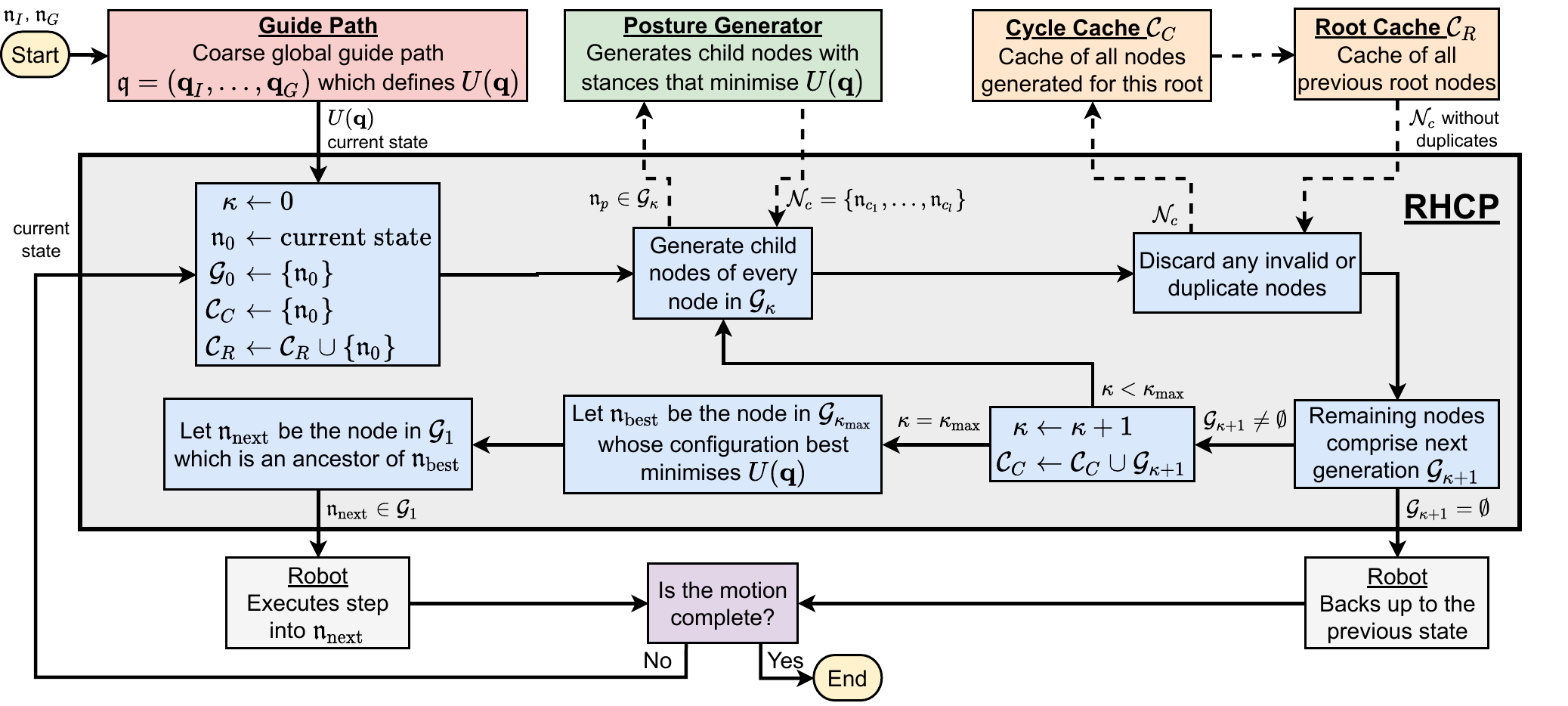}
\caption{Flowchart illustrating the RHCP planning algorithm.\label{fig:RHCP-Flowchart}}
\end{figure*}
A flowchart of the proposed Receding Horizon Contact Planning (RHCP)
algorithm is given in Figure~\ref{fig:RHCP-Flowchart}. First, a
coarse global path is generated (shown in \emph{red}) to guide the
contact planning. Then, the main planning loop begins, with the initial
node being re-designated as the root node $\rootnode$ and the root
generation\footnote{Except for $\generation_{0}$, which contains only the root node,
the $\kappa$th generation, $\generation_{\kappa}$, is written as
a set of nodes that contains all valid, non-duplicate child nodes
of all nodes in the generation $\generation_{\kappa-1}$.} being defined as $\generation_{0}=\left\{ \rootnode\right\} $. The
posture generator (shown in \emph{green}) is then called to find the
child nodes of $\rootnode$, with each one representing a possible
step that the robot could take. The cycle cache $\cache_{C}$ and
root cache $\cache_{R}$ (shown in \emph{orange}) are used to find
and prune any duplicate nodes, and those that remain comprise the
next generation $\generation_{1}$. If the horizon has not been reached,
then the current step $\kappa$ is incremented and the posture generator
is called again to populate the new generation $\generation_{2}$.
Once the horizon is reached (\emph{i.e.,} $\indexgeneration=\kmax$),
the best node in generation $\generation_{\kmax}$ is found and the
robot executes the step in $\generation_{1}$ that brings it towards
this state. The process then repeats, with the resulting robot state
becoming the new root node, until the goal is reached. The next subsections
detail each of those mechanisms.

\subsubsection{Potential Field Guide}

The guide path is comprised of a sequence of waypoint configurations
$\configguidepath\triangleq\left(\config_{I},\ldots,\config_{\goal}\right)$,
much like in \citep{CVBFP-Escande2013MainPaper}, and the waypoint
configurations may be provided by the user or generated autonomously
\citep{CVBFP-Bouyarmane2009GuidePathGeneration}. Also like \citep{CVBFP-Escande2013MainPaper},
a potential field $\potfield:\configspace\rightarrow\left[0,\infty\right)$
is defined based on $\guidepath$ that incentivises the robot to move
towards the goal state while keeping close to the guide path. The
formulation of $\potential{\config}$ is discussed further in \prettyref{sec:Potential-Field}.

\subsubsection{Child Generation}

The main planning loop begins by setting the generation count $\indexgeneration\isnonnegativeint$
to $0$ and storing the current state in the root node $\rootnode$.
It is assumed that all contact patches $\patch\in\allcontactpatches$
are in contact in the root stance $\stance_{0}$. The root generation
$\generation_{0}$, cycle cache $\cache_{C}$, and root cache $\cache_{R}$
are also each defined at this stage as sets containing only $\rootnode$.
Every parent node $\node_{p}\in\generation_{0}$ is then expanded
(in this case only $\rootnode$), with the posture generator being
called to generate the $\ell$ child nodes, each designated as $\node_{c_{k}}=\fullnode{p_{k}}{c_{k}},\,k\in\left\{ 1,\ldots,\ell\right\} $.

One child node is specified for each contact patch-area pair $\left(\patch_{\indexrobot},\area_{\indexenv}\right)$,
each containing a stance $\stance_{c_{k}}$ that differs from its
parent stance $\stance_{p_{k}}$. The chosen contact patch $\patch_{\indexrobot}$
is moved to the location on contact area $\area_{\indexenv}$ that
best minimises $\potential{\config}$. All of the remaining contacts
in $\stance_{p_{k}}$ are maintained in $\stance_{c_{k}}$. 

\subsubsection{Pruning}

Any invalid or duplicate child nodes are then rejected. A node is
invalid if the posture generator fails to find a feasible configuration
or linking trajectory (\emph{e.g.}, the requested contact cannot be
safely broken). A node is considered a duplicate if its stance $\stance_{c}$
is too similar to that of any node in the cycle cache $\cache_{C}$
or the root cache $\cache_{R}$. Two stances, $\stance_{1}$ and $\stance_{2}$,
are considered `too similar' if, for every contact $\contact_{i}\in\stance_{1}$,
there exists a contact $\contact_{j}\in\stance_{2}$ involving the
same contact patch-area pair as in $\contact_{i}$ such that the distance
between $\contact_{i}$ and $\contact_{j}$ is below a user-defined
threshold distance $\mindist\ispositive$.

\subsubsection{Completing the Cycle}

The remaining valid, non-duplicate child nodes comprise the next generation
$\generation_{1}$ and are added to the cycle cache $\cache_{C}$.
If at least one such node is returned (\emph{i.e.}, $\generation_{1}\neq\emptyset$),
then the generation count $\indexgeneration$ is incremented. If $\indexgeneration<\horizondepth$,
then the cycle repeats, this time expanding every node in $\generation_{1}$,
producing the next generation $\generation_{2}$, and so on until
$\generation_{\kmax}$.

Once $\indexgeneration=\horizondepth$, the node in the final generation
$\generation_{\horizondepth}$ that best minimises the potential field
$\potential{\config}$ is identified, designated $\bestnode$. A backward
search is conducted to find the node in $\generation_{1}$ from which
$\bestnode$ descends, designated $\nextnode$, and the motion connecting
$\rootnode$ to $\nextnode$ is executed by the robot. 

If this motion brings the robot into the goal state, then the planning
concludes. If not, the planner is called again to determine the \emph{next}
step to take, with the current robot state becoming the new root node
and being added to the root cache $\cache_{R}$, preventing it from
being generated again in future cycles.

\subsubsection{Retreating}

If no valid, non-duplicate child nodes remain to populate a generation
$\generation_{\kappa+1}$, then the robot returns to its previous
state (\emph{i.e.}, the parent node of $\rootnode$) because all descendants
of $\node_{0}$ will eventually lead to a dead end at the $\kappa$th
step (\emph{i.e.}, there are no valid stances after $\generation_{\kappa}$).
The planning cycle then repeats, as depicted in Figure~\ref{fig:RHCP-Flowchart},
in order to identify the \emph{next-best} line of exploration. Since
this time $\rootnode$ is already in the root cache $\cache_{R}$,
it is considered a duplicate and is excluded from $\generation_{1}$.
The planner is thus prevented from returning to previously occupied
root nodes that are known to be `dead ends'. 

If the robot is instructed to retreat while in its initial state,
when there is no previous state to retreat to, then the planning is
considered to have failed, and no route is returned.

\subsection{Analysis}

The maximum number of posture generator calls per cycle is

\begin{equation}
n_{\max}=\sum_{\indexgeneration=1}^{\horizondepth}\left(\numpatches\numareas\right)^{\indexgeneration},\label{eq:number-of-calls}
\end{equation}
where $\numpatches,\numareas\ispositiveint$ denote the number of
contact patches in $\allcontactpatches$ and the number of contact
areas in $\allareas$, respectively. It is clear from \eqref{eq:number-of-calls}
that the maximum number of calls increases polynomially with respect
to the number of contact patches or contact areas, and exponentially
with respect to the horizon depth parameter. 

Note, however, that fewer total calls may be made if one or more nodes
are rejected (\emph{i.e.,} they are invalid or duplicates) and hence
their children are not generated. The number of calls may be further
reduced by only attempting placements on contact areas within the
robot's reach. 

Another important observation is that, as RHCP generates only one
child node per contact patch-area pair (for a given parent node),
it is not possible for the algorithm to guarantee completeness. That
is to say, there may be possible motion plans that the planner fails
to find due to it not considering other locations at which contacts
could be formed between the same contact patch-area pair. 

Finally, compared to other CBM motion planners, \allowbreak  RHCP
is especially well suited to rapid re-planning in response to new
information or changes in the environment. This is because the motion
plan may be updated by merely repeating the previous planning cycle
to generate a new next-step recommendation, rather than repeating
a costly global planning operation. This is particularly useful if
the environment is not fully mapped beforehand, as the robot can quickly
update the plan as its sensors gather more data.

\section{Potential-field-based Guide Path\label{sec:Potential-Field}}

Recall that the configuration-space guide path $\guidepath\in\configspace$
comprises a sequence of waypoint configurations 
\[
\configguidepath\triangleq\left(\config_{1},\ldots,\config_{G}\right).
\]
We also define workspace waypoints for each contact patch $\patch_{\indexrobot}$,
written as $\patchguidepath_{\indexrobot}\triangleq\left\{ \fkm i{\config}\ispurequat\::\ \config\in\configguidepath\right\} $,
where $\fkm{\indexrobot}{\config}:\configspace\rightarrow\purequatset$
denotes the workspace position of the coordinate frame attached to
$\patch_{\indexrobot}$ in configuration $\config$.

\begin{figure}[t]
\centering{}%
\fcolorbox{black}{white}{\includegraphics[width=1\columnwidth]{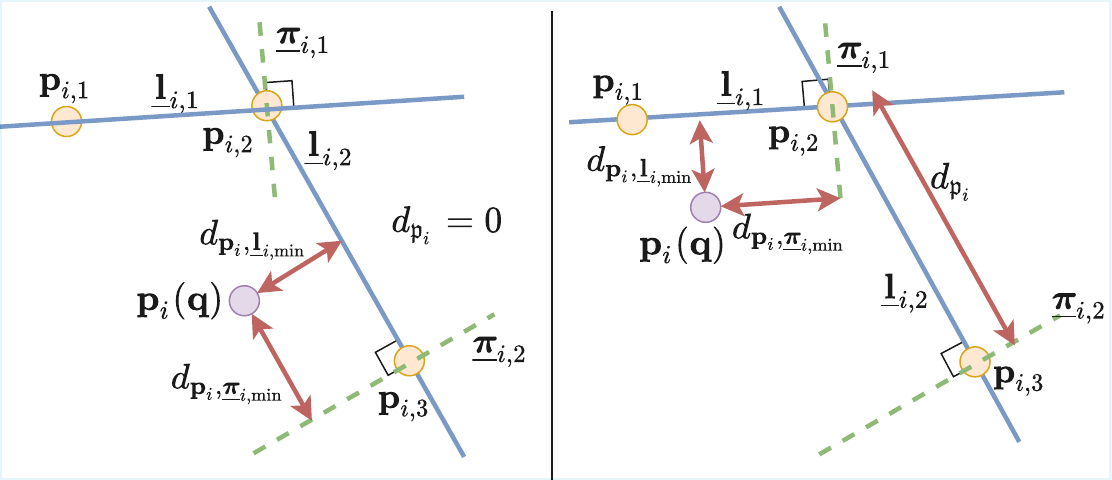}} \caption{Illustration of primitives used in contact patch-specific potential
field terms. Two scenarios are given for different positions of the
contact patch (shown as a \emph{purple} \emph{circle}). The remaining
path distance $\protect\monodistance[\protect\patchguidepath_{\protect\indexrobot}]$
is shown in \emph{red} in the scenario on the \emph{right} but is
absent from the \emph{left} scenario, since $\protect\monodistance[\protect\patchguidepath_{\protect\indexrobot}]=0$
in this case. \label{fig:Field-diagram-part-1}}
\end{figure}

Each successive pair of waypoints $\left(\point_{\indexrobot,j},\point_{\indexrobot,j+1}\right)$
is used to define a dual quaternion line $\line_{i,j}\in\mathcal{H}_{p}\cap\dq{\mathcal{S}}$
(illustrated in Figure~\ref{fig:Field-diagram-part-1}), written
as \citep{DQ-Adorno2017Fundamentals}
\begin{equation}
\line_{\indexrobot,j}\triangleq\direction_{\indexrobot,j}+\dual\left(\crossproduct{\point_{\indexrobot,j}}{\direction_{\indexrobot,j}}\right),\label{eq:line}
\end{equation}
where $\direction_{\indexrobot,j}\isunitpurequat$ is given by
\begin{equation}
\direction_{\indexrobot,j}\triangleq\frac{\point_{\indexrobot,j+1}-\point_{\indexrobot,j}}{\norm{\point_{\indexrobot,j+1}-\point_{\indexrobot,j}}}.\label{eq:line-direction}
\end{equation}
 We also define a dual quaternion plane $\plane_{\indexrobot,j}\isplane$
attached to $\point_{\indexrobot,j+1}$ whose normal vector is equal
to $-\direction_{\indexrobot,j}$, written as \citep{DQ-Adorno2017Fundamentals}
\begin{equation}
\plane_{\indexrobot,j}\triangleq-\direction_{\indexrobot,j}+\dual\innerproduct{\point_{\indexrobot,j+1}}{-\direction_{\indexrobot,j}}.\label{eq:plane}
\end{equation}

The closest line segment to $\fkm{\indexrobot}{\config}$ is denoted
$\line_{\indexrobot,\closest}$.\footnote{Each line segment is treated as finite and bounded by the waypoints
at each end for the purposes of this check.} If multiple line segments are equally close to $\fkm{\indexrobot}{\config}$,
then that which results in the lowest overall potential is chosen.
The waypoint terminating $\line_{\indexrobot,\closest}$ is denoted
$\point_{\indexrobot,\closest}$, while the plane intersecting that
point is denoted $\plane_{\indexrobot,\closest}$. We thus define
the contact patch-specific sub-field for contact patch $\patch_{i}$
as
\begin{equation}
\potfield_{\patch_{\indexrobot}}\left(\config\right)\triangleq\squaredistance{\point_{\indexrobot}}{\line_{\indexrobot,\closest}}+\potfieldweighting\left(\squaredistance{\point_{\indexrobot}}{\plane_{\indexrobot,\closest}}+\lumppot{\indexrobot}\right),\label{eq:field-patch-subterm}
\end{equation}
where $\potfieldweighting\ispositive$ is a user-defined weighting
term, and the terms $\squaredistance{\point_{\indexrobot}}{\line_{\indexrobot,\closest}}\isnonnegative$
and $\squaredistance{\point_{\indexrobot}}{\plane_{\indexrobot,\closest}}\isnonnegative$
denote the point-to-line square distance between $\fkm{\indexrobot}{\config}$
and $\line_{\indexrobot,\closest}$, and the point-to-plane square
distance between $\fkm{\indexrobot}{\config}$ and $\plane_{\indexrobot,\closest}$,
respectively \citep{DQ-Marinho2019VFIs}.\footnote{Note that in Figure~\ref{fig:Field-diagram-part-1} the signed distances
$\signdistance{\point_{\indexrobot}}{\line_{\indexrobot,\closest}}$,
$\signdistance{\point_{\indexrobot}}{\plane_{\indexrobot,\closest}}$and
$\monodistance[\patchguidepath_{\indexrobot}]$ are shown, whereas
\eqref{eq:field-patch-subterm} uses the square distances, given by
$\squaredistance{\point_{\indexrobot}}{\line_{\indexrobot,\closest}}=\signdistance[2]{\point_{\indexrobot}}{\line_{\indexrobot,\closest}}$,
$\squaredistance{\point_{\indexrobot}}{\plane_{\indexrobot,\closest}}=\signdistance[2]{\point_{\indexrobot}}{\plane_{\indexrobot,\closest}}$
and $\lumppot{\indexrobot}=\monodistance[\patchguidepath_{\indexrobot}]^{2}$.
This is because the square distances are differentiable everywhere,
whereas the signed distance time derivatives are not defined at zero
\citep{DQ-Marinho2019VFIs}.} The term $\lumppot{\indexrobot}\isnonnegative$ denotes the square
of the distance\emph{ along the path} from $\point_{\indexrobot,\closest}$
to the final waypoint, given by
\begin{equation}
\lumppot{\indexrobot}\triangleq\sum_{j=k+1}^{\numwayps-1}\squaredistance{\point_{\indexrobot,j}}{\point_{\indexrobot,j+1}},\label{eq:lump-potential}
\end{equation}
where $\numwayps$ is the number of waypoints, $\line_{\indexrobot,k}=\line_{\indexrobot,\closest}$,
and $\squaredistance{\point_{\indexrobot,j}}{\point_{\indexrobot,j+1}}$
denotes the point-to-point square distance from $\point_{\indexrobot,j}$
to $\point_{\indexrobot,j+1}$ \citep{DQ-Marinho2019VFIs}.

Each sub-field $\potfield_{\patch_{\indexrobot}}\left(\config\right)$
is minimised by keeping $\patch_{\indexrobot}$ close to its guide
path (minimising $\squaredistance{\point_{\indexrobot}}{\line_{\indexrobot,\closest}}$)
and moving along the path towards the goal (minimising $\squaredistance{\point_{\indexrobot}}{\plane_{\indexrobot,\closest}}+\lumppot{\indexrobot}$).

\begin{figure*}[t]
\noindent\begin{minipage}[t]{1\columnwidth}%
\centering %
\fcolorbox{black}{white}{\includegraphics[width=1\textwidth]{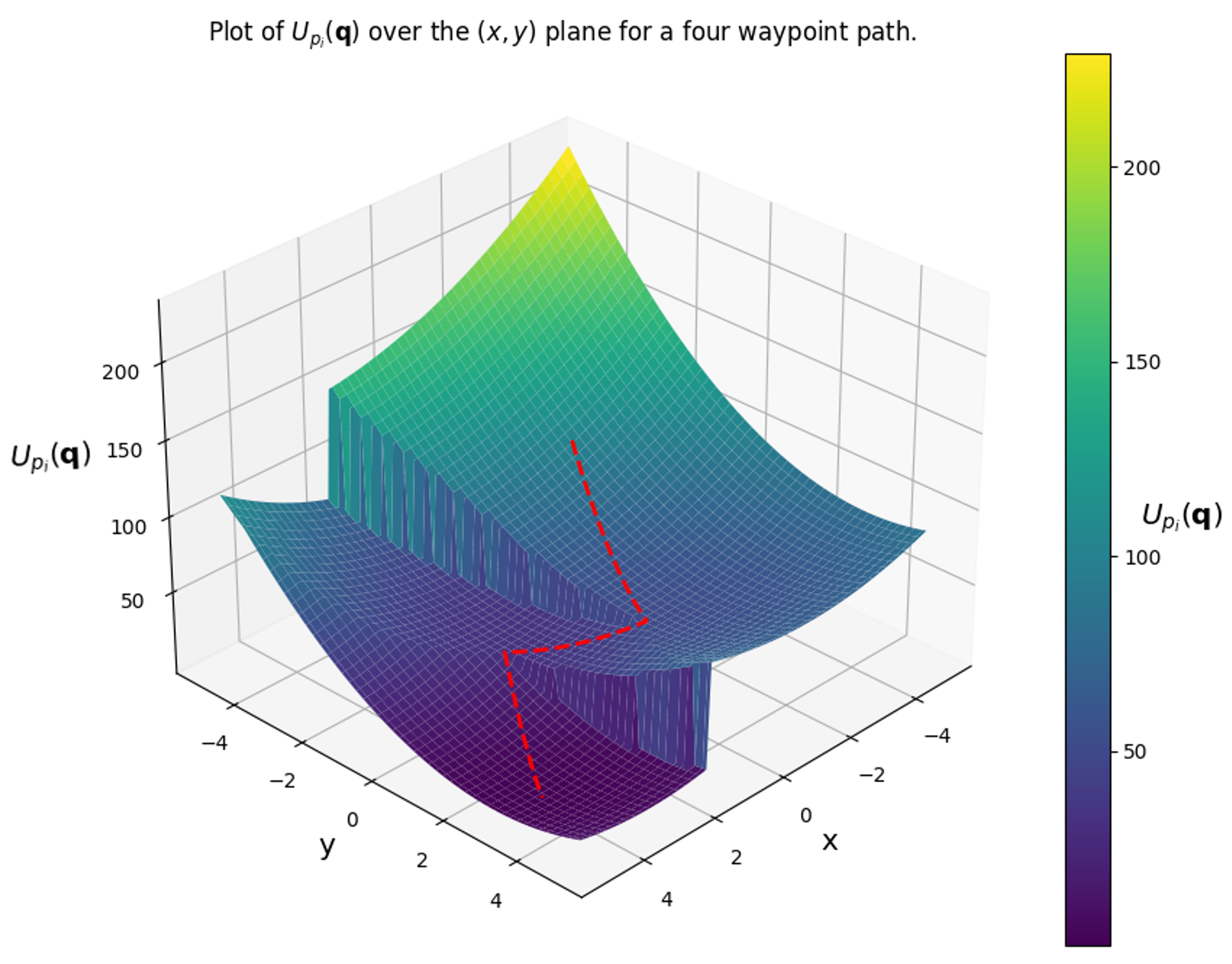}} \caption{3D plot of the potential field term $\protect\potfield_{\protect\patch_{\protect\indexrobot}}\left(\protect\config\right)$
over the $(x,y)$ plane for an example guide path (shown in \emph{red}).
\label{fig:potential-field-3D-plot}}
\end{minipage}\hfill{}%
\noindent\begin{minipage}[t]{1\columnwidth}%
\centering %
\fcolorbox{black}{white}{\includegraphics[width=1\textwidth]{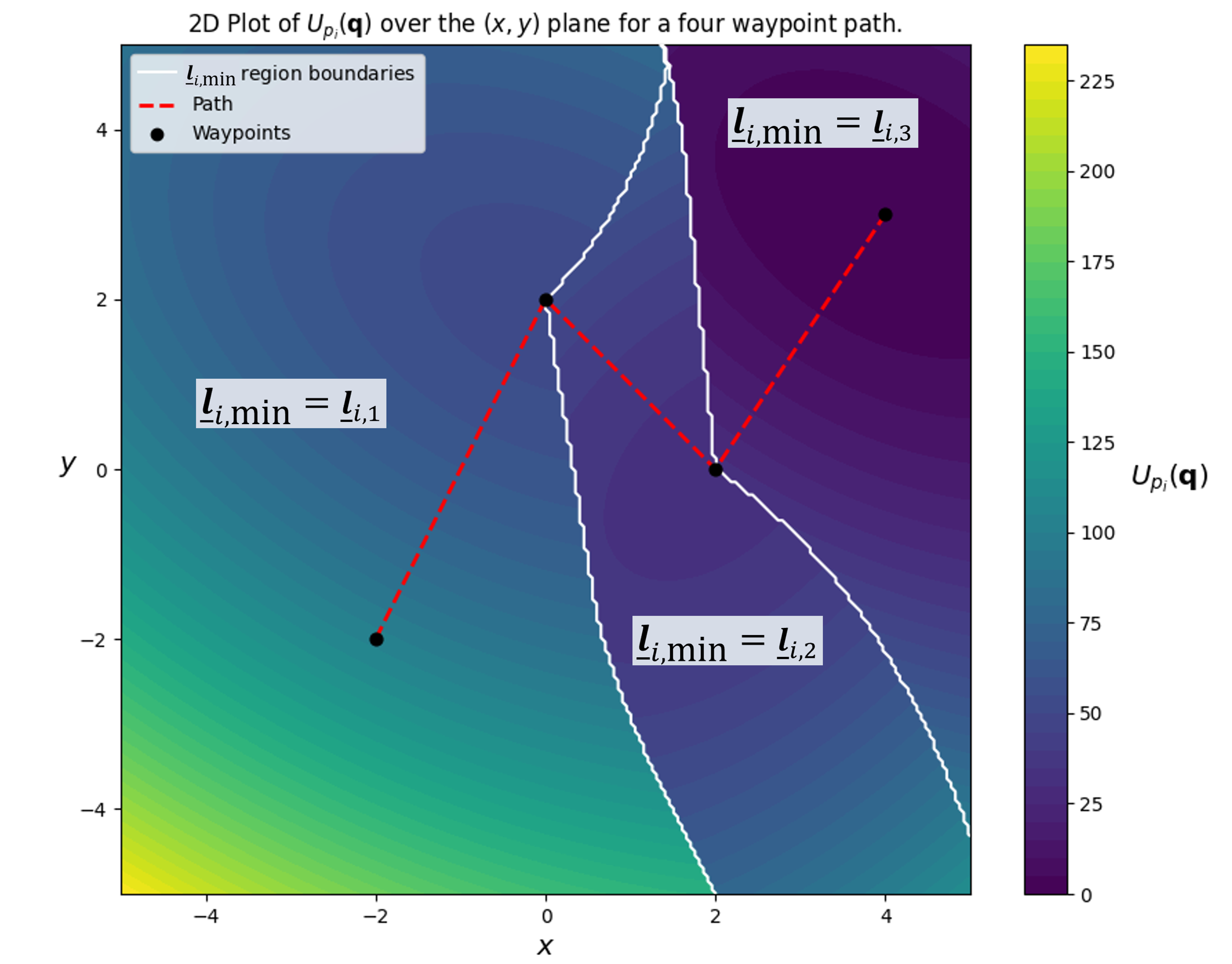}} \caption{2D cross-section plot of $\protect\potfield_{\protect\patch_{\protect\indexrobot}}\left(\protect\config\right)$
over the $(x,y)$ plane for the example path in Figure~\ref{fig:potential-field-3D-plot}.
The \emph{white} boundaries indicate which line segment serves as
$\protect\line_{\protect\indexrobot,\protect\closest}$. \label{fig:potential-field-2D-plot}}
\end{minipage}
\end{figure*}

For a given line segment, $\potfield_{\patch_{\indexrobot}}\left(\config\right)$
is non-negative and continuously differentiable with respect to $\config$.
An illustration of $\potfield_{\patch_{\indexrobot}}\left(\config\right)$
for a guide path containing four waypoints over a 2D cross-section
of the workspace is shown in Figure~\ref{fig:potential-field-3D-plot},
whereas which line segment is serving as $\line_{\indexrobot,\closest}$
is highlighted by the regions shown in Figure~\ref{fig:potential-field-2D-plot}.
Both figures show how the potential field monotonically decreases
as the contact patch $\patch_{\indexrobot}$ moves from the initial
waypoint $\point_{i}\left(\config_{1}\right)$ to the final waypoint
$\point_{i}\left(\config_{4}\right)$.

Having defined the sub-field for each patch $\patch_{\indexrobot}$,
the full potential field is written
\begin{equation}
\potential{\config}\triangleq\sum_{i=1}^{\numpatches}\potfield_{\patch_{\indexrobot}}\left(\config\right).\label{eq:pot-field-v1}
\end{equation}

\section{Posture Generation\label{sec:Posture-Generation}}

\begin{figure*}[t]
\centering{}\includegraphics[width=1\textwidth]{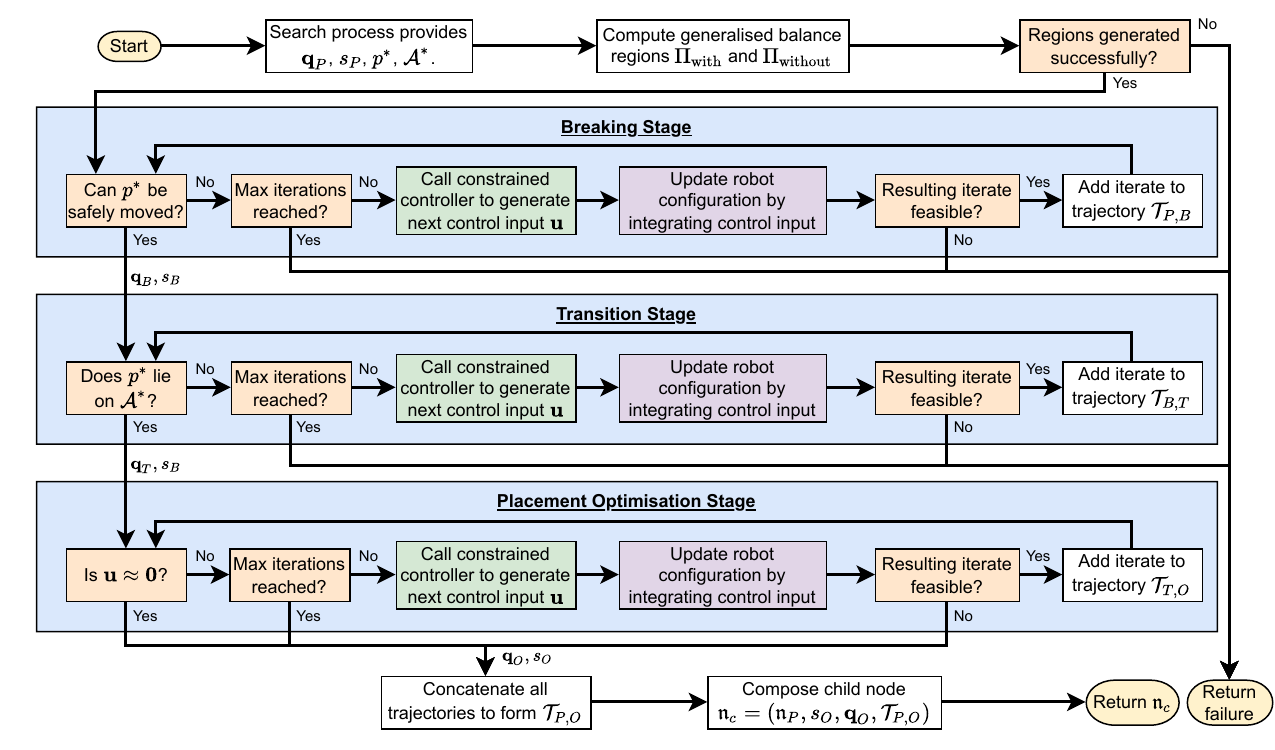}
\caption{Flowchart summarising our proposed posture generator. \label{fig:PG-flowchart}}
\end{figure*}

To guarantee that a given stance is feasible, multi-contact planners
must find safe whole-body transition configurations that allow the
robot to realise the intended stance while respecting certain constraints
(\emph{e.g.}, avoiding collisions, maintaining balance, \emph{etc.}).
The task of finding such a configuration for a given stance is known
as the `\emph{posture generation problem}'. Expanding upon our previous
work \citep{TAROSPaper-2025-Derwent2025Multi}, we propose a novel
posture generation method that improves upon the state-of-the-art
by:
\begin{itemize}
\item Guaranteeing, for every node returned, that a feasible linking trajectory
exists for the robot to safely transition into that node's stance. 
\item Generating the stance, transition configuration and linking trajectory
\emph{simultaneously}, removing the need for secondary planning stages
and hence reducing the overall complexity of the planning pipeline.
\item Returning nodes faster than traditional methods by re-formulating
the problem as a series of linearly constrained quadratic programs,
rather than as a single non-linear optimisation problem.
\end{itemize}
The posture generator is based on a VFI constrained controller in
the same form as \eqref{eq:standard-control-problem}, which drives
a model of the robot into an appropriate stance. This allows us to
simultaneously plan new stances while implicitly generating the transition
configuration and whole-body trajectory, which can be tracked by the
real robot. A flowchart summarising the approach is given in Figure~\ref{fig:PG-flowchart}.

\begin{figure}[t]
\centering{}\includegraphics[width=1\columnwidth]{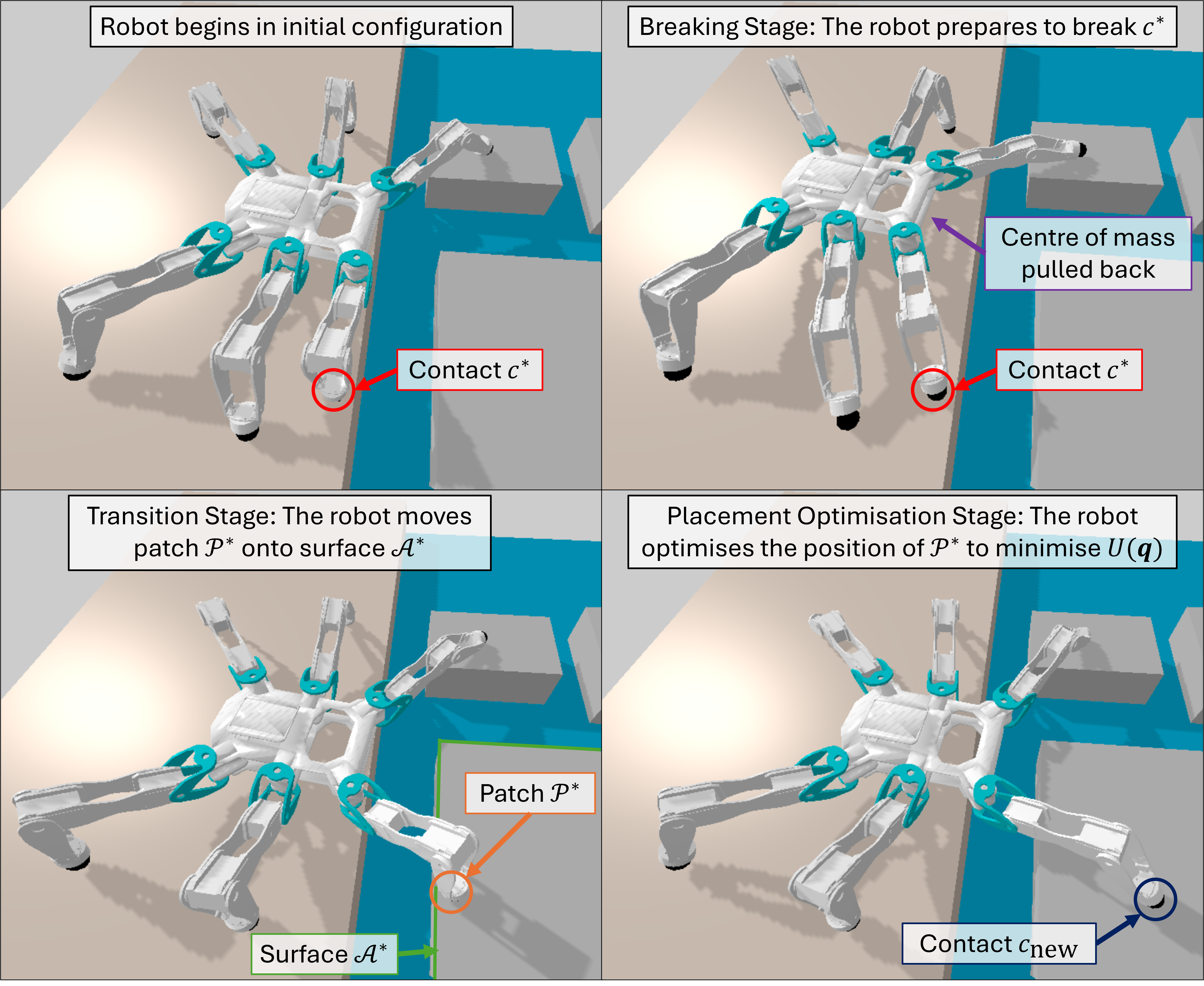}
\caption{Illustration of each posture generation stage. \label{fig:Posture-Gen-Stages}}
\end{figure}

Each posture generator call comprises the following three stages (depicted
in Figure~\ref{fig:Posture-Gen-Stages}):
\begin{itemize}
\item Given the parent configuration $\config_{P}$ and stance $\stance_{P}$,
the breaking stage finds a configuration that allows the robot to
safely break contact $\contact^{*}$, returning the stance $\stance_{B}=\stance_{P}\setminus\left\{ \contact^{*}\right\} $,
configuration $\config_{B}\in\feasiblespace_{\stance_{P}}\cap\feasiblespace_{\stance_{B}}$,
and trajectory $\trajectory PB$ linking $\config_{P}$ to $\config_{B}$. 
\item The transition stage moves the desired contact patch $\patch^{*}$
to the intended contact area $\area^{*}$, returning the configuration
$\config_{T}\in\feasiblespace_{\stance_{B}}$ and trajectory $\trajectory BT$. 
\item The placement optimisation stage optimises the location of $\patch^{*}$
on $\area^{*}$, forming a new contact $\contact_{\text{new}}$, and
returning the stance $\stance_{O}=\stance_{B}\cup\left\{ \contact_{\text{new}}\right\} $,
configuration $\config_{O}\in\feasiblespace_{\stance_{B}}\cap\feasiblespace_{\stance_{O}}$
and trajectory $\trajectory TO$. 
\item The final result is a child node $\node_{c}=\fullnode PO$, where
$\trajectory PO$ is the concatenation of $\trajectory PB$, $\trajectory BT$
and $\trajectory TO$.
\end{itemize}
Each time a quadratic program is solved to find the instantaneous
control input, numerical integration is used to update the robot configuration
for the next time step in the corresponding configuration-space trajectory.
The functions of $\boldsymbol{q}$ (\emph{e.g.}, the Jacobian matrices)
are then recalculated, and the cycle repeats. This continues until
either the stage's stopping criteria are satisfied or an evaluation
limit is reached.

\subsection{VFIs under discretisation effects\label{subsec:VFIs-under-discretisation}}

When operating in continuous time, VFI constraints ensure that any
robot entities outside of a restricted zone at time $t=0$ will continue
to be so for all times $t>0$ . However, in RHCP we use discrete numerical
integration steps to obtain the next configuration every time an instantaneous
control input is returned by the posture generator. In order to minimise
the number of control cycles, and hence also reduce the amount of
time required to plan, it is desirable to use the largest integration
step size possible. Therefore the assumption of continuous time is
violated and, since each integration step acts as a zero-order hold,
the robot may briefly enter the restricted zone, risking unsafe behaviour
or an unfeasible optimisation problem. To address this, we artificially
inflate the boundaries of each restricted zone by a buffer $b_{d}\in\left(0,\infty\right)$,
which can be given as a function of the robot's maximum speed and
the integration period.

\begin{figure}[t]
\centering{}\includegraphics[width=1\columnwidth]{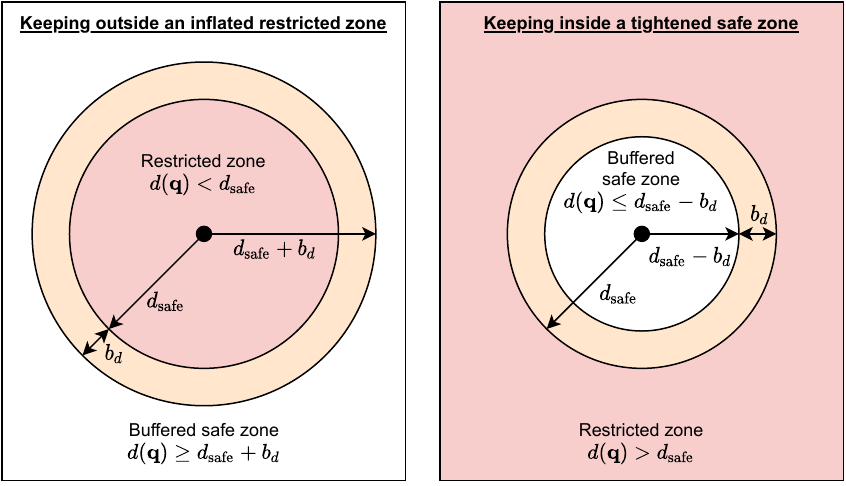}
\caption{Buffer zones applied to VFI constraints when keeping outside an inflated
restricted zone (\emph{left}) \eqref{eq:outside-inflated-boundary}
and inside a tightened safe zone (\emph{right}) \eqref{eq:inside-tightened-constrained}
\label{fig:buffers}}
\end{figure}

When keeping a robot body outside of a restricted zone, the boundary
of that zone is expanded by the buffer (shown on the \emph{left} in
Figure~\ref{fig:buffers}). We thus re-write the constraint \eqref{eq:outside-safe-zone}
for keeping outside a restricted zone as 
\begin{align}
-\mymatrix J_{d}\dot{\myvec q}\leq\eta_{d}\left(\tilde{d}(\myvec q)-b_{d}\right).\label{eq:outside-inflated-boundary}
\end{align}
Likewise, when staying inside a safe zone, the constraint is \emph{tightened}
by the buffer (shown on the \emph{right} in Figure~\ref{fig:buffers}),
and constraint \eqref{eq:inside-safe-zone} is rewritten 
\begin{align}
\mymatrix J_{d}\dot{\myvec q}\leq-\eta_{d}\left(\tilde{d}(\myvec q)+b_{d}\right).\label{eq:inside-tightened-constrained}
\end{align}

\begin{figure}
\centering{}\includegraphics[width=1\columnwidth]{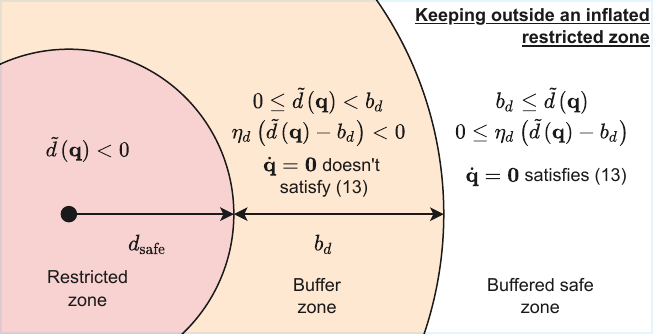}\\
\includegraphics[width=1\columnwidth]{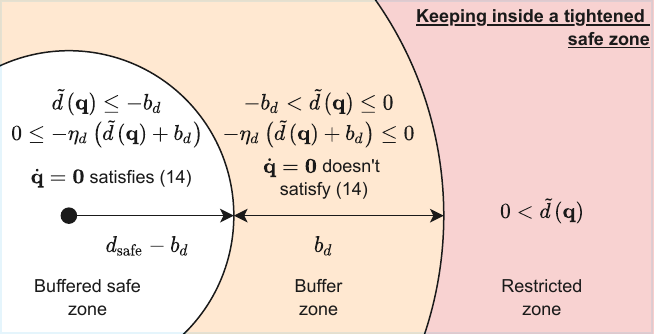}
\caption{Slack variables for keeping outside an inflated restricted zone (\emph{top})
or inside a tightened safe zone (\emph{bottom}). \label{fig:slack}}
\end{figure}

Furthermore, to ensure that there is always at least one solution
to problem \eqref{eq:standard-control-problem}, we improve upon constraints
\eqref{eq:outside-inflated-boundary} and \eqref{eq:inside-tightened-constrained}
in order to guarantee that they are always satisfied by $\dot{\myvec q}=\myvec 0$.
For example, when $\dot{\myvec q}=\myvec 0$, \eqref{eq:outside-inflated-boundary}
is reduced to $0\leq\tilde{d}(\myvec q)-b_{d}$. If the robot violates
the inflated boundaries but does not violate the original boundaries
(\emph{i.e., $0\leq\tilde{d}(\myvec q)<b_{d}$}), we have $\tilde{d}(\myvec q)-b_{d}<0$,
and hence $\dot{\myvec q}=\myvec 0$ no longer satisfies \eqref{eq:outside-inflated-boundary}
(shown in Figure~\ref{fig:slack}). 

To overcome this problem, we add a slack variable $s_{d}\in[0,\textrm{max}(0,-\eta_{d}(\tilde{d}(\myvec q)-b_{d}))]$
to the constraint, resulting in 
\begin{align}
-\mymatrix J_{d}\dot{\myvec q} & \leq\eta_{d}\left(\tilde{d}(\myvec q)-b_{d}\right)+s_{d}.\label{eq:relaxed-ouside-inflated-boundary}
\end{align}
Therefore, if\emph{ $0\leq\tilde{d}(\myvec q)<b_{d}$} and the robot
takes the maximum allowable slack $s_{d}=-\eta_{d}(\tilde{d}(\myvec q)-b_{d})$,
then \eqref{eq:relaxed-ouside-inflated-boundary} becomes $-\mymatrix J_{d}\dot{\myvec q}\leq0$
to which $\dot{\myvec q}=\myvec 0$ is a valid solution. Note that
$-\mymatrix J_{d}\dot{\myvec q}\leq0\iff\mymatrix J_{d}\dot{\myvec q}\geq0\iff\dot{d}\geq0$;
therefore, the robot cannot make any further progress towards the
boundary of the restricted zone. Additionally, if the robot is inside
the buffered safe zone (and hence $\tilde{d}(\myvec q)-b_{d}\geq0$),
then $-\eta_{d}(\tilde{d}(\myvec q)-b_{d}))\leq0$, meaning that the
maximum allowable slack is zero. Hence, only the minimum slack required
for $\dot{\myvec q}=\myvec 0$ to be a valid solution is ever permitted.

Analogously, when inside a safe zone, if the robot body lies inside
the original safe zone but outside of the buffered safe zone (as shown
in Figure~\ref{fig:slack}), we have $-b_{d}<\tilde{d}(\myvec q)\leq0$,
which implies that $\tilde{d}(\myvec q)+b_{d}>0$ and hence $\dot{\myvec q}=\myvec 0$
does not satisfy \eqref{eq:inside-tightened-constrained}. In this
case, the slack variable is written as $s_{d}\in[0,\textrm{max}(0,\eta_{d}(\tilde{d}(\myvec q)+b_{d}))]$
so that \eqref{eq:inside-tightened-constrained} is rewritten as
\begin{equation}
\mymatrix J_{d}\dot{\myvec q}\leq-\eta_{d}\left(\tilde{d}(\myvec q)+b_{d}\right)+s_{d},\label{eq:relaxed-inside-tightened-boundary}
\end{equation}
 making $\dot{\myvec q}=\myvec 0$ a feasible solution. 

Finally, problem \eqref{eq:standard-control-problem} is rewritten
as
\begin{equation}
\argminimfour{\myvec u\in}{\dot{\boldsymbol{q}},\boldsymbol{s}}{\Psi\left(\config,\dot{\myvec q}\right)+\regweighting^{2}\norm{\dot{\myvec q}}_{2}^{2}+\slackweighting^{2}\norm{\myvec s}_{2}^{2}}{\boldsymbol{W}_{1}\dot{\boldsymbol{q}}\preceq\boldsymbol{w}_{1}+\myvec s}{\boldsymbol{W}_{2}\dot{\boldsymbol{q}}=\boldsymbol{w}_{2}}{\qvmin\preceq\dot{\myvec q}\preceq\qvmax}{\veczero_{\numslackvariables}\preceq\boldsymbol{s}\preceq\boldsymbol{s}_{\max},}\label{eq:optimisation-problem-with-slack}
\end{equation}
where $\slackweighting,\regweighting\in\left(0,\infty\right)$ and
$\myvec s=\begin{bmatrix}s_{1} & \cdots & s_{\numslackvariables}\end{bmatrix}^{T}$
is the vector of slack variables. The vector pairs $\qvmin,\qvmax$
and $\veczero_{\numslackvariables},\boldsymbol{s}_{\max}$ impose
lower and upper limits on $\dot{\boldsymbol{q}}$ and $\boldsymbol{s}$,
respectively. The task function $\Psi\left(\config,\dot{\myvec q}\right)$
defines the task for a given stage (breaking, transition, placement
optimisation) in the form 
\begin{equation}
\Psi\left(\dot{\myvec q}\right)=\sum_{i=1}^{\numtasks}\norm{\mymatrix J_{\tilde{\myvec x}_{i}}\dot{\myvec q}+\eta_{o}\tilde{\myvec x}_{i}}_{2}^{2}.\label{eq:form-of-task-function}
\end{equation}

\subsection{Constraints and Objectives For the Posture Generator\label{subsec:Constraints-and-Objectives}}

In this section the VFI constraints and objective functions $\Psi\left(\dot{\myvec q}\right)$
for each stage of the posture generator are derived.

\subsubsection{Common Constraints to All Stages\label{subsec:Common-Constraints}}

Five types of constraints apply to all stages of the posture generator:
avoiding collisions and self-collisions; maintaining balance; preventing
contacts from sliding; avoiding excessive torso tilting; and respecting
the limits and topologies of the optimisation variables.

\paragraph{Collision and self-collision avoidance constraints}

\begin{figure}[!t]
\centering{}\includegraphics[width=1\columnwidth]{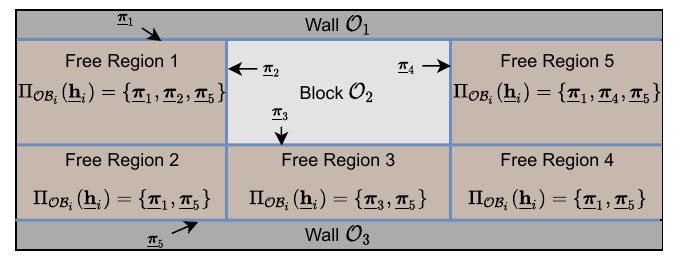}
\caption{Illustration of how collision planes are assigned for a given robot
body part based on its position in a non-convex free space. The collision
plane with the ground is omitted for clarity. \label{fig:non-convex-free-spaces}}
\end{figure}

We define $\collisionbodies\subset\mathcal{H}$ as the set of dual
quaternion primitives representing each robot body part $\pose_{\indexrobot}\in\collisionbodies$
for the purposes of collisions, which may be planes, cylinders or
spheres \citep{DQ-Marinho2019VFIs}. The environment is modelled as
a set of convex obstacles, whose surfaces are described by the set
$\allcollisionsurfaces\triangleq\left\{ \dq{\pi}_{1},\ldots,\dq{\pi}_{\numobstacles}\right\} $,\LyXZeroWidthSpace{}
wherein each element $\dq{\pi}_{\indexenv}\isplane$ is an infinite
plane whose normal vector is directed outside of the obstacle. The
subset of planes that are forbidden from colliding with the robot
body part $\pose_{\indexrobot}$ is written as $\bodycollisionsurfaces i\left(\pose_{i}\right)\subseteq\allcollisionsurfaces$.
The planes in $\bodycollisionsurfaces i$ are selected based on the
position of the robot primitive $\pose_{i}$ in the workspace, similar
to the approach used to model small obstacles in \citep{DQ-Pereira2022CollisionAvoidanceAndCones}.
An illustration of this process is shown in Figure~\ref{fig:non-convex-free-spaces}.
For instance, when a robot body part is in the free region 3, planes
$\plane_{3}$ and $\plane_{5}$ prevent the robot part from colliding
with block $\obstacle_{2}$ and wall $\obstacle_{3}$ and allow the
robot body part to be moved to free region 2 or free region 4. If
the robot part moves to free region 2, then only planes $\plane_{1}$
and $\plane_{5}$ are activated, allowing the robot part to move to
either free region 1 or free region 3. Although this process is simplified
in Figure~\ref{fig:non-convex-free-spaces}, it its general and we
use it to represent non-convex tridimensional free workspaces.

We thus write the signed distance between a given robot body primitive
$\pose_{\indexrobot}\in\collisionbodies$ and a given plane $\plane_{\indexenv}\in\bodycollisionsurfaces i$
as $\signdistance{\pose_{\indexrobot}}{\plane_{\indexenv}}\isreal$.
The Jacobian matrix of $\signdistance{\pose_{\indexrobot}}{\plane_{\indexenv}}$
is written as $\mymatrix J_{\pose_{i},\dq{\pi}_{\indexenv}}\triangleq\mymatrix J_{\pose_{i},\dq{\pi}_{\indexenv}}\left(\config\right)$
\citep{DQ-Marinho2019VFIs}. Hence, the unbuffered distance term is
written as $\signederror{\pose_{i}}{\plane_{j}}\triangleq\signdistance{\pose_{i}}{\plane_{j}}-\safedistance{\obstacle}$
and the collision avoidance requirement is stated as $\signederror{\pose_{i}}{\plane_{j}}\geq b_{d}$.
The collision avoidance constraint is thus defined, in the same form
as \eqref{eq:relaxed-ouside-inflated-boundary}, 
\begin{multline}
-\mymatrix J_{\pose_{i},\dq{\pi}_{\indexenv}}\dot{\myvec q}\leq\eta_{d}\left(\tilde{d}_{\pose_{i},\dq{\pi}_{\indexenv}}-b_{d}\right)+s_{\pose_{i},\dq{\pi}_{\indexenv}},\\
\forall\dq{\pi}_{\indexenv}\in\bodycollisionsurfaces{.i},\:\forall\pose_{i}\in\collisionbodies.\label{eq:object-collision}
\end{multline}

These constraints can be represented in stacked matrix/vector form
as 
\begin{equation}
\begin{aligned}\mymatrix W_{\obstacle\robotEXCEPTION}= & \left[\begin{array}{ccc}
-\jac_{1}^{\transpose} & \ldots & -\jac_{n_{\obstacle\mathcal{B}}}^{\transpose}\end{array}\right]^{\transpose}\\
\myvec w_{\obstacle\robotEXCEPTION}= & \left[\begin{array}{ccc}
\eta_{d}\left(\tilde{d}_{1}-b_{d}\right) & \ldots & \eta_{d}\left(\tilde{d}_{n_{\obstacle\mathcal{B}}}-b_{d}\right)\end{array}\right]^{\transpose}\\
\myvec s_{\obstacle\robotEXCEPTION}= & \left[\begin{array}{ccc}
s_{1} & \ldots & s_{n_{\obstacle\mathcal{B}}}\end{array}\right]^{\transpose},
\end{aligned}
\label{eq:stackng-units}
\end{equation}
where $n_{\obstacle\mathcal{B}}$ is the number of robot body parts
in $\collisionbodies$ multiplied by the number of planes in each
corresponding set $\bodycollisionsurfaces i$. Thus, the complete
collision avoidance constraint may be evaluated as
\begin{equation}
\mymatrix W_{\obstacle\robotEXCEPTION}\dot{\config}\preceq\myvec w_{\obstacle\robotEXCEPTION}+\myvec s_{\obstacle\robotEXCEPTION}.\label{eq:stacked-obj-collision}
\end{equation}

The process for self-collision avoidance is similar. We denote by
$\selfcollisionbodies(\pose)\subset\collisionbodies$ the subset of
primitives in $\collisionbodies$ that are forbidden from colliding
with $\pose\in\mathcal{\collisionbodies}$. The self-collision constraint
is thus written as
\begin{equation}
-\mymatrix J_{\pose,\dq y}\dot{\myvec q}\leq\eta_{d}\left(\tilde{d}_{\pose,\dq y}-b_{d}\right)+s_{\pose,\dq y},\:\forall\dq y\in\selfcollisionbodies(\pose),\:\forall\pose\in\collisionbodies.\label{eq:self-collision}
\end{equation}
 These constraints may be expressed in stacked matrix/vector form
in the same manner as \eqref{eq:stackng-units}, being written as 

\begin{equation}
\mymatrix W_{\robotEXCEPTION\robotEXCEPTION}\dot{\config}\preceq\myvec w_{\robotEXCEPTION\robotEXCEPTION}+\myvec s_{\robotEXCEPTION\robotEXCEPTION}.\label{eq:stacked-self-collision}
\end{equation}

\paragraph{Balance constraint}

In order to maintain quasi-static balance during the posture generation
process, the robot must position its COM in such a way that feasible
forces can be generated at the contacts that will counteract gravity,
result in no net force or moment, and prevent contacts from slipping.
Bretl and Lall \citep{RHCP-Bretl2008SupportRegions} show that, for
any set of point contacts, the region of COM positions satisfying
these conditions is a vertical prism with a convex cross-section on
the horizontal plane. They also present an efficient interior-point
algorithm to compute a conservative approximation of this region (\emph{the
generalised support region}) to within a user-defined level of precision.
We adopt this method for our posture generation process with two small
modifications. 

\begin{figure}[t]
\centering{}\includegraphics[width=1\columnwidth]{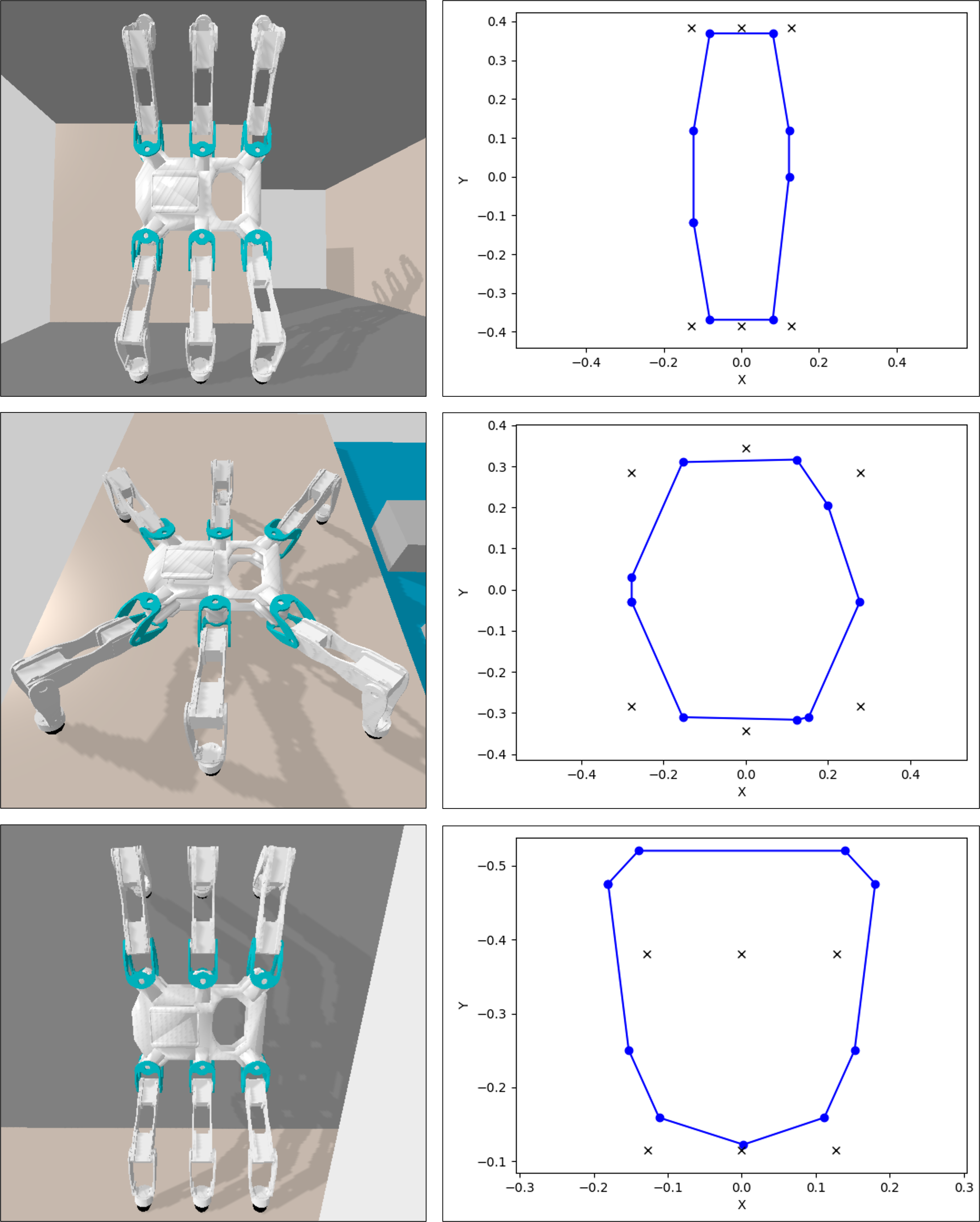}
\caption{The Corin hexapod in different configurations (\emph{left}) and their
associated generalised support regions (\emph{right}). In each figure,
the positions of the contacts are shown as black crosses. \label{fig:Balance-regions}}
\end{figure}

Firstly, we apply limits on the magnitude of the forces that the robot
is permitted to generate at each contact, constraining each 3D force
vector $\force_{i}$ such that $\force_{\min}\preceq\force_{i}\preceq\force_{\max}$.
Secondly, we use polyhedral approximations of the Coulomb friction
cones \citep{General-SpringerHandbook-ContactModellingChapter}, reducing
the algorithm in \citep{RHCP-Bretl2008SupportRegions} to a series
of linear optimisation problems with linear constraints, allowing
it to be solved extremely quickly. Examples of generalised support
regions for three different scenarios are shown in Figure~\ref{fig:Balance-regions}.
\footnote{The support regions shown in Figure~\ref{fig:Balance-regions} do
not meet the contact locations, even in the middle case where they
all lie on a horizontal surface, because the forces realised at each
contact are limited. When the COM is very close to a single contact,
the forces required at that contact to support the robot's balance
become large enough to exceed this upper limit, and hence the support
regions are cut-off before reaching the contact points.}

\begin{figure}
\noindent\begin{minipage}[t]{1\columnwidth}%
\begin{center}
\mbox{%
\includegraphics[width=1\textwidth]{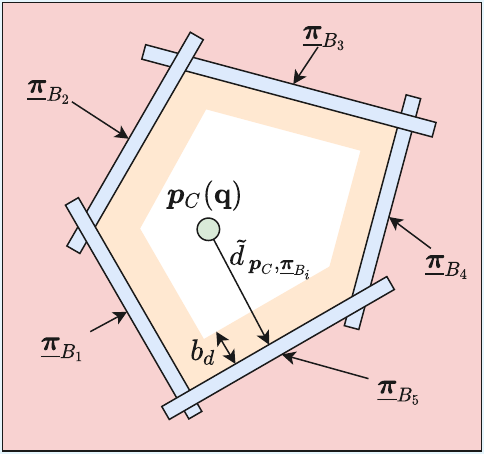}%
} \caption{Illustration of the quasi-static balance VFI constraint \eqref{eq:balance}
for a 2D cross section, with $\protect\plane_{B_{1}},\ldots,\protect\plane_{B_{5}}\in\Pi_{\textrm{balance}}$.\label{fig:VFI-Balance}}
\par\end{center}%
\end{minipage}
\end{figure}

Having computed the generalised support region, we define $\Pi_{\textrm{balance}}\triangleq\{\dq{\pi}_{B_{1}},\ldots,\dq{\pi}_{B_{m}}\}$
as the set of vertical planes describing the boundaries of the region.
The normal vector of each plane is directed outside of the region.
Writing the point-to-plane signed distance between the centre of mass
position $\COM\left(\config\right)\in\workspace$ and the plane $\plane_{B_{\indexrobot}}\in\Pi_{\text{balance}}$
as $\signdistance{\COM}{\plane_{B_{i}}}\isreal$, and the Jacobian
matrix of $\signdistance{\COM}{\plane_{B_{i}}}$ as $\mymatrix J_{\COM,\dq{\pi}_{B_{\indexrobot}}}$
\citep{DQ-Marinho2019VFIs}, the unbuffered distance term is given
as $\signederror{\COM}{\plane_{B_{\indexrobot}}}\triangleq\signdistance{\COM}{\plane_{i}}-\safedistance B$
and the requirement is stated as $\signederror{\COM}{\plane_{B_{\indexrobot}}}\leq-b_{d}$,
illustrated in Figure~\ref{fig:VFI-Balance}. 

The static balance constraint is thus defined, in the same form as
\eqref{eq:relaxed-inside-tightened-boundary}, 
\begin{equation}
\mymatrix J_{\quat p_{C},\dq{\pi}_{B_{i}}}\dot{\myvec q}\!\leq\!-\eta_{d}\left(\tilde{d}_{\quat p_{C},\dq{\pi}_{B_{i}}}\!+\!b_{d}\right)+s_{\quat p_{C},\dq{\pi}_{B_{i}}}\!,\forall\dq{\pi}_{B_{i}}\!\in\!\Pi_{\textrm{balance}}.\label{eq:balance}
\end{equation}
As before, these constraints are expressed in stacked matrix/vector
form as 
\begin{equation}
\mymatrix W_{\text{Bal}}\dot{\config}\preceq\myvec w_{\text{Bal}}+\myvec s_{\text{Bal}}.\label{eq:stacked-balance}
\end{equation}

\paragraph{Preventing foot sliding}

When the posture generator is called, it first determines which contact
patches are free to move during a particular stage. For instance,
during the breaking stage no contact patches are free to move, while
in the transition and placement optimisation stages only $\patch^{*}$
is free to move. Those contact patches that are \emph{not} free to
move make up the set $\allcontactpatches_{\text{fixed}}\subset\allcontactpatches$,
and the posture generator must ensure that their positions do not
change.

\begin{figure}
\begin{minipage}[t]{0.49\columnwidth}%
\mbox{%
\includegraphics[width=1\textwidth]{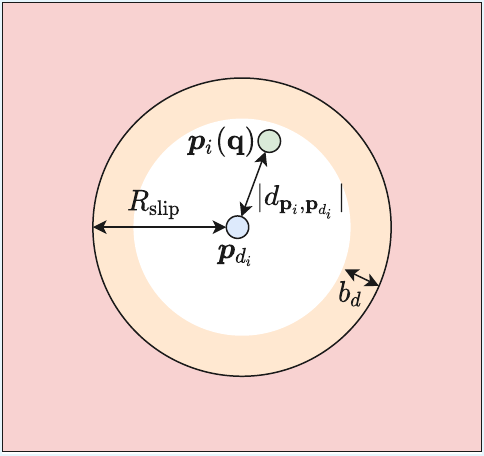}%
} \caption{Illustration of the contact slippage VFI constraint \eqref{eq:contact-slippage}
for a 2D cross section. Note that $\protect\squaredistance{\protect\point_{\protect\indexrobot}}{\protect\point_{d_{\protect\indexrobot}}}=\protect\signdistance[2]{\protect\point_{i}}{\protect\point_{d_{i}}}$.\label{fig:VFI-Slippage}}
\end{minipage}\hfill{}%
\begin{minipage}[t]{0.49\columnwidth}%
\mbox{%
\includegraphics[width=1\textwidth]{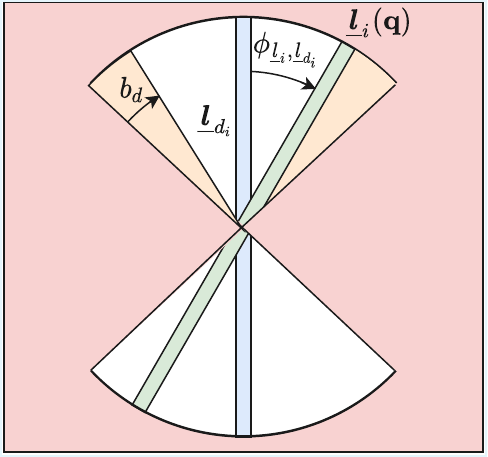}%
} \caption{Illustration of the angular VFI constraint \eqref{eq:orientation}for
a 2D cross section.\label{fig:VFI-Orientation}}
\end{minipage}
\end{figure}

Therefore, we constrain the centroid of each fixed contact patch,
written as $\point_{\indexrobot}$ for $\patch_{i}\in\allcontactpatches_{\text{fixed}}$,
to remain inside a small sphere of radius $R_{\text{slip}}$ centred
on its desired location $\point_{d_{\indexrobot}}$. Writing the point-to-point
square distance between $\point_{\indexrobot}$ and $\point_{d_{\indexrobot}}$
as $\squaredistance{\point_{\indexrobot}}{\point_{d_{\indexrobot}}}$,
we define the unbuffered distance term as $\tilde{D}_{\quat p_{i},\quat p_{d_{i}}}\triangleq\squaredistance{\point_{\indexrobot}}{\point_{d_{\indexrobot}}}-R_{\text{slip}}^{2}$
and state the requirement as $\tilde{D}_{\quat p_{i},\quat p_{d_{i}}}\leq-b_{d}^{2}$
(illustrated in Figure~\ref{fig:VFI-Slippage}). The constraint is
thus given as 
\begin{equation}
\mymatrix J_{\quat p_{i},\quat p_{d_{i}}}\dot{\myvec q}\leq-\eta_{d}\left(\tilde{D}_{\quat p_{i},\quat p_{d_{i}}}+b_{d}^{2}\right)+s_{\quat p_{i},\quat p_{d_{i}}},\:\forall\patch_{i}\in\allcontactpatches_{\text{fixed}},\label{eq:contact-slippage}
\end{equation}
and is expressed in stacked matrix/vector form as before 
\begin{equation}
\mymatrix W_{\allcontactpatches}\dot{\config}\preceq\myvec w_{\text{\ensuremath{\allcontactpatches}}}+\myvec s_{\text{\ensuremath{\allcontactpatches}}}.\label{eq:stacked-contact-slippage}
\end{equation}

\paragraph{Preventing excessive tilting:}

In order to prevent the robot from assuming extreme root (\emph{i.e.},
the robot base/torso) orientations that are not conducive to continued
progress, we constrain the root orientation to lie within a maximum
tolerance from a desired value.

The desired root orientation is defined as that of the closest point
on the whole-body guide path $\configguidepath$ to the current configuration.
We then obtain Plücker lines in dual quaternion form \citep{DQ-Adorno2017Fundamentals}
describing the $x$, $y$ and $z$ axes of the robot's torso frame
with respect to the world frame ($\dq l_{x},\dq l_{y},\dq l_{z}\in\mathcal{H}_{p}\cap\dq{\mathcal{S}}$,
respectively) as well as those for the desired torso orientation (denoted
$\dq l_{d_{x}},\dq l_{d_{y}},\dq l_{d_{z}}\in\mathcal{H}_{p}\cap\dq{\mathcal{S}}$).
Using the angular distance function between two lines $d_{\phi_{\dq l_{1}},\phi_{\dq l_{2}}}$
and its Jacobian matrix $\mymatrix J_{\phi_{\dq l_{1}},\phi_{\dq l_{2}}}$
as described in \citep{DQ-Quiroz-Omana2019AngularConstraint}, we
define the error term as $\tilde{d}_{\phi_{\dq l_{1},}\phi_{\dq l_{2}}}\triangleq d_{\phi_{\dq l_{1}},\phi_{\dq l_{2}}}-\safedistance{\phi}$
and state the requirement as $\tilde{d}_{\phi_{\dq l_{1}},\phi_{\dq l_{2}}}\leq-b_{d}$
(illustrated in Figure~\ref{fig:VFI-Orientation}). The constraint
is thus given as 
\begin{equation}
\mymatrix J_{\phi_{\dq l_{i}},\phi_{\dq l_{d_{i}}}}\dot{\myvec q}\!\leq\!-\eta_{d}\left(\tilde{d}_{\phi_{\dq l_{i}},\phi_{\dq l_{d_{i}}}}\!+\!b_{d}\right)+s_{\phi_{\dq l_{i}},\phi_{\dq l_{d_{i}}}},\forall\dq l_{i}\!\in\!\{\dq l_{x},\dq l_{y},\dq l_{z}\}.\label{eq:orientation}
\end{equation}
As before, this is written in matrix/vector form as
\begin{equation}
\mymatrix W_{\phi}\dot{\config}\preceq\myvec w_{\text{\ensuremath{\phi}}}+\myvec s_{\phi}.\label{eq:stacked-orientation}
\end{equation}

\paragraph{Respecting variable limits and topologies:}

As discussed in Section~\ref{sec:Mathematical-Preliminaries}, this
work assumes that configuration vectors take the form $\config=\left[\begin{array}{ccc}
\vecfour{\rot}^{\transpose} & \vecthree{\point}^{\transpose} & \jointangles^{\transpose}\end{array}\right]^{\transpose}$, where the term $\quat r\in\mathbb{S}^{3}$ represents the orientation
of the robot root. Due to the presence of this term, the optimisation
must be constrained to ensure that $\dot{\myvec q}$ respects the
properties of the underlying topology of unit quaternions\emph{—i.e.,}
the condition $\norm{\quat r}=\vecfour{\quat r}^{T}\vecfour{\quat r}=1$
must be maintained. By taking the time derivative of this expression,
we obtain the constraint
\begin{equation}
\vecfour{\rot}^{T}\vecfour{\dot{\quat r}}=0,\label{eq:unit-quat-constraint}
\end{equation}
which can be expressed in terms of $\dot{\config}$ by defining 
\[
\mymatrix R=\left[\begin{array}{cc}
\vecfour{\rot}^{T} & \mymatrix 0_{3+\numtheta}^{\transpose}\end{array}\right],
\]
and hence writing $\mymatrix R\dot{\config}=0.$

Additionally, given limits on $\myvec q$ in the form $\myvec q_{\text{min}},\myvec q_{\text{max}}\isvector n$,
we constrain $\dot{\myvec q}$ such that $\config_{\min}\preceq\myvec q\preceq\config_{\max}$
(similar to \citep{DQ-Quiroz-Omana2019AngularConstraint}) by writing
the constraint
\begin{equation}
\eta_{d}\left(\myvec q_{\text{min}}-\myvec q+b_{d}\right)\preceq\dot{\myvec q}\preceq\eta_{d}\left(\myvec q_{\text{max}}-\myvec q-b_{d}\right).\label{eq:config-limits}
\end{equation}
We define 
\begin{align*}
\myvec w_{\min} & =\eta_{d}\left(\myvec q_{\text{min}}-\myvec q+b_{d}\right),\\
\myvec w_{\max} & =\eta_{d}\left(\myvec q_{\text{max}}-\myvec q-b_{d}\right),
\end{align*}
and thus \eqref{eq:config-limits} is rewritten as $\myvec w_{\min}\preceq\dot{\myvec q}\preceq\myvec w_{\max}$.

\subsubsection{Breaking Stage\label{subsec:Breaking-Stage}}

The breaking stage brings the robot's centre of mass into a position
such that contact $\contact^{*}$ is not required for the robot's
balance and can thus be safely broken. To begin, the posture generator
computes two generalised support regions \citep{RHCP-Bretl2008SupportRegions},
one \emph{with} and one \emph{without} contact $\contact^{*}$, whose
respective sets of planes are denoted $\Pi_{\textrm{with}}$ and $\Pi_{\textrm{without}}$.
We can thus formalise the breaking stage's requirement as 
\begin{equation}
d_{\COM,\dq{\pi}_{B_{\indexrobot}}}\leq-b_{d},\,\forall\dq{\pi}_{B_{i}}\in\Pi_{\textrm{without}},\label{eq:breaking-goal}
\end{equation}
where $\signdistance{\COM}{\plane_{B_{i}}}$ denotes the point-to-plane
signed distance between the robot's centre of mass $\COM$ and each
plane $\plane_{B_{i}}\in\Pi_{\text{without}}$ \citep{DQ-Marinho2019VFIs}. 

\begin{figure}
\centering{}\includegraphics[width=1\columnwidth]{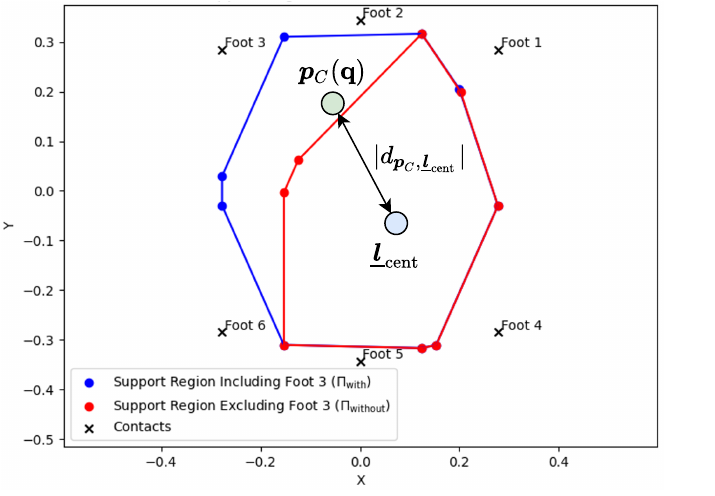}
\caption{Illustration of the breaking stage task function \eqref{eq:breaking-task-func}.
The goal is to move $\protect\COM$ toward $\protect\dq l_{\textrm{cent}}$
so that the contact, in this case foot 3, can be safely broken. \label{fig:Task-Func-Breaking}}
\end{figure}

If condition \eqref{eq:breaking-goal} is satisfied by the starting
configuration, then the breaking stage is skipped, with the robot
proceeding to the transition stage. If the breaking stage is not skipped,
then we define $\dq l_{\textrm{cent}}\in\mathcal{H}_{p}\cap\dq{\mathcal{S}}$
as the line parallel to the gravity vector that intersects the centroid
of the region $\Pi_{\textrm{without}}$. Writing the point-to-line
square distance between $\COM$ and $\dq l_{\textrm{cent}}$ as $\squaredistance{\COM}{\dq l_{\textrm{cent}}}\isnonnegative$,
we define the task function as
\begin{equation}
\Psi_{B}\left(\dot{\myvec q}\right)=\norm{\mymatrix J_{\quat p_{C},\dq l_{\text{cent}}}\dot{\myvec q}+\eta_{o}D_{\quat p_{C},\dq l_{\text{cent}}}}_{2}^{2},\label{eq:breaking-task-func}
\end{equation}
where $\mymatrix J_{\quat p_{C},\dq l_{\text{cent}}}$ denotes the
Jacobian matrix of $\squaredistance{\COM}{\dq l_{\textrm{cent}}}$.
Minimising $\Psi_{B}\left(\dot{\myvec q}\right)$ drives the robot's
COM from $\Pi_{\text{with}}$ into $\Pi_{\text{without}}$ (illustrated
in Figure~\ref{fig:Task-Func-Breaking}).

The constraints imposed during the breaking stage include avoiding
collisions \eqref{eq:object-collision} and self-collisions \eqref{eq:self-collision},
maintaining quasi-static balance (by keeping $\COM$ inside the safe
zone described by $\Pi_{\text{with}}$) \eqref{eq:balance}, maintaining
all existing contacts (\emph{i.e.}, $\allcontactpatches_{\text{fixed}}=\allcontactpatches$)
\eqref{eq:contact-slippage}, avoiding excessive root rotations \eqref{eq:orientation},
and respecting the variable bounds and topologies \eqref{eq:unit-quat-constraint},
\eqref{eq:config-limits}. The optimisation problem is hence written
as
\begin{equation}
\argminimfive{\continput\in}{\dot{\boldsymbol{q}},\boldsymbol{s}}{\Psi_{B}+\regweighting^{2}\norm{\dot{\myvec q}}_{2}^{2}+\slackweighting^{2}\norm{\myvec s}_{2}^{2}}{\boldsymbol{W}_{B}\dot{\boldsymbol{q}}\preceq\boldsymbol{w}_{B}+\myvec s_{B}}{\myvec R\dot{\boldsymbol{q}}=0}{\myvec w_{\min}\preceq\dot{\myvec q}\preceq\myvec w_{\max}}{\qvmin\preceq\dot{\myvec q}\preceq\qvmax}{\veczero_{\numslackvariables}\preceq\boldsymbol{s}\preceq\boldsymbol{s}_{\max},}\label{eq:breaking-stage-optimisation}
\end{equation}
where 
\begin{equation}
\mymatrix W_{B}=\left[\begin{array}{c}
\mymatrix W_{\obstacle\robotEXCEPTION}\\
\mymatrix W_{\robotEXCEPTION\robotEXCEPTION}\\
\mymatrix W_{\text{Bal}}\\
\mymatrix W_{\allcontactpatches}\\
\mymatrix W_{\phi}
\end{array}\right],\;\myvec w_{B}=\left[\begin{array}{c}
\myvec w_{\obstacle\robotEXCEPTION}\\
\myvec w_{\robotEXCEPTION\robotEXCEPTION}\\
\myvec w_{\text{Bal}}\\
\myvec w_{\allcontactpatches}\\
\myvec w_{\phi}
\end{array}\right],\;\myvec s_{B}=\left[\begin{array}{c}
\myvec s_{\obstacle\robotEXCEPTION}\\
\myvec s_{\robotEXCEPTION\robotEXCEPTION}\\
\myvec s_{\text{Bal}}\\
\myvec s_{\allcontactpatches}\\
\myvec s_{\phi}
\end{array}\right].\label{eq:constraints-breaking-stage}
\end{equation}
The control input $\myvec u$ is numerically integrated to update
the robot configuration. If the resulting configuration satisfies
condition \eqref{eq:breaking-goal}, then the breaking stage terminates
and the process proceeds to the transition stage.

If condition \eqref{eq:breaking-goal} is \emph{not} satisfied, then
the process repeats, generating a new control input. If a user-defined
maximum number of integration steps is reached without satisfying
condition \eqref{eq:breaking-goal}, then the posture generator returns
a failure state.

\subsubsection{Transition Stage\label{subsec:Transition-Stage}}

The transition stage brings patch $\patch^{*}$ into contact with
the area $\area^{*}$, modelled as a plane $\dq{\pi}_{\area^{*}}\in\dq{\mathcal{S}}$
with normal pointing outwards from the perspective of the object on
which it is attached, with boundaries described by the set of planes
$\Pi_{\textrm{bound}}$ with normal pointing outwards the bounded
area. Writing the position of $\patch^{*}$ as $\point^{*}\in\workspace$,
the transition stage is satisfied if
\begin{align}
d_{\point^{*},\dq{\pi}_{i}}\leq & -b_{d},\,\forall\dq{\pi}_{i}\in\Pi_{\textrm{bound}}\label{eq:boundary-condition}\\
D_{\point^{*},\dq{\pi}_{\area^{*}}}\leq & b_{d}^{2}\label{eq:plane-condition}
\end{align}
where $d_{\point^{*},\dq{\pi}_{i}}$ denotes the point-to-plane signed
distance between $\point^{*}$ and $\dq{\pi}_{i}$, and $D_{\point^{*},\dq{\pi}_{\area^{*}}}$
denotes the point-to-plane square distance between $\point^{*}$ and
$\dq{\pi}_{\area^{*}}$ \citep{DQ-Marinho2019VFIs}. If conditions
\eqref{eq:boundary-condition} and \eqref{eq:plane-condition} are
both already satisfied, then the transition stage is skipped.

\begin{figure}
\begin{minipage}[t]{0.49\columnwidth}%
\mbox{%
\includegraphics[width=1\textwidth]{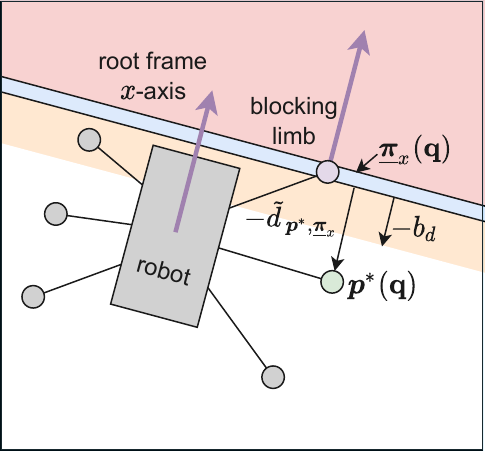}%
} \caption{Illustration of the limb-crossing VFI constraint \eqref{eq:criss-cross-constraint-1}.\label{fig:VFI-Criss-cross}}
\end{minipage}\hfill{}%
\begin{minipage}[t]{0.49\columnwidth}%
\mbox{%
\includegraphics[width=1\textwidth]{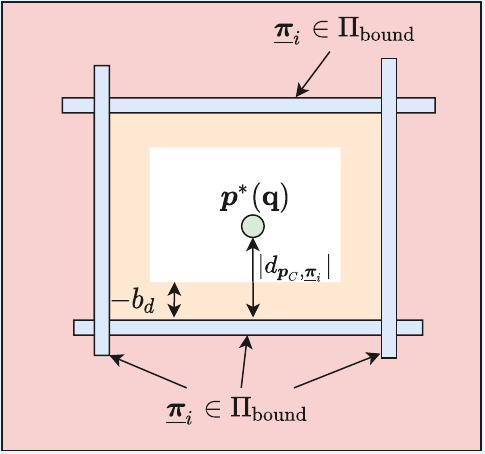}%
} \caption{Illustration of the VFI constraint for respecting the bounds of $\protect\area^{*}$
\eqref{eq:respect-all-boundary-planes}.\label{fig:VFI-boundaries}}
\end{minipage}
\end{figure}

Additionally, $\patch^{*}$ is constrained to prevent it from crossing
in-front of any preceding limbs. For example, the middle-right foot
of the hexapod shown in Figure~\ref{fig:VFI-Criss-cross} is prevented
from crossing in-front of the front right foot (designated the \emph{blocking
limb}). Let $\dq{\pi}_{x}\isplane$ denote a plane intersecting the
blocking limb's end effector with a normal vector parallel to the
$x$-axis of the robot's root frame, which points forward with respect
to the body. The signed point-to-plane distance between $\point^{*}$
and $\dq{\pi}_{x}$ is written $d_{\point^{*},\dq{\pi}_{x}}$, and
the associated error term as $\tilde{d}_{\point^{*},\dq{\pi}_{x}}\triangleq d_{\point^{*},\dq{\pi}_{x}}-\safedistance x$
with Jacobian matrix $\mymatrix J_{\point^{*},\plane_{x}}$ \citep{DQ-Marinho2019VFIs}.
The task-space requirement is thus given as $\tilde{d}_{\point^{*},\dq{\pi}_{x}}\leq-b_{d}$
and hence we define the VFI constraint 
\begin{equation}
\mymatrix J_{\point^{*},\plane_{x}}\dot{\boldsymbol{q}}\leq-\eta_{d}\left(\tilde{d}_{\point^{*},\dq{\pi}_{x}}+b_{d}\right)+s_{\point^{*}\dq{\pi}_{x}},\label{eq:criss-cross-constraint-1}
\end{equation}
 written in matrix/vector form as 
\begin{equation}
\mymatrix W_{x}\dot{\config}\preceq\myvec w_{\text{\ensuremath{x}}}+\myvec s_{\text{\ensuremath{x}}}.\label{eq:stacked-criss-cross}
\end{equation}

The remaining formulations are dependent on condition \eqref{eq:boundary-condition}.
If condition \eqref{eq:boundary-condition} \emph{is} \emph{satisfied},
then we apply the constraint 
\begin{equation}
\mymatrix J_{\point^{*},\dq{\pi}_{i}}\dot{\boldsymbol{q}}\leq-\eta_{d}\left(\tilde{d}_{\point^{*},\dq{\pi}_{i}}+b_{d}\right)+s_{\point^{*},\dq{\pi}_{i}},\,\forall\dq{\pi}_{i}\in\Pi_{\textrm{bound}},\label{eq:respect-all-boundary-planes}
\end{equation}
which constrains $\point^{*}$ to remain inside the boundaries of
$\area^{*}$ (illustrated in Figure~\ref{fig:VFI-boundaries}). This
is written in matrix/vector form as 
\begin{equation}
\mymatrix W_{\text{Bnd}}\dot{\config}\preceq\myvec w_{\text{Bnd}}+\myvec s_{\text{\text{Bnd}}}.\label{eq:stacked-boundaries}
\end{equation}

\begin{figure}
\begin{minipage}[t]{0.49\columnwidth}%
\mbox{%
\includegraphics[width=1\textwidth]{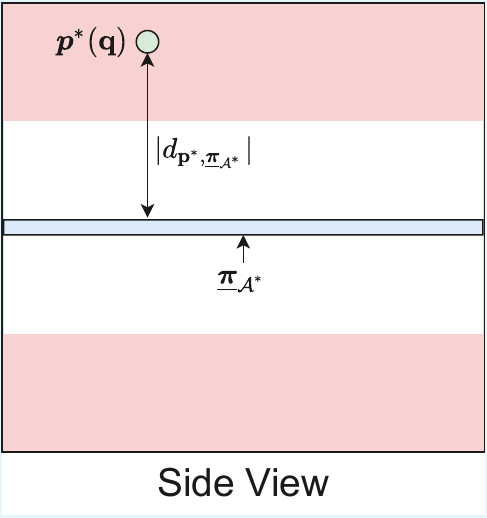}%
} \caption{Illustration of transition stage task function $\Psi_{T_{1}}$, given
in \eqref{eq:transition-task-func-1}, to drive $\protect\point^{*}\left(\protect\config\right)$
to the $\protect\plane_{\mathcal{A}^{*}}$. \label{fig:Task-Func-Transition-1}}
\end{minipage}\hfill{}%
\begin{minipage}[t]{0.49\columnwidth}%
\mbox{%
\includegraphics[width=1\textwidth]{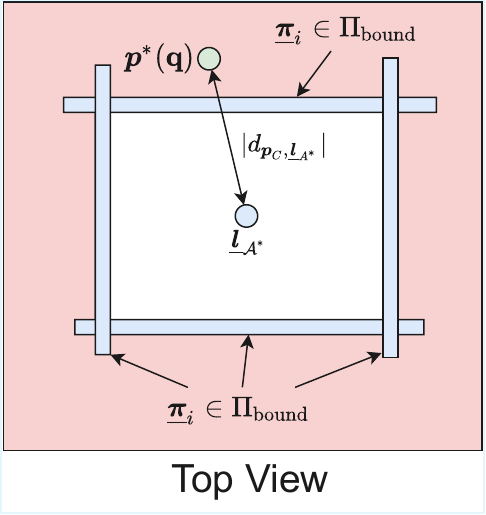}%
} \caption{Illustration of transition stage task function $\Psi_{T_{2}}$, given
in \eqref{eq:transition-task-func-2}..\label{fig:Task-Func-Transition-2}}
\end{minipage}
\end{figure}

The task function is thus given as
\begin{equation}
\Psi_{T_{1}}\left(\dot{\myvec q}\right)\triangleq\norm{\mymatrix J_{\point^{*},\dq{\pi}_{\area^{*}}}\dot{\myvec q}+\eta_{o}D_{\point^{*},\dq{\pi}_{\area^{*}}}}_{2}^{2},\label{eq:transition-task-func-1}
\end{equation}
which drives $\point^{*}$ towards $\dq{\pi}_{\area^{*}}$, as illustrated
in Figure~\ref{fig:Task-Func-Transition-1}. The complete optimisation
problem is written 
\begin{equation}
\argminimfive{\continput\in}{\dot{\boldsymbol{q}},\boldsymbol{s}}{\Psi_{T_{1}}+\regweighting^{2}\norm{\dot{\myvec q}}_{2}^{2}+\slackweighting^{2}\norm{\myvec s}_{2}^{2}}{\boldsymbol{W}_{T_{1}}\dot{\boldsymbol{q}}\preceq\boldsymbol{w}_{T_{1}}+\myvec s_{T_{1}}}{\myvec R\dot{\boldsymbol{q}}=0}{\myvec w_{\min}\preceq\dot{\myvec q}\preceq\myvec w_{\max}}{\qvmin\preceq\dot{\myvec q}\preceq\qvmax}{\veczero_{\numslackvariables}\preceq\boldsymbol{s}\preceq\boldsymbol{s}_{\max},}\label{eq:transition-stage-optimisation-1}
\end{equation}
where
\[
\mymatrix W_{T_{1}}=\left[\begin{array}{c}
\mymatrix W_{B}\\
\mymatrix W_{x}\\
\mymatrix W_{\text{Bnd}}
\end{array}\right],\;\myvec w_{T_{1}}=\left[\begin{array}{c}
\myvec w_{B}\\
\myvec w_{x}\\
\myvec w_{\text{Bnd}}
\end{array}\right],\;\myvec s_{T_{1}}=\left[\begin{array}{c}
\myvec s_{B}\\
\myvec s_{x}\\
\myvec s_{\text{Bnd}}
\end{array}\right],
\]
with $\mymatrix W_{B}$ given in \ref{eq:constraints-breaking-stage}.

In this stage, the balance constraint \eqref{eq:balance} is defined
using the region $\Pi_{\text{without}}$ and the contact slippage
constraint \eqref{eq:contact-slippage} uses $\allcontactpatches_{\text{fixed}}=\allcontactpatches\setminus\left\{ \patch^{*}\right\} $.

Alternatively, if condition \eqref{eq:boundary-condition}\emph{ }is
\emph{not} \emph{satisfied}, then constraint \eqref{eq:respect-all-boundary-planes}
is not applied and a new task function is written
\begin{equation}
\Psi_{T_{2}}\left(\dot{\myvec q}\right)\triangleq\norm{\mymatrix J_{\point^{*},\dq l_{\area^{*}}}\dot{\myvec q}+\eta_{o}\tilde{D}_{\point^{*},\dq l_{\area^{*}}}}_{2}^{2},\label{eq:transition-task-func-2}
\end{equation}
where $\dq l_{\area^{*}}\in\mathcal{H}_{p}\cap\dq{\mathcal{S}}$ is
a line perpendicular to the contact plane intersecting its centroid,
similar to that in \eqref{eq:breaking-task-func}. Thus this additional
task function brings $\point^{*}$ into the region described by $\Pi_{\textrm{bound}}$,
as illustrated in Figure~\ref{fig:Task-Func-Transition-2}. The problem
is hence re-written to include \emph{both} task functions

\begin{equation}
\argminimfive{\continput\in}{\dot{\boldsymbol{q}},\boldsymbol{s}}{\Psi_{T_{2}}+\Psi_{T_{1}}+\regweighting^{2}\norm{\dot{\myvec q}}_{2}^{2}+\slackweighting^{2}\norm{\myvec s}_{2}^{2}}{\boldsymbol{W}_{T_{2}}\dot{\boldsymbol{q}}\preceq\boldsymbol{w}_{T_{2}}+\myvec s_{T_{2}}}{\myvec R\dot{\boldsymbol{q}}=0}{\myvec w_{\min}\preceq\dot{\myvec q}\preceq\myvec w_{\max}}{\qvmin\preceq\dot{\myvec q}\preceq\qvmax}{\veczero_{\numslackvariables}\preceq\boldsymbol{s}\preceq\boldsymbol{s}_{\max},}\label{eq:transition-stage-optimisation-2}
\end{equation}
where
\[
\mymatrix W_{T_{2}}=\left[\begin{array}{c}
\mymatrix W_{B}\\
\mymatrix W_{x}
\end{array}\right],\;\myvec w_{T_{2}}=\left[\begin{array}{c}
\myvec w_{B}\\
\myvec w_{x}
\end{array}\right],\;\myvec s_{T_{2}}=\left[\begin{array}{c}
\myvec s_{B}\\
\myvec s_{x}
\end{array}\right].
\]

As in the breaking stage, either problem \eqref{eq:transition-stage-optimisation-1}
or \eqref{eq:transition-stage-optimisation-2} is solved (as appropriate)
and numerical integration is used to find the next configuration.
Once conditions \eqref{eq:boundary-condition} and \eqref{eq:plane-condition}
are both satisfied, the transition stage terminates and the posture
generator progresses to the placement optimisation stage. If either
condition is not satisfied, then the process repeats, returning a
failure state if a user-defined maximum number of integration step
cycles is reached.

\subsubsection{Placement Optimisation Stage\label{subsec:Optimisation-Stage}}

Finally, in the placement optimisation stage the posture generator
optimises the position of the new contact with respect to the potential
field $\potential{\config}$, given in \eqref{eq:pot-field-v2}. Since
only the contact patch $\patch^{*}$ is free to move in this stage,
only the contact patch-specific sub-field $\potfield_{\patch^{*}}\left(\config\right)$,
given by \eqref{eq:field-patch-subterm}, is relevant to this optimisation,
since the other sub-terms remain constant. The time derivative of
$\potfield_{\patch^{*}}\left(\config\right)$ is given by 
\begin{equation}
\begin{aligned}\frac{d}{dt}\potfield_{\patch^{*}}\left(\config\right) & =\squaredistanceROC{\point^{*}}{\line_{*,\closest}}+\potfieldweighting\squaredistanceROC{\point^{*}}{\plane_{*,\closest}}\end{aligned}
,\label{eq:Pot-field-ROC}
\end{equation}
where $\squaredistance{\point^{*}}{\line_{*,\min}}$ and $\squaredistance{\point^{*}}{\plane_{*,\min}}$
have the same meanings as in \eqref{eq:field-patch-subterm}. The
Jacobian matrix of $\potfield_{\patch^{*}}\left(\config\right)$ (denoted
$\mymatrix J_{\potfield_{\patch^{*}}}$) is straightforwardly defined
by writing
\begin{equation}
\begin{aligned}\squaredistanceROC{\point^{*}}{\line_{*,\closest}}+\potfieldweighting\squaredistanceROC{\point^{*}}{\plane_{*,\closest}} & =\underbrace{\left(\mymatrix J_{\point^{*},\line_{*,\closest}}+\potfieldweighting\mymatrix J_{\point^{*},\plane_{*,\closest}}\right)}_{\mymatrix J_{\potfield_{\patch^{*}}}}\dot{\config}\end{aligned}
.\label{eq:Pot-field-jacobian}
\end{equation}

With this, the task function for the placement optimisation stage
is given as 
\begin{equation}
\Psi_{O}\left(\dot{\myvec q}\right)=\norm{\mymatrix J_{\potfield_{\patch^{*}}}\dot{\myvec q}+\eta_{o}\potfield_{\patch^{*}}\left(\config\right)}_{2}^{2},\label{eq:placement-task-func}
\end{equation}
which drives the robot to move contact patch $\patch^{*}$ to a position
that minimises $\potfield_{\patch^{*}}\left(\config\right)$ and,
consequently, $\potential{\config}$. 

\begin{figure}
\centering
\includegraphics[width=0.5\columnwidth]{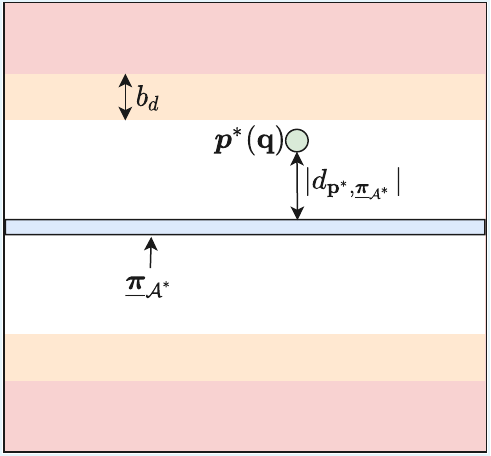}
\caption{Illustration of the VFI constraint \eqref{eq:max-FOI-to-contact-plane-dist}
for keeping $\protect\patch^{*}$ close to $\protect\area^{*}$ while
its location is optimised.\label{fig:VFI-Plane-Dist}}
\end{figure}

Finally, we limit the maximum square distance from $\point^{*}$ to
$\dq{\pi}_{\mathrm{\area^{*}}}$ (illustrated in Figure~\ref{fig:VFI-Plane-Dist})
by applying the constraint 
\begin{equation}
\mymatrix J_{\quat p^{*},\dq{\pi}_{\area^{*}}}\dot{\myvec q}\leq-\eta_{d}\left(D_{\quat p^{*},\dq{\pi}_{\area^{*}}}+b_{d}^{2}\right)+s_{\quat p^{*},\dq{\pi}_{\area^{*}}},\label{eq:max-FOI-to-contact-plane-dist}
\end{equation}
written in matrix/vector form as 
\[
\mymatrix W_{\area^{*}}\dot{\config}\preceq\myvec w_{\area^{*}}+\myvec s_{\area^{*}}.
\]

The complete optimisation problem can thus be written as 
\begin{equation}
\argminimfive{\continput\in}{\dot{\boldsymbol{q}},\boldsymbol{s}}{\Psi_{O}+\regweighting^{2}\norm{\dot{\myvec q}}_{2}^{2}+\slackweighting^{2}\norm{\myvec s}_{2}^{2}}{\boldsymbol{W}_{O}\dot{\boldsymbol{q}}\preceq\boldsymbol{w}_{O}+\myvec s_{O}}{\myvec R\dot{\boldsymbol{q}}=0}{\myvec w_{\min}\preceq\dot{\myvec q}\preceq\myvec w_{\max}}{\qvmin\preceq\dot{\myvec q}\preceq\qvmax}{\veczero_{\numslackvariables}\preceq\boldsymbol{s}\preceq\boldsymbol{s}_{\max},}\label{eq:placement-stage-optimisation}
\end{equation}
where
\[
\mymatrix W_{O}=\left[\begin{array}{c}
\mymatrix W_{T_{1}}\\
\mymatrix W_{\area^{*}}
\end{array}\right],\;\myvec w_{O}=\left[\begin{array}{c}
\myvec w_{T_{1}}\\
\myvec w_{\area^{*}}
\end{array}\right],\;\myvec s_{O}=\left[\begin{array}{c}
\myvec s_{T_{1}}\\
\myvec s_{\area^{*}}
\end{array}\right].
\]
As in the transition stage, the balance constraint \eqref{eq:balance}
is defined using the region $\Pi_{\text{without}}$ and the contact
slippage constraint \eqref{eq:contact-slippage} uses $\allcontactpatches_{\text{fixed}}=\allcontactpatches\setminus\left\{ \patch^{*}\right\} $.

Also like the previous stages, problem \eqref{eq:placement-stage-optimisation}
is solved and numerical integration is used to find the next configuration.
If $\myvec u\approx\veczero$, then the placement optimisation stage
terminates and the posture generation process is complete. Alternatively,
if $\myvec u\not\approx\veczero$, then the process repeats. If the
user-defined maximum number of integration step cycles is reached,
then the process terminates and the most recent iterate is returned.

\section{Contact-critical Local Minima\label{sec:Local-Minima}}

This section discusses contact-critical local minima (CLM) problems
that might affect CBM planners that rely on potential-field-based
guide paths. The causes of these local minima are examined, and heuristics
are developed to mitigate them.

\subsection{Causes of Local Minima\label{subsec:Explanation-of-Problem}}

\begin{figure*}[t]
\noindent\begin{minipage}[t]{1\columnwidth}%
\centering%
\fcolorbox{black}{white}{\includegraphics[width=1\textwidth]{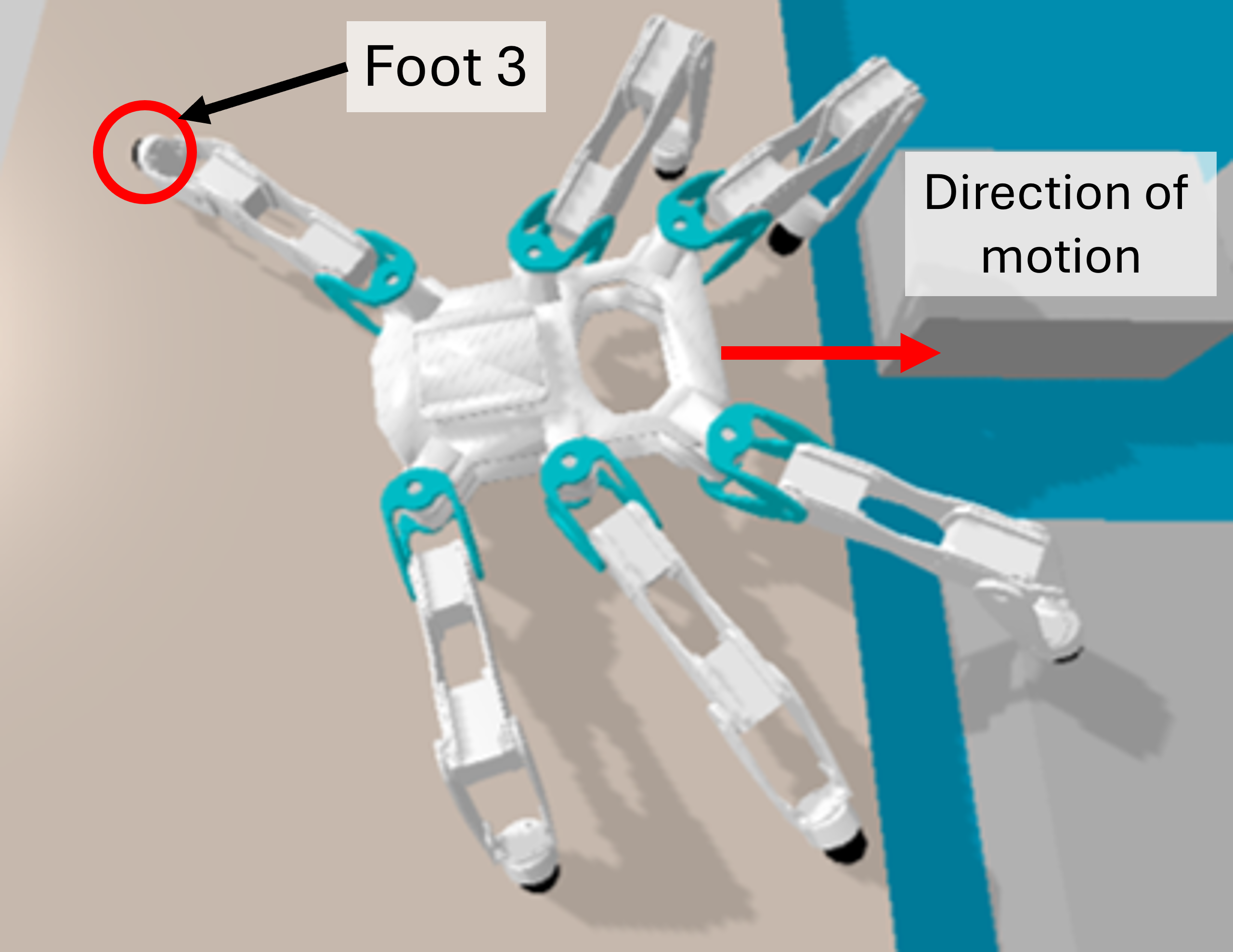}} \caption{The Corin hexapod during a CLM.\label{fig:Stuck-Config-stepping-stones}}
\end{minipage}\hfill{}%
\noindent\begin{minipage}[t]{1\columnwidth}%
\centering %
\fcolorbox{black}{white}{\includegraphics[width=1\textwidth]{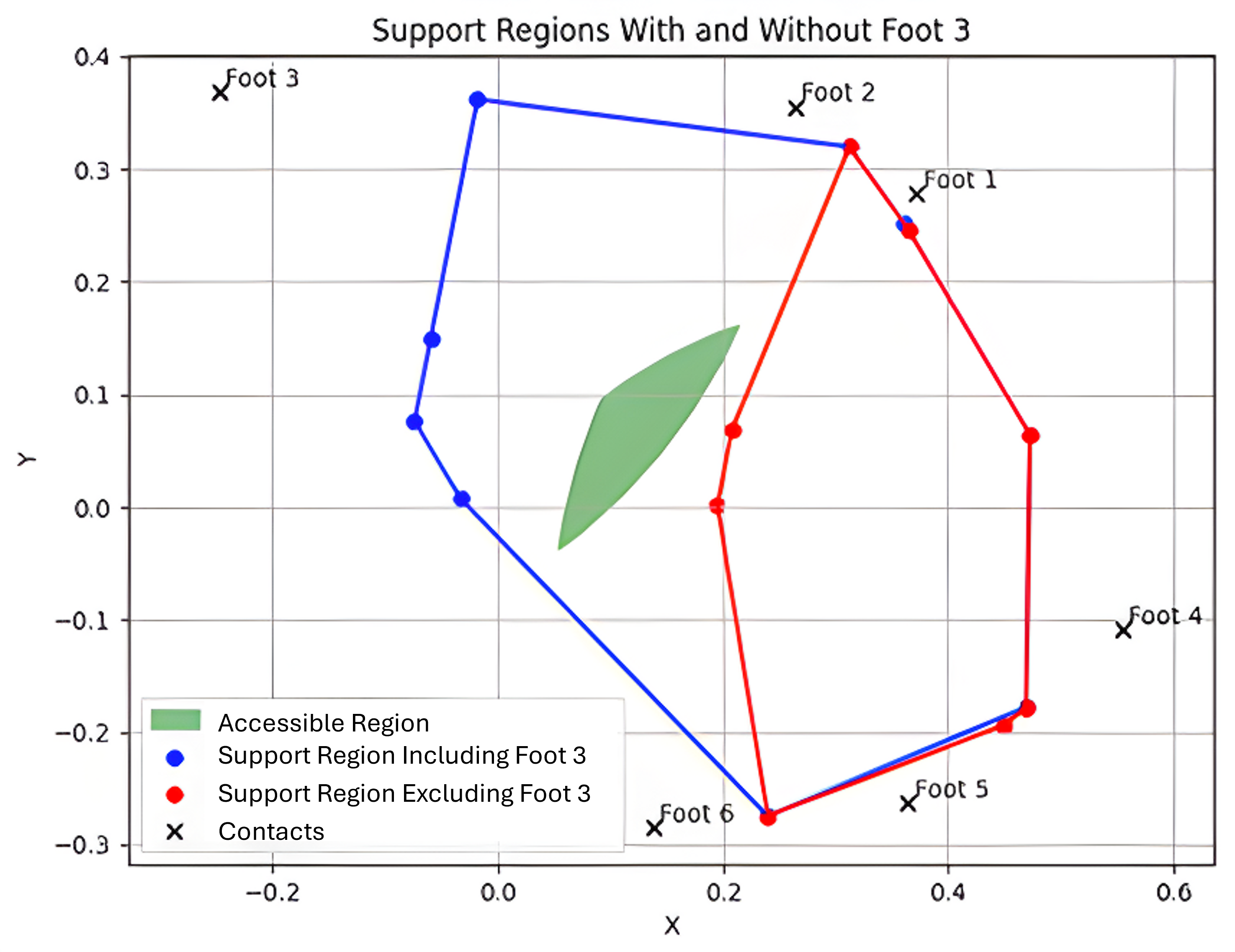}} \caption{The support regions with (\emph{blue}) and without (\emph{red}) the
rear left foot (foot 3) for the Corin hexapod in the configuration
shown in Figure~\ref{fig:Stuck-Config-stepping-stones}. \label{fig:stuck-config-support-regions}}
\end{minipage}
\end{figure*}

CLM are a form of local minima that coincide with the confluence of
the following three conditions:
\begin{itemize}
\item \textbf{Condition 1} - The robot cannot meaningfully progress without
breaking a particular contact, denoted $\contact^{*}=\left(\patch^{*},\area^{*},\point^{\area^{*}}\right)$.
This most often occurs when the other contact patches have each been
moved from their previous position and cannot be moved much further
until $\patch^{*}$ is also moved.
\item \textbf{Condition 2} - The contact $\contact^{*}$ is critical to
the robot's balance, meaning that if $\contact^{*}$ were broken in
the current configuration, then the robot would fall. Thus, in order
to safely break contact $\contact^{*}$, the robot must first move
its centre of mass (COM) to a region where $\contact^{*}$ is not
required for balance. A horizontal cross-section of this region for
a hexapod robot in the configuration shown in Figure~\ref{fig:Stuck-Config-stepping-stones}
is shown in \emph{red} in Figure~\ref{fig:stuck-config-support-regions}.
\item \textbf{Condition 3} - The robot cannot move its COM into the required
region without breaking contact $\contact^{*}$ \emph{first}. An approximation
of the reachable area for the robot torso (which typically contains
the COM) is shown in \emph{green} in Figure~\ref{fig:stuck-config-support-regions}.
Since there is no overlap between the green region and the red polygon
in Figure~\ref{fig:stuck-config-support-regions}, the robot cannot
reach a configuration from which foot 3 can be safely lifted.
\end{itemize}
\begin{figure}[t]
\centering{}\includegraphics[width=1\columnwidth]{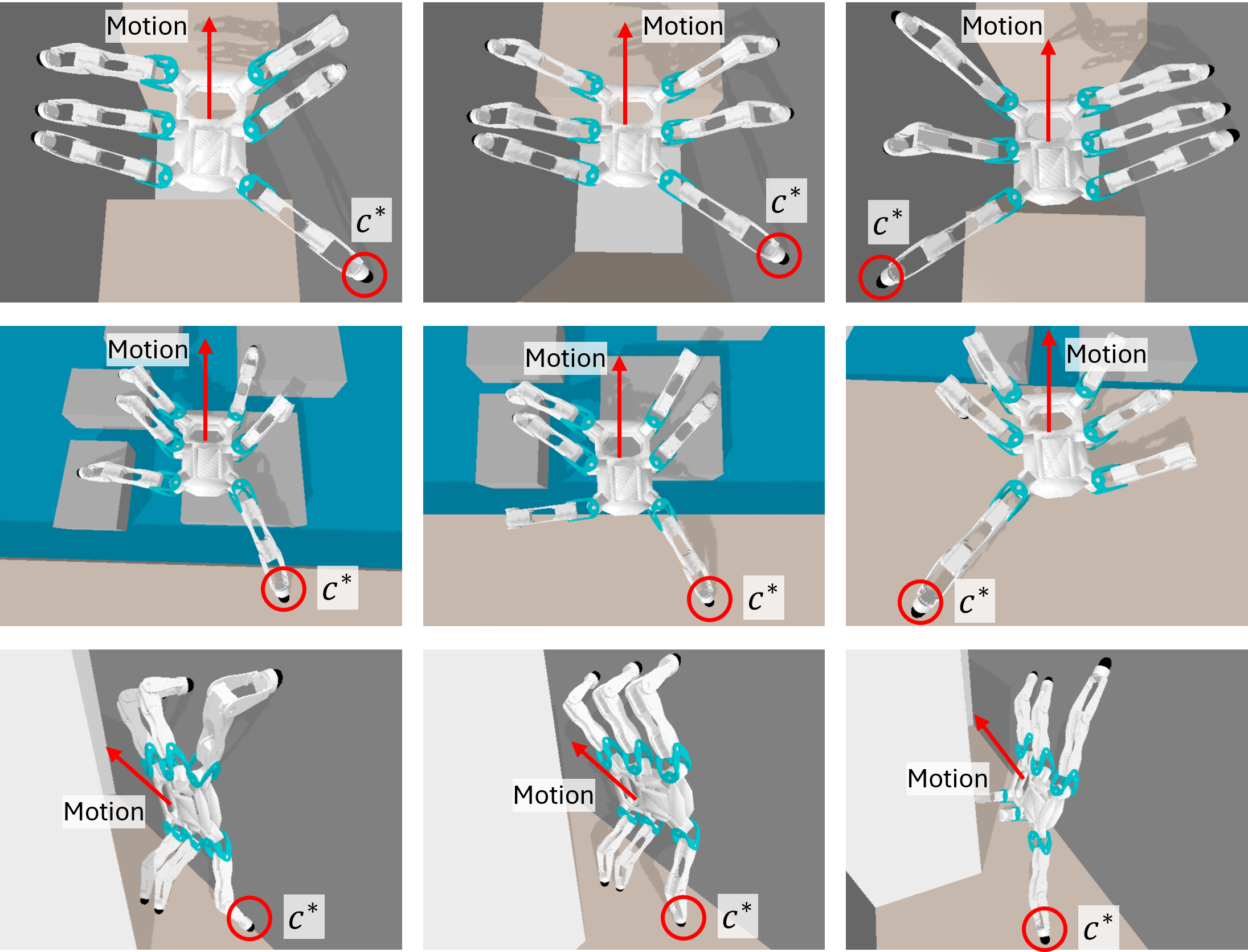}
\caption{Examples of the Corin hexapod during contact-critical local minima.
In each case, contact $\protect\contact^{*}$ is indicated by a \emph{red
circle}, and the direction of motion by the \emph{red arrow}.\label{fig:Several-Stuck-Examples}}
\end{figure}

Figure~\ref{fig:Several-Stuck-Examples} shows examples of the robot
during CLM in various scenarios. In principle, the robot could escape
from CLM by moving another limb to relieve contact $\contact^{*}$,
making it no-longer critical for balance. However, the potential fields
that guide CBM planners, such as CVBFP and RHCP, strongly disincentivise
moving contact patches away from the goal \citep{CVBFP-Escande2013MainPaper}.
Thus, the algorithm's respective posture generators rarely form contacts
that can relieve $\contact^{*}$, and when they do their search algorithms
will not typically choose to expand these nodes. Hence, escaping from
CLM can be very difficult for both planners.

While trapped in a CLM, the planning processes may generate very large
quantities of nodes descending from the `stuck' position, exploring
many different future step sequences, none of which result in significant
forward movement, since meaningful progress remains impossible without
breaking contact $\contact^{*}$. This typically continues until the
planners become unable to generate any more nodes descending from
the stuck position and are forced to retreat to a prior state. Other
avenues may then be explored in which contact $\contact^{*}$ \emph{can}
be broken, allowing progress to resume. Both planners prevent duplicate
positions from being repeated, guaranteeing that they will both break
out of CLM after a finite time. However, in practice this may take
too long.

CLM-causing configurations are favoured by search processes, despite
their deleterious impacts, because the potential fields guiding CBM
planners incentivise the robot to move its limbs forward, with larger
steps being preferred. The front legs of the Corin hexapod, for example,
typically have more space to move than others, meaning that the front
legs often move first, followed by the middle legs and finally the
rear legs. This increases the likelihood that one rear leg will become
critical for balance and unable to be safely lifted.

\subsection{Improving Resilience\label{subsec:Improving-Resilience}}

In response to the problems posed by CLM, we make two changes to the
basic RHCP shown in Figure~\ref{fig:RHCP-Flowchart}. First, the
potential field is reformulated to discourage the robot from entering
positions resembling those shown in Figure~\ref{fig:Several-Stuck-Examples}.
Second, we add a heuristic mechanism that detects the presence of
CLM and takes corrective action to escape. These measures were chosen
because they do not disallow any configurations that were previously
permitted, meaning that the planner is not made any more conservative.

\subsubsection{Encouraging Movement of Farthest Legs}

To discourage the robot from taking on configurations resembling those
in Figure~\ref{fig:Several-Stuck-Examples}, the potential field
is reformulated such that movements of the feet that are farthest
away from the end position are prioritised over those closer to the
end position.

\begin{figure}[t]
\centering{}\includegraphics[width=1\columnwidth]{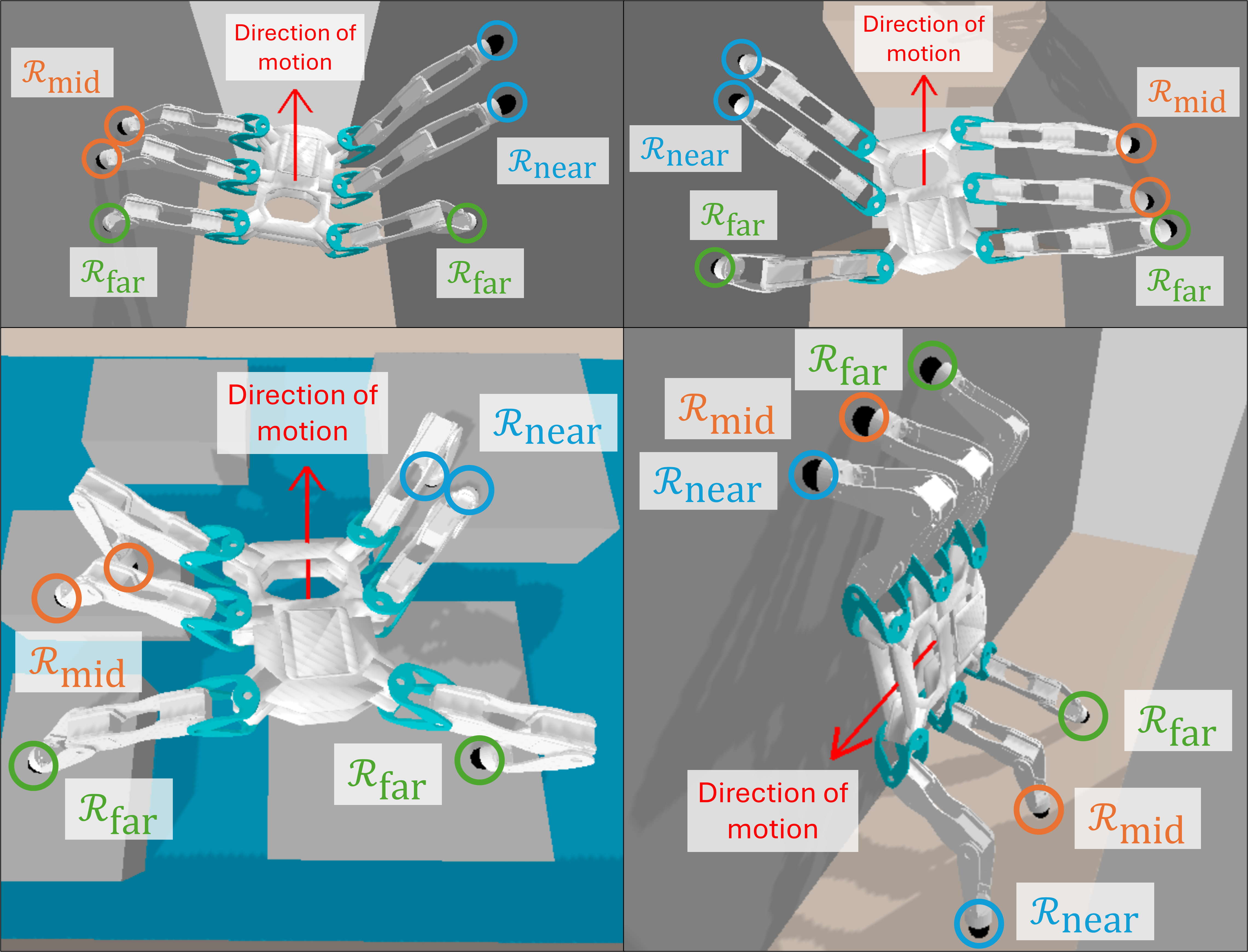}
\caption{Example configurations in different scenarios with the contacts assigned
to either $\protect\nearpatches$, $\protect\midpatches$, or $\protect\farpatches$.}
\label{fig:leg sets}
\end{figure}

Let the point $\point_{\stance_{\goal}}\in\workspace$ be the mean
3D location of all of the contacts comprising the goal stance $\stance_{\goal}$.
When the potential is determined for a configuration $\config$, each
of the robot's contact patches $\patch\in\allcontactpatches$ are
automatically assigned to one of three sets based on their distance
from $\point_{\stance_{\goal}}$. The nearest two patches are assigned
to $\nearpatches$, the furthest two to $\farpatches$, and the remaining
two to $\midpatches$. Figure~\ref{fig:leg sets} shows examples
of configurations in different scenarios with their contacts assigned
to these sets.

The revised potential field is then given as

\begin{equation}
\potential{\config}=\sum_{i=1}^{\numpatches}\fieldpatchweighting\potfield_{\patch_{\indexrobot}}\left(\config\right),\label{eq:pot-field-v2}
\end{equation}

where $\fieldpatchweighting\in\left(0,2\right)$ is determined based
on the set to which the contact patch $\patch_{\indexrobot}$ belongs,

\begin{equation}
\fieldpatchweighting=\begin{cases}
1-\fieldpatchoffset & \text{if }\patch_{\indexrobot}\in\nearpatches\\
1 & \text{if }\patch_{\indexrobot}\in\midpatches\\
1+\fieldpatchoffset & \text{if }\patch_{\indexrobot}\in\farpatches,
\end{cases}\label{eq:foot-offsets}
\end{equation}
 where $\fieldpatchoffset\in\left[0,1\right)$. The term $\potfield_{\patch_{\indexrobot}}\left(\config\right)$
is calculated as given in \eqref{eq:field-patch-subterm}.

This change causes the search process to prioritise the limbs that
are the furthest away from the goal stance before moving those closest.
This causes the planning process to disfavour moves that will result
in stuck positions like those shown in Figure~\ref{fig:Several-Stuck-Examples}.

One drawback of this approach is that the contact patches of different
robots must be sorted into sets differently depending on their geometries.
For example, a quadruped may assign the two closest patches to $\nearpatches$,
the two farthest to $\farpatches$, and leave $\midpatches$ empty.
Likewise, a humanoid may allocate one patch each to $\nearpatches$
and $\farpatches$. This need to specify how patches should be allocated
for different robot geometries somewhat reduces the generality of
the approach, but these adaptations are judged unlikely to pose a
significant obstacle to generalisation in most cases.

\subsubsection{Intervention Mechanism}

\begin{figure*}[t]
\centering{}\includegraphics[width=1\textwidth]{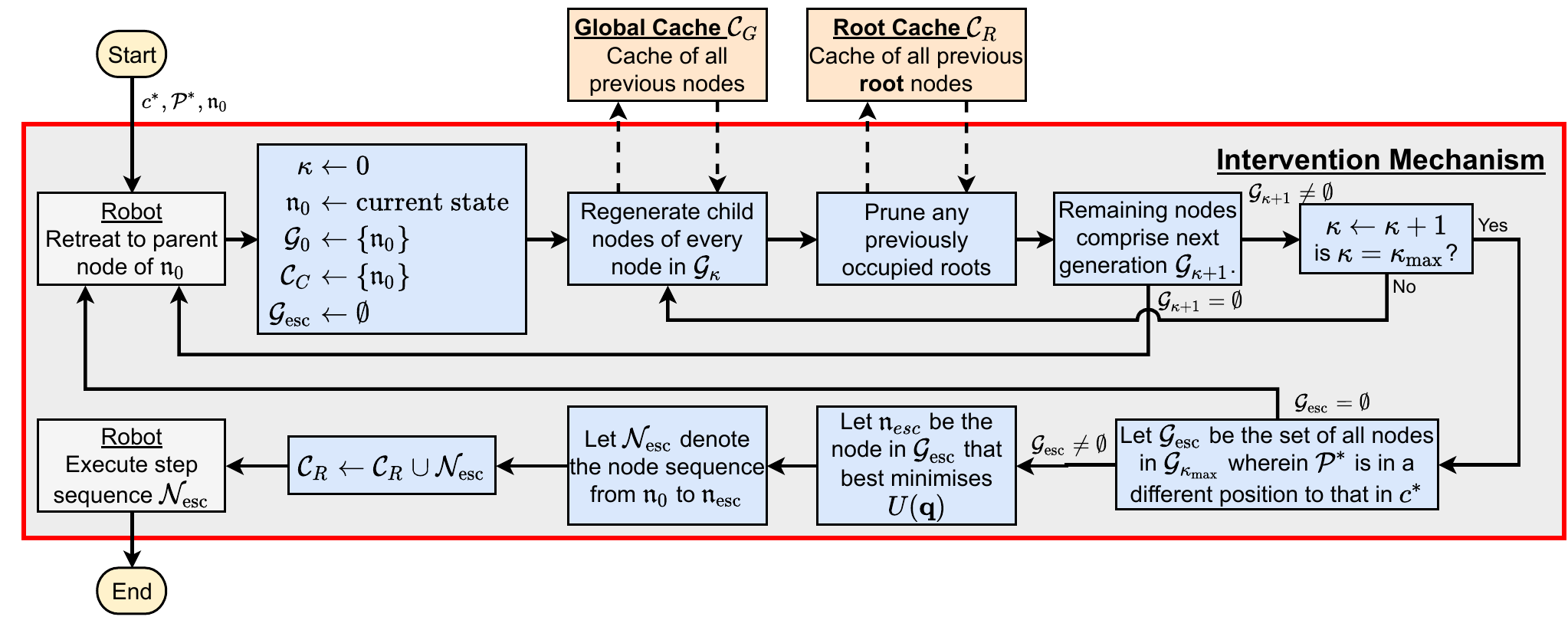}
\caption{Flowchart illustrating the intervention mechanism that mitigates CLM.\label{fig:Int-Mech-Flowchart}}
\end{figure*}

Though the revised potential field makes CLM issues\emph{ less likely}
to occur, it does not prevent them entirely. Therefore, we also design
a heuristic intervention mechanism that detects CLM and takes corrective
action to break out. The mechanism is triggered if either of two conditions
are met:
\begin{enumerate}
\item There exists a contact $\contact^{*}$ that has been impossible to
break safely in all of the $\interlim\ispositiveint$ most recently
occupied root nodes.
\item The planning process has encountered a dead end (\emph{i.e.}, $\horizon=\emptyset$)
\emph{and} there exists a contact $\contact^{*}$ that cannot be safely
broken from the current root node.
\end{enumerate}
When the mechanism is triggered, it takes corrective action to break
the search process out of the local minimum (illustrated in Figure~\ref{fig:Int-Mech-Flowchart}).
This begins with the robot retreating to the previous state (the parent
node of the current root node). When the robot has arrived in that
state, it is redesignated $\rootnode$ and becomes the new root node.
The child generation process is done by restoring child nodes from
a global cache, denoted $\cache_{G}$ in Figure~\ref{fig:Int-Mech-Flowchart},
until generation $\generation_{\horizondepth}$ is reached.

As each generation of child nodes is restored, any nodes that have
previously been the root node are purged from the set. This prevents
the system from repeating previous stances that are known to be ineffective.

Once $\indexgeneration=\kmax$, the nodes comprising $\generation_{\horizondepth}$
are each checked in order to determine whether contact patch $\patch^{*}$
has moved to a different location, thus breaking contact $\contact^{*}$.
Any nodes that meet this condition are added to a set of `escape
nodes' denoted $\escapenodes\subset\horizon$. The lowest-potential
node in $\escapenodes$, designated $\node_{\text{esc}}$, is then
identified and the sequence of nodes leading from $\rootnode$ to
$\node_{\text{esc}}$ is designated $\nodesequence_{\text{esc}}$.
The robot then performs the series of motions required to execute
the sequence $\nodesequence_{\text{esc}}$, arriving at $\node_{\text{esc}}$,
which becomes the new root node. The nodes in $\nodesequence_{\text{esc}}$
are hence added to the root cache $\cache_{R}$, and planning resumes
as normal.

\begin{figure*}[!t]
\centering{}\includegraphics[width=1\textwidth]{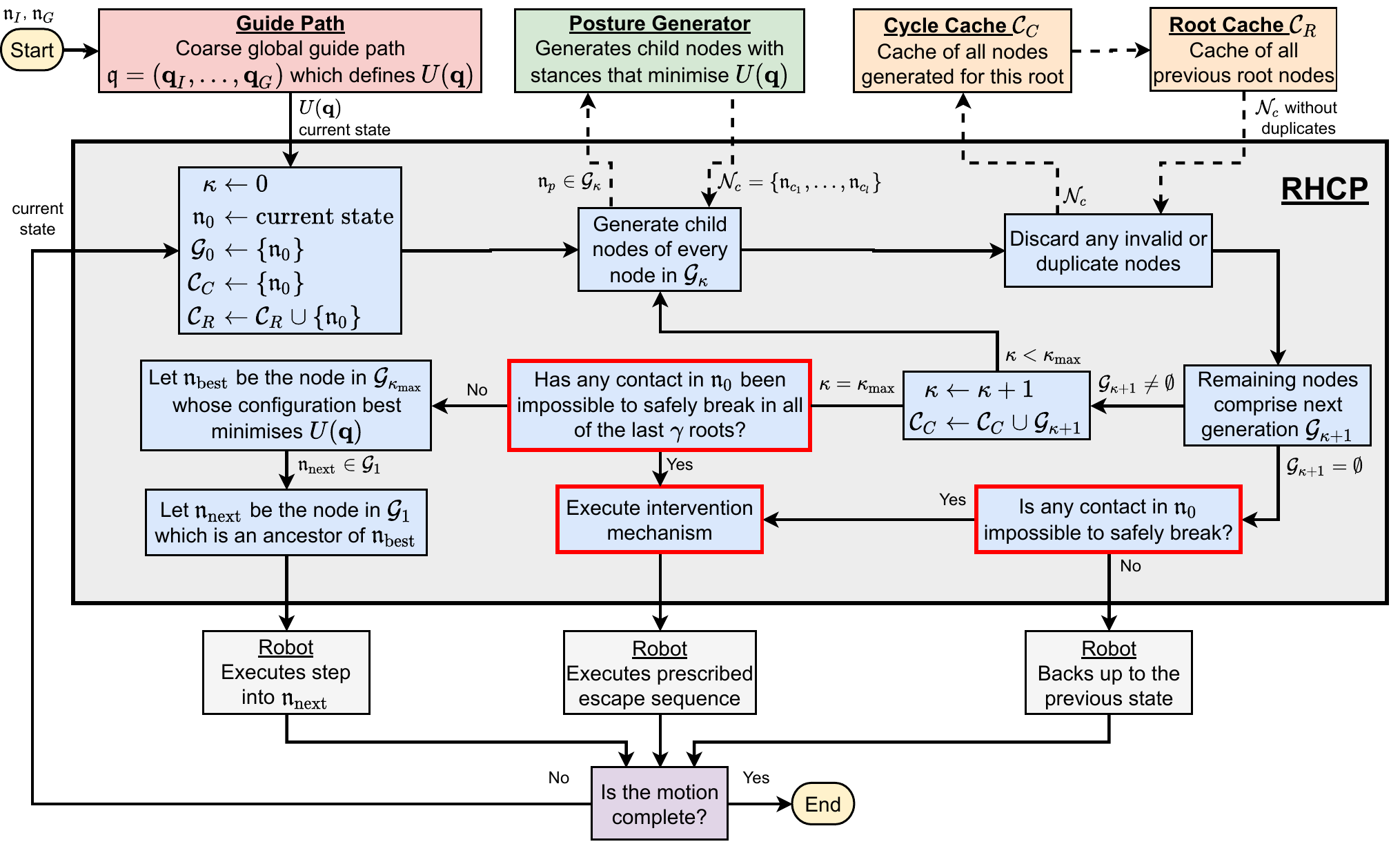}
\caption{Updated flowchart illustrating the RHCP planning algorithm with the
improvements, highlighted in \emph{red}\textbf{\emph{,}} to mitigate
CLM: an improved potential-field guide-path and an intervention mechanism
that finds escape nodes. \label{fig:Updated-RHCP-Flowchart}}
\end{figure*}

If no valid escape nodes can be found (\emph{i.e.}, $\escapenodes=\emptyset$),
then the robot retreats again to the parent node of $\rootnode$ and
repeats the process. This continues until either a valid escape node
is found, or the process has backed up to the initial state $\node_{\initial}$
(which has no parent node) at which point the planning is considered
to have failed. An updated flowchart displaying the revised algorithm
is shown in Figure~\ref{fig:Updated-RHCP-Flowchart}.

\section{Configuration Update by Numerical Integration With Contact Drift
Correction \label{sec:Numerical-Integration}}

Once a control input $\myvec u$ is generated by solving the optimisation
problems \eqref{eq:breaking-stage-optimisation} (breaking stage),
\eqref{eq:transition-stage-optimisation-1} (transition stage), \eqref{eq:transition-stage-optimisation-2}
(transition stage when \eqref{eq:plane-condition} is not satisfied),
or \eqref{eq:placement-stage-optimisation} (placement optimisation),
the robot configuration is updated via numerical integration. First,
the configuration velocity vector,
\[
\dot{\myvec q}\triangleq\begin{bmatrix}\vecfour{\dot{\quat r}(t)} & \vecthree{\dot{\quat p}(t)} & \dot{\boldsymbol{\theta}}(t)\end{bmatrix}^{T},
\]
is extracted from $\myvec u=\left(\dot{\myvec q},\myvec s\right)$,
and the torso pose derivative is calculated as 
\begin{equation}
\dot{\dq x}(t)=\dot{\quat r}(t)+\varepsilon\frac{1}{2}\left(\dot{\quat p}(t)\quat r(t)+\quat p(t)\dot{\quat r}(t)\right).\label{eq:torso-pose-derivative}
\end{equation}
Then, the next torso pose $\dq x(t+\tau)$ and joint angle vector
$\boldsymbol{\theta}(t+\tau)$, where $\tau\in(0,\infty]$ is the
integration step, are calculated as
\begin{align}
\dq x(t+\tau) & =\exp\left(\tau\dot{\dq x}(t)\dq x(t)^{*}\right)\dq x(t),\label{eq:torso-pose-numerical-integration}\\
\boldsymbol{\theta}(t+\tau) & =\boldsymbol{\theta}(t)+\tau\dot{\boldsymbol{\theta}}(t),\label{eq:joint-angs-numerical-integration}
\end{align}
where the (dual quaternion) exponential map and group operation in
\eqref{eq:torso-pose-numerical-integration} ensures that the underlying
topology of unit dual quaternions is respected \citep{DQ-Adorno2017Fundamentals}.
The pose $\dq x(t+\tau)$ can hence be decomposed into $\quat r(t+\tau)$
and $\quat p(t+\tau)$, and used to form the overall configuration
\begin{equation}
\myvec q(t+\tau)\triangleq\begin{bmatrix}\vecfour{\rot(t+\tau)} & \vecthree{\point(t+\tau)} & \jointangles(t+\tau)\end{bmatrix}^{T}.\label{eq:full-config-numerical-integration}
\end{equation}

\subsection{Contact drift correction}

As discussed in Section~\ref{subsec:VFIs-under-discretisation},
discrete integration steps can cause robot bodies to enter restricted
zones, which we partially resolve with the addition of constraint
buffers. However, the relatively small safe zone of the contact sliding
constraint \eqref{eq:contact-slippage}, coupled with relatively large
integration step sizes, causes two problems:
\begin{enumerate}
\item To guarantee that the next configuration will remain within the safe
zone, the size of the buffer region would need to be greater than
the radius of the safe zone itself (\emph{i.e.,} $b_{d}>R_{\text{slip}}$
in Figure~\ref{fig:VFI-Slippage}), making satisfying the requirement
that $\squaredistance{\point_{\indexrobot}}{\point_{d_{\indexrobot}}}-R_{\text{slip}}^{2}\leq-b_{d}^{2}$
impossible.
\item Constraint \eqref{eq:contact-slippage} limits how fast $\point_{i}$
may approach the \emph{edge} of the safe zone, but the velocity of
$\point_{i}$ towards the \emph{centre} of the zone is not restricted.
Since the spherical safe zone is very small, $\point_{i}$ may pass
the centre of the zone with sufficient speed to cross the boundary
on the other side, due to discretisation, before the antipodal point's
VFI starts having an effect.
\end{enumerate}
These problems may be mitigated by using smaller integration steps,
however, this would mean more control cycles are necessary to plan
the same motion, directly increasing the execution time of the planner.
Instead, if any contact patch $\patch_{i}\in\allcontactpatches_{\text{fixed}}$
has drifted too far from its desired location, and hence violated
the constraint, a secondary constrained controller corrects the drift.

We define $\myvec D_{\text{drift}}$ as the stacked vector of square
distances between each fixed contact patch and its desired location,
given by 
\[
\myvec D_{\text{drift}}\triangleq\left[\begin{array}{c}
\squaredistance{\point_{1}}{\point_{d,1}}\\
\vdots\\
\squaredistance{\point_{n}}{\point_{d,n}}
\end{array}\right],\,\forall\patch_{i}\in\allcontactpatches_{\text{fixed}}.
\]
Writing the Jacobian matrix of $\myvec D_{\text{drift}}$ as $\mymatrix J_{\text{drift}}=\partial\myvec D_{\text{drift}}/\partial\myvec q$,
we hence design the drift-correction controller as
\begin{equation}
\argminimthree{\myvec u\in}{\dot{\myvec q}}{\norm{\mymatrix J_{\text{drift}}\dot{\myvec q}+\eta_{o}\myvec D_{\text{drift}}}_{2}^{2}+\lambda^{2}\norm{\dot{\myvec q}}_{2}^{2}}{\myvec R\dot{\boldsymbol{q}}=0}{\myvec w_{\min}\preceq\dot{\myvec q}\preceq\myvec w_{\max}}{\qvmin\preceq\dot{\myvec q}\preceq\qvmax,}\label{eq:drift-correction-problem}
\end{equation}
where $\myvec R,\myvec w_{\min},\myvec w_{\max}$ are defined in Section~\ref{subsec:Constraints-and-Objectives}.

\subsection{Variable integration step sizes}

Drift correction reduces the frequency of contact sliding. However,
relatively small values of $\tau$ remain necessary to avoid violating
other constraints (or else using very conservative buffers). We resolve
this by varying $\tau$ between $\tau_{\min}$ and $\tau_{\max}$,
making large steps where possible and smaller steps where necessary.

\begin{figure}[t]
\centering{}\includegraphics[width=1\columnwidth]{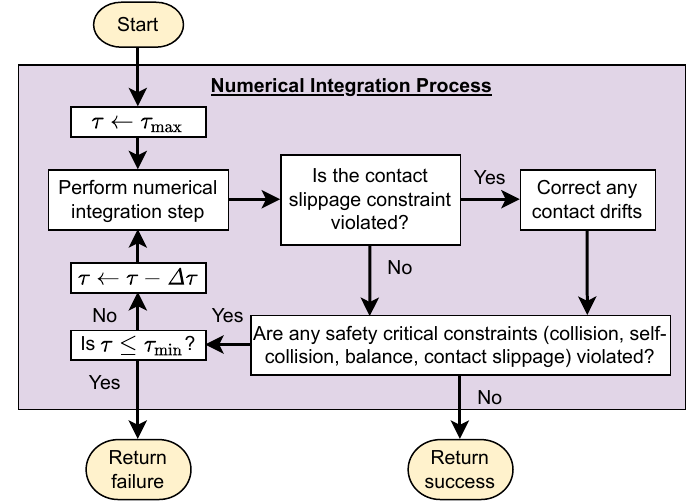}
\caption{Flowchart of the numerical integration process with variable integration
step sizes.\label{fig:NI-flowchart}}
\end{figure}

Whenever a value of $\boldsymbol{u}$ is returned, $\myvec q(t+\tau)$
is calculated using \eqref{eq:full-config-numerical-integration}
with $\tau=\tau_{\max}$. Any contact drifts are corrected, and the
resulting configuration is checked to verify that the constraints
are satisfied. If so, then the posture generator progresses to the
next optimisation. If, however, $\myvec q(t+\tau)$ is \emph{invalid}
(i.e., any safety critical constraint are violated when robot is at
configuration $\myvec q(t+\tau)$), then $\tau$ is decremented by
$\varDelta\tau$ and $\myvec q(t+\tau)$ is recalculated. This continues
until either a valid step is found or $\tau\leq\tau_{\min}$, in which
case the posture generator terminates and returns a failure. A flowchart
for this process is provided in Figure~\ref{fig:NI-flowchart}.

\section{Planning Simulations and Statistical Evaluation\label{sec:Results}}

\subsection{Experimental Setup}

\begin{figure}[t]
\centering{}\includegraphics[width=1\columnwidth]{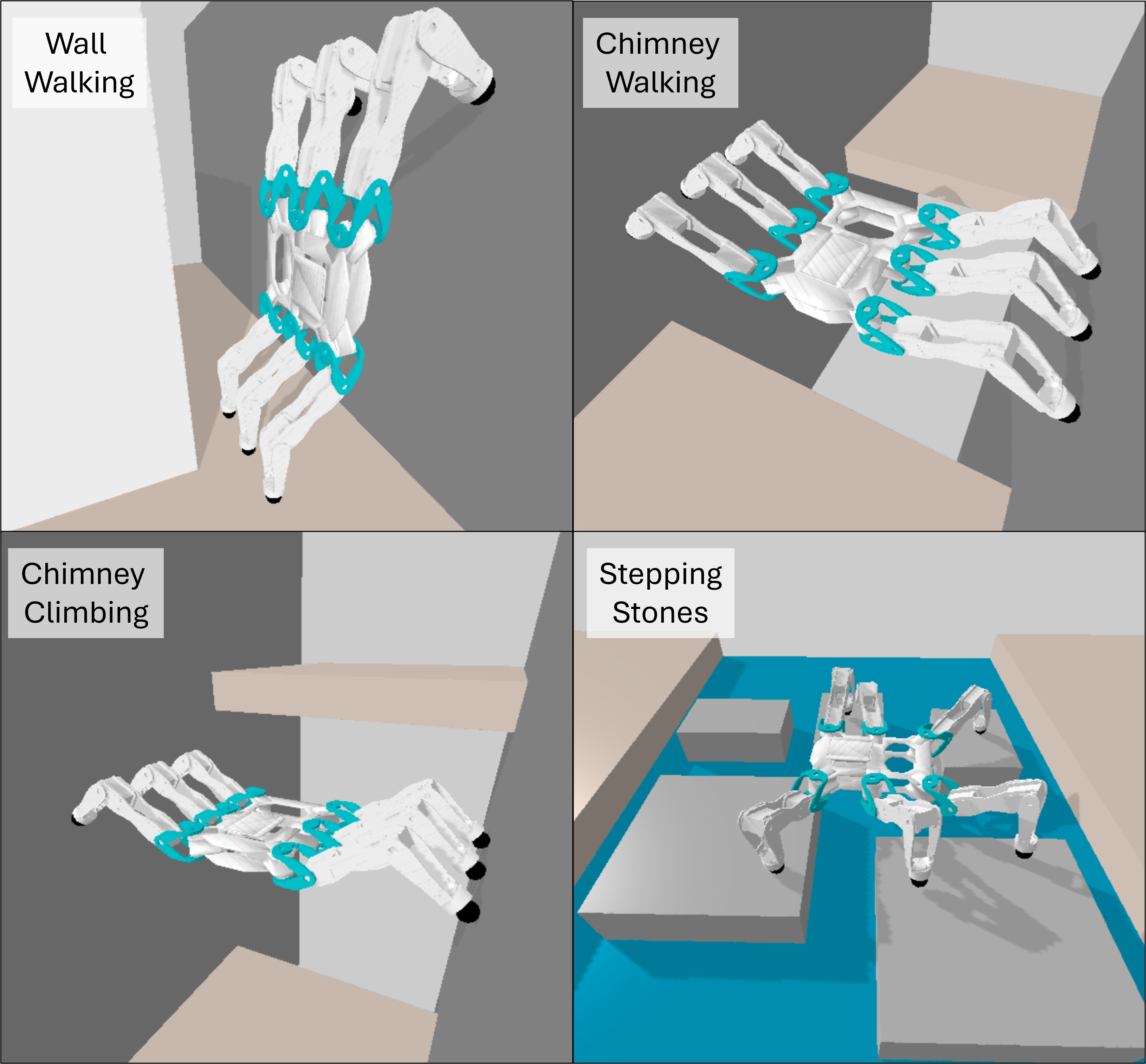}
\caption{Testing scenarios used for experimentation, referred to as Wall Walking
(\emph{top left}), Chimney Walking (\emph{top right), }Chimney Climbing
(\emph{bottom left}) and Stepping Stones (\emph{bottom} \emph{right}).}
\label{fig:environments}
\end{figure}

\begin{figure*}[!t]
\centering{}\includegraphics[width=1\textwidth]{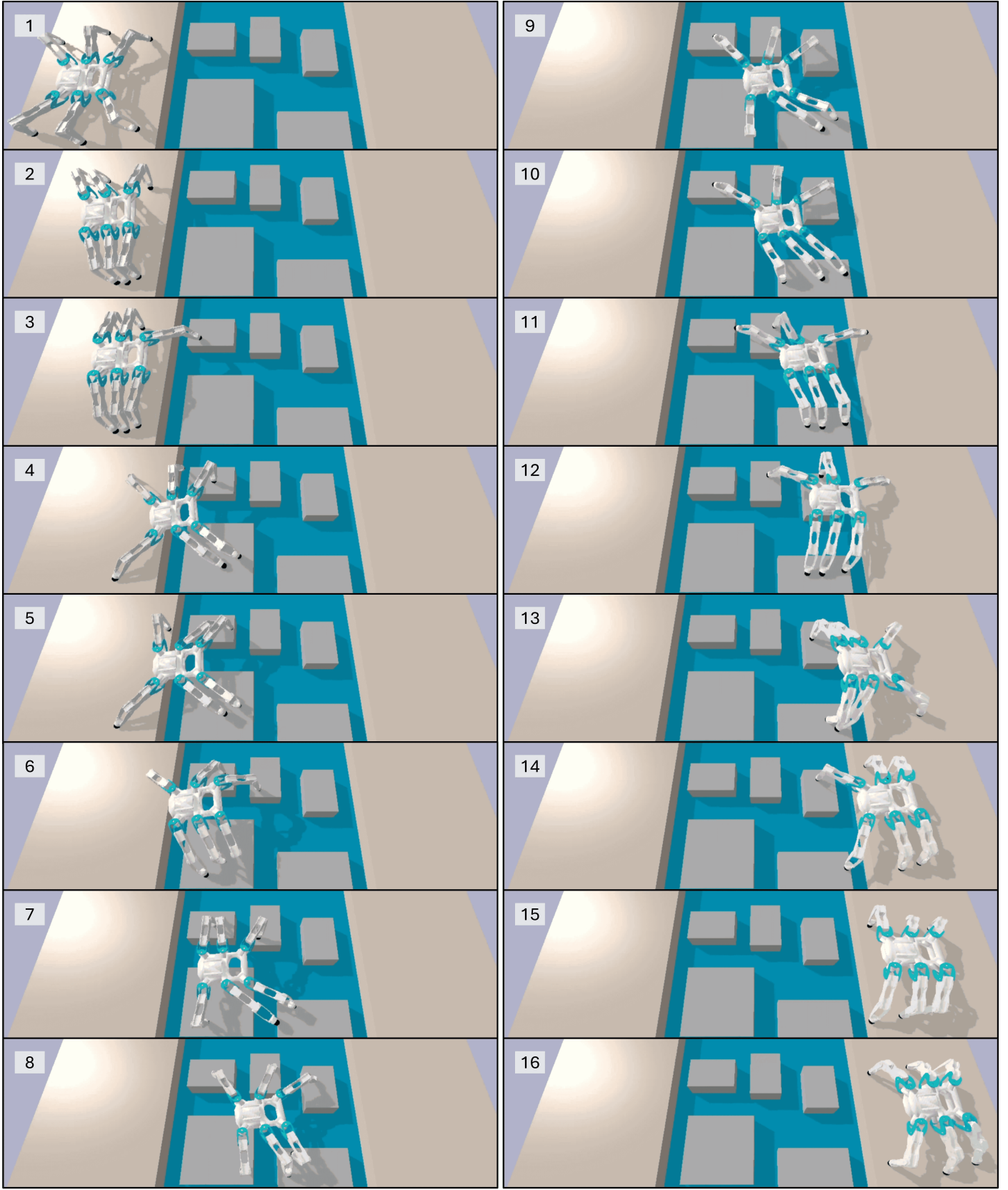}
\caption{Snapshots of a motion plan generated by RHCP in the Stepping Stones
scenario.}
\label{fig:motion-snapshots}
\end{figure*}

The RHCP planning framework was compared to CVBFP in the four simulated
scenarios depicted in Figure~\ref{fig:environments}. Each scenario
requires different forms of motion to be planned for the Corin hexapod.
In the Wall Walking scenario, the robot moves through a narrow one-meter
corridor by forming contacts on a vertical wall. In the Chimney Walking
scenario, the robot crosses a one-meter gap in the ground by using
the vertical surfaces exclusively. In the Chimney Climbing case, the
robot vertically climbs a one-meter-tall chimney. In the Stepping
Stones environment, the robot crosses a two-meter-wide `river' using
sparse, non-repeating surfaces. An animation of the robot executing
an example motion plan (generated by RHCP with $\kmax=2$) in each
scenario is available in the supplementary video for this article.
Snapshots of a motion plan in the Stepping Stones scenario are also
shown in Figure~\ref{fig:motion-snapshots}. 

Both planners generated motion plans for the Corin hexapod \citep{Corin-WebsiteEntry-UOM2020Corin}
in each scenario from partially randomised starting configurations.
RHCP used a variety of horizon depths for the tests, varying from
$\kmax=1$ to $\kmax=4$. CVBFP, and RHCP with $\kmax\in\left\{ 1,2,3\right\} $,
each generated 100 motion plans per scenario. Due to the large planning
times required, only 50 motion plans were generated for RHCP with
$\kmax=4$. 

The total time required to plan each motion was recorded, as well
as the number of total stance changes required to execute the resulting
motion plan. Note that lifting and then placing a foot is counted
as two separate stance changes. All tests were run in a computer running
Ubuntu 20.04.6 LTS (64 bits) with an Intel® Core™ i9-7900X Processor
and 62 GB RAM. Posture generator calls were batched and made concurrently
with a maximum of 20 threads. For CVBFP, all children/siblings of
a given node were generated in one batch, whereas in RHCP all nodes
in a given generation formed one batch. Both posture generators were
written in C++ and both the tree search methods were written in Python.
Where objects in the environment were modelled by dual quaternion
primitives, the kinematic calculations were handled by the DQ Robotics
C++ library \citep{DQRobotics}. Otherwise, the simulator Bullet was
used, accessed via its fast C++ interface \citep{BulletPhysics}.

This experimental set-up is very similar to that in our previous work
\citep{TAROSPaper-2025-Derwent2025Multi}, though with some important
differences. One difference is that here we analyse multiple horizon
depths, whereas in \citep{TAROSPaper-2025-Derwent2025Multi} only
a two-step horizon (\emph{i.e.,} $\kmax=2$) was considered. Additionally,
in \citep{TAROSPaper-2025-Derwent2025Multi} the balance criterion
of CVBFP was implemented as described in \citep{CVBFP-Escande2013MainPaper},
with no upper bound on the magnitudes of the contact forces. This
was also the case for RHCP's balance criterion, described in \citep{RHCP-Bretl2008SupportRegions}.
The effect of this is that CLM posed a much smaller problem for both
planners in \citep{TAROSPaper-2025-Derwent2025Multi} than they do
in this present work, which is more realistic, and hence the experimental
results in this present work are more insightful and physically plausible
than those given in \citep{TAROSPaper-2025-Derwent2025Multi}.

\subsection{Statistical Evaluation}

Bayesian analysis techniques were used to inform a probabilistic model
of the performance of each planner with respect to each metric in
each environment \citep{Stats-Kruschke2013BayesianEstimationSupersedes}.
Raw data distributions for tests that encountered a large number of
CLM often showed two distinct peaks, one for when CLM occur and one
for when they do not. Therefore, we model each metric as a bimodal
$t$-distribution \citep{Stats-Student1908Tdists}, written as
\begin{equation}
m\sim\weighting_{A}\tdistsub AA+\weighting_{B}\tdistsub BB,\label{eq:Bayesian-model}
\end{equation}
where $\mean_{A}$ and $\mean_{B}$ denote the means of components
A and B respectively. Likewise, $\scale_{A}$ and $\text{\ensuremath{\scale}}_{B}$
denote the components' respective scale parameters, whereas $\normality_{A}$
and $\normality_{B}$ denote their respective normality parameters
(also called degrees of freedom). The two components are weighted
by the terms $\weighting_{A}\in\left[0,1\right]$ and $\weighting_{B}\triangleq1-\weighting_{A}$.
If the median posterior estimate for either weighting is above 0.9
in a given model, then that model is considered \emph{effectively}
unimodal and the less-weighted component is excluded from the results.
This is because, in such cases, the less-weighted component has little
impact on the posterior probabilities, and thus its parameters may
fail to converge to estimates informed by the data. Therefore, analysing
such parameter estimates is of little value.

The priors used for each model parameter are given in Table~\ref{tab:priors},
and describe broad distributions based on the sample mean $\samplemean_{\metric}$
and sample standard deviation $\samplestdev_{\metric}$ of the dataset,
reflecting little pre-existing knowledge about how the algorithms
would compare.

\begin{table}
\centering{}\caption{Prior distributions used for the parameters of model \eqref{eq:Bayesian-model}.}
\label{tab:priors} %
\begin{tabular}{ll}
\toprule 
Parameter & Prior Distribution\tabularnewline
\midrule 
Means $\mu_{A}$, $\mu_{B}$ & $\text{Normal\,}(\mu=M_{m},\sigma=100S_{m})$\tabularnewline
Scales $\scale_{A}$, $\scale_{B}$ & $\text{Uniform\,}(L=\frac{S_{m}}{1000},H=1000S_{m})$\tabularnewline
Normalities $\nu_{A}$, $\normality_{B}$ & $\text{Exponential\,}(\lambda=30)$\tabularnewline
Weighting $\weighting_{A}$ & $\text{Uniform\,}(L=0,H=1)$\tabularnewline
\bottomrule
\end{tabular}
\end{table}

For each combination of metric, horizon depth, and scenario, we present
the probability distributions describing the most credible range of
mean performance differences between CVBFP and RHCP (planning times
in Figure~\ref{fig:Planning-Times} and number of stance changes
in Figure~\ref{fig:Stance Changes}). These differences are given
in the form $\mean_{\text{RHCP}}-\mean_{\text{CVBFP}}$, meaning that
a negative result indicates that RHCP exhibits a reduction in this
metric with respect to CVBFP, and vice versa. 

The most credible range of estimates for the mean difference, given
the data, is described by the 95\% Highest Density Interval (HDI)
of each posterior distribution (indicated by the \emph{black} horizontal
bars on each distribution in Figures~\ref{fig:Planning-Times} and
\ref{fig:Stance Changes}). This is the range that contains 95\% of
the probability density, such that all estimates within the range
are more credible than all estimates outside the range. Thus, there
is a 95\% chance that the \emph{true} mean difference lies within
this range. The median estimate is also shown in Figures~\ref{fig:Planning-Times}
and \ref{fig:Stance Changes} by the \emph{green} horizontal bar for
each distribution.

The red dashed lines in Figures~\ref{fig:Planning-Times} and \ref{fig:Stance Changes}
denote the Region of Practical Equivalence (ROPE), defined as the
range of differences too small to be considered practically meaningful
for our purposes. For the planning time comparisons, we chose a ROPE
equal to $\pm10\%$ of the median planning time for CVBFP in that
scenario. For the number of stance changes, we chose $\pm5\%$ of
the number of stance changes that CVBFP would be expected to make
if it took the smallest allowable step in our implementation (10~cm).
For example, in a one-metre-long scenario a six-legged robot making
10~cm steps would need to execute 60 steps, or 120 stance changes.
Hence, a ROPE of $\pm6$ stance changes is used in such a case.

If the 95\% HDI of a given distribution overlaps the ROPE, then there
is a credible possibility that the true difference is not practically
meaningful. Thus, we cannot be confident that a practical difference
exists in these cases. Alternatively, if the 95\% HDI does not overlap
with the ROPE, then we can be confident that a practically meaningful
difference exists.

A two-hour maximum time limit was imposed on each test. A total of
34 out of 100 motion planning attempts by CVBFP in the Wall Walking
scenario exceeded the time limit and were terminated. Therefore, the
data from these tests are excluded from the analysis in this section.
Additionally, one out 100 RHCP motion planning attempt (when $\kmax=1$,
also in the Wall Walking environment) failed to return a motion plan
due to the planner's lack of completeness (mentioned in Section~\ref{subsec:High-Level-Summary}).
All other planning attempts concluded successfully and without issue.

\subsubsection{Planning time}

\begin{figure*}[!t]
\centering{}\includegraphics[width=1\textwidth]{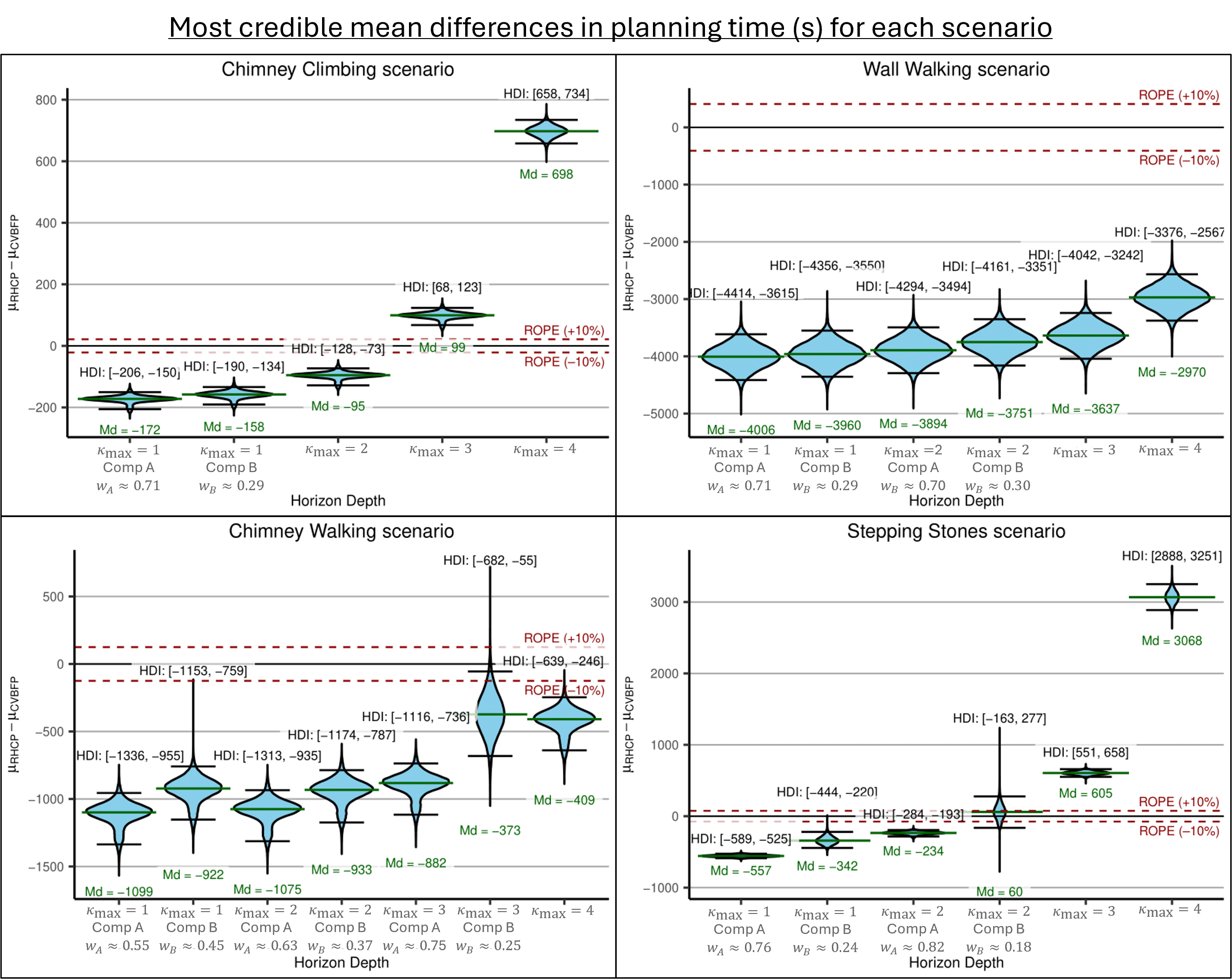}
\caption{Probability distributions of the difference in planning times (s)
between RHCP and CVBFP in each scenario. Each violin denotes a different
vale of $\protect\kmax$. Where the distribution for a given horizon
depth is bimodal, the two components (A and B) are shown separately.
The 95\% HDI is marked on each violin by \emph{black horizontal bars},
while the median is shown in \emph{green}. The ROPE for each environment
is shown as \emph{red dashed horizontal lines}. Violins below the
ROPE's lowerbound favour RHCP in a statistically significant way.
Conversely, violins above the ROPE's upperbound favour CVBFP. When
violins overlap with the ROPE, no strong conclusions can be drawn.\label{fig:Planning-Times}}
\end{figure*}

As expected, Figure~\ref{fig:Planning-Times} shows that longer horizon
depths result in longer planning times. Indeed, the distributions
display an approximately exponential curve with respect to horizon
depth in most cases, in line with expectations given by expression
\eqref{eq:number-of-calls}.

Figure~\ref{fig:Planning-Times} also shows that, when $\kmax=1$,
RHCP executes meaningfully faster than CVBFP across all scenarios,
with the median time saving ranging from 2.6 minutes in the Chimney
Climbing scenario (component B) to 66.8 minutes in the Wall Walking
scenario (component A). This is also true in most cases when $\kmax=2$
(as was our finding in \citep{TAROSPaper-2025-Derwent2025Multi}),
with the exception of component B in the Stepping Stones scenario,
whose 95\% HDI spans the ROPE. Thus, in this case it is not possible
to be confident in the true mean difference, with an increase, a decrease
or no practical difference with respect to CVBFP all being credible
possibilities. We note, however, that the weighting of component B
in this case is approximately $0.18$, meaning that an individual
motion plan is most likely to belong to component A, whose mean planning
time is meaningfully faster than CVBFP's (3.9 minutes faster).

For longer horizons ($\kmax>2$), the planning time comparisons are
more mixed and depend on the scenario. In the Chimney Climbing and
Stepping Stones scenarios, RHCP with $\kmax=3$ has a meaningfully
slower mean execution time than CVBFP (1.7 minutes and 10 minutes
slower, respectively), as does RHCP with $\kmax=4$ (12 minutes and
51 minutes slower, respectively). However, in the Chimney Walking
and Wall Walking scenarios, RHCP with $\kmax=4$ remains faster than
CVBFP on average (6.8 minutes and 50 minutes faster, respectively).
When $\kmax=3$, RHCP is approximately 60 minutes faster than CVBFP
in the Wall Walking scenario, while the comparison in the Chimney
Climbing scenario depends on the component of the bimodal distribution.
Component A (with a weighting of 0.75) is approximately 15 minutes
faster than CVBFP, while the 95\% HDI of component B partially intersects
the ROPE, meaning it remains a credible possibility that no practical
difference is present in this case.
\begin{figure*}[!t]
\centering{}\includegraphics[width=1\textwidth]{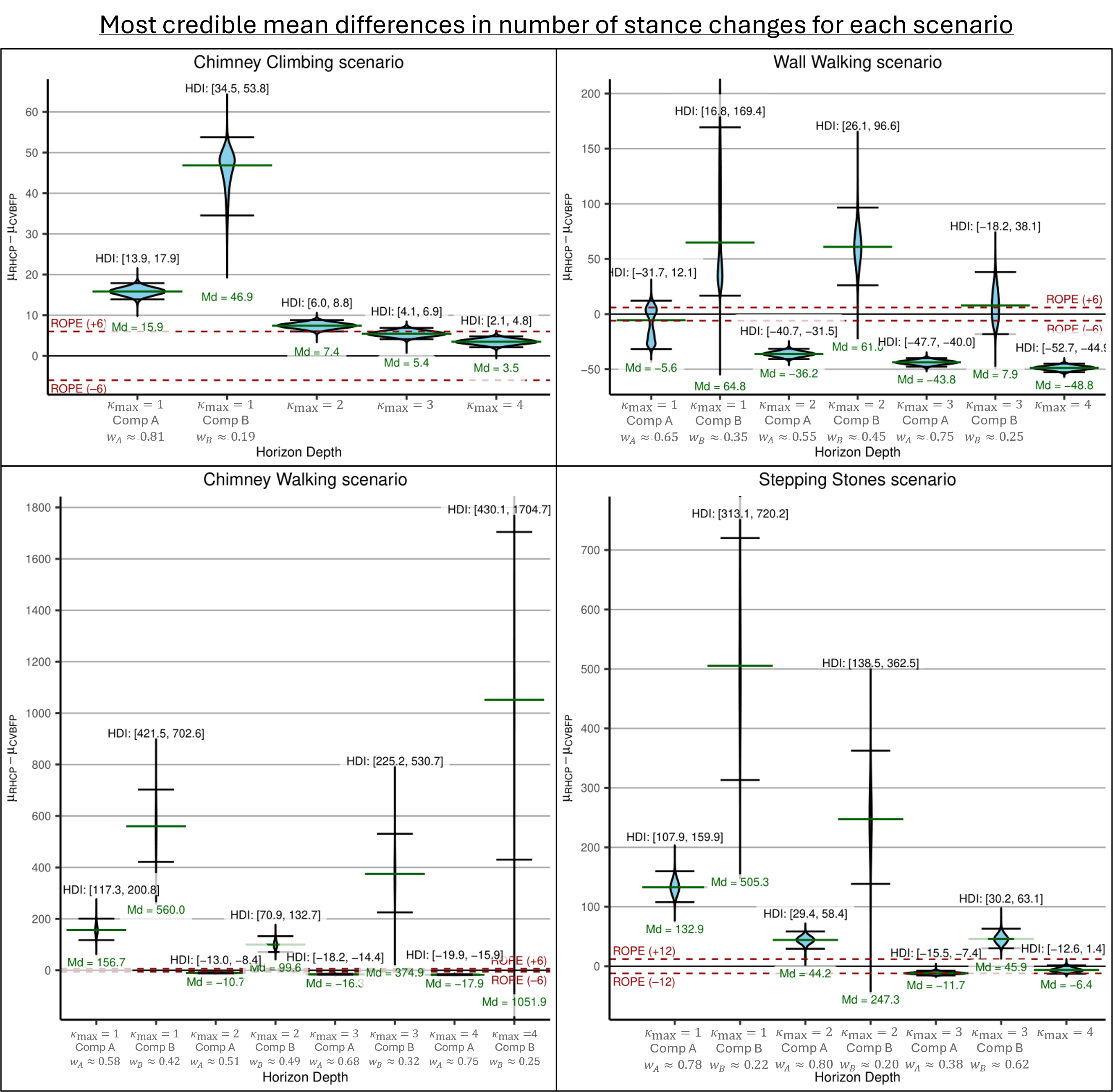}
\caption{Probability distributions of the difference in stance changes between
RHCP and CVBFP in each scenario, plotted according to the same conventions
as Figure~\ref{fig:Planning-Times}\label{fig:Stance Changes}. The
ROPE for each environment is shown as \emph{red dashed horizontal
lines}. Violins below the ROPE's lowerbound favour RHCP in a statistically
significant way. Conversely, violins above the ROPE's upperbound favour
CVBFP. When violins overlap with the ROPE, no strong conclusions can
be drawn. When the HDI is completely inside the ROPE, both planners
are considered to be statistically equivalent.}
\end{figure*}

The magnitudes of the planning time differences are very different
between scenarios. For example, the greatest mean time saving in the
Chimney Climbing scenario was only 2.9 minutes ($\kmax=1$, component
A) compared to over an hour in the Wall Walking scenario ($\kmax=1$,
component A). One explanation for this variation is the relative prevalence
of CLM in different scenarios. For example, CVBFP encountered at least
one CLM in 97\% of tests in the Wall Walking scenario, which may have
contributed to significantly increasing its mean planning time, resulting
in a large disparity between it and RHCP. By contrast, CVBFP encountered
CLM in only 4\% of tests in the Stepping Stones scenario, resulting
in much closer comparisons between the two planners when $\kmax\leq3$.

Additionally, as the horizon depth increases, the planning time distributions
typically become more consolidated into a single component. For example,
in the Chimney Walking scenario, the weighting of component A is $0.55$
when $\kmax=1$, increasing to $0.63$ when $\kmax=2$ and $0.75$
when $\kmax=3$, before becoming unimodal when $\kmax=4$. This is
further demonstrated by the fact that the data for $\kmax=1$ is bimodal
in all scenarios, while that for $\kmax=4$ is unimodal in all scenarios.
This is likely because CLM occur less frequently for longer horizon
depths.

\subsubsection{Number of stance changes}

Figure~\ref{fig:Stance Changes} shows that the number of stance
changes also follows the expected pattern, with longer horizon depths
generally corresponding to more efficient motion plans featuring fewer
stance changes. However, this trend is somewhat occluded by the prevalence
of bimodal distributions whose two components often have very different
means. The relative prevalence of bimodality, as well as the large
disparities between the two components, is likely due to CLM having
a much greater impact on the number of stance changes made by RHCP
than on the algorithm's planning time. This is to be expected, as
the robot is often forced to retreat in the face of CLM, with the
stance changes made during this retreat being counted towards this
data. By contrast, the additional time needed to plan a retreat is
often small, since the previous node can be restored from the global
cache $\cache_{G}$ rather than being generated anew.

Figure~\ref{fig:Stance Changes} also shows that, in the Wall Walking
scenario, RHCP with $\kmax=4$ generates motion plans requiring meaningfully
fewer stance changes than CVBFP on average (48.8 fewer, 47\%), while
in the Chimney Climbing scenario the 95\% HDI for this planning horizon
is entirely contained within the ROPE, suggesting that the difference
in path efficiency is not practically meaningful. In the Stepping
Stones scenario, the distribution for RHCP with $\kmax=4$ is only
partially contained by the ROPE, suggesting that the true difference
may be a very modest decrease (6.4 stance changes) with respect to
CVBFP, though the most likely case is that the true difference is
not meaningful. 

The Chimney walking scenario does not follow this trend, however.
Here, the data for RHCP with $\kmax=4$ is bimodal, with only component
A (weighted approximately 0.75) reflecting a meaningful improvement
over CVBFP (17.9, or 23\%, fewer stance changes on average). 

Component B, on the other hand, reflects a very large increase (1050
additional changes on average, or 1344\%). Indeed, the data for the
Chimney Walking scenario shows a peculiar trend where the secondary
components of each distribution exhibit larger numbers of stance changes
as the horizon depth increases — the opposite of the trend seen elsewhere
in this data. This is somewhat offset by the approximate value of
$\weighting_{B}$ decreasing as the horizon depth increases. Taken
together, these trends suggest that planning attempts in the Chimney
Walking scenario with longer horizon depths are less likely to encounter
a CLM, but the consequences of doing so become more dire.

\begin{figure}
\centering{}\includegraphics[width=1\columnwidth]{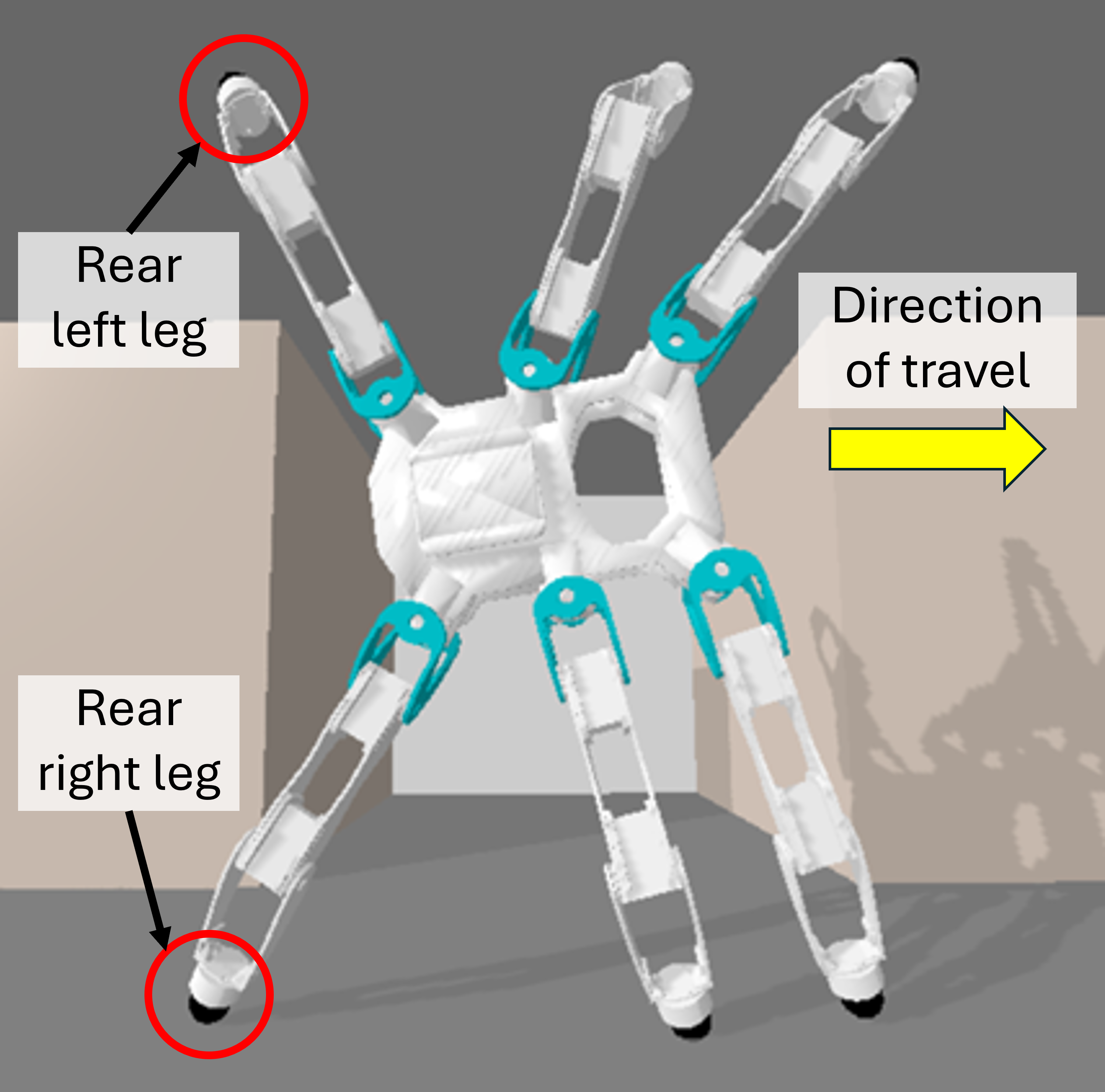}
\caption{Image of the Corin hexapod in a configuration resulting in especially
pernicious CLM behaviour.}
\label{fig:Problem_Config}
\end{figure}

This is believed to be caused by an unintended emergent behaviour
of the CLM intervention mechanism, discussed in Section~\prettyref{subsec:Improving-Resilience}.
The motion plans requiring the largest numbers of stance changes often
involved the robot becoming trapped in a configuration resembling
that shown in Figure~\ref{fig:Problem_Config}. From this configuration,
if the robot lifts either of its rear feet, then it will become unable
to lift the other without loss of balance, causing a CLM.

Ordinarily, the intervention mechanism would resolve the CLM by forcing
the robot to retreat and pre-emptively lift the foot that is becoming
trapped. However, in this case the heuristic worsens the problem because
lifting the trapped foot causes \emph{another} CLM in which the other
rear foot is trapped. This then triggers the intervention mechanism
again, causing \emph{another} CLM, and so on. This loop continues,
making a very large number of unnecessary stance changes, until the
robot has occupied every node on the horizon from the perspective
of that shown in Figure~\ref{fig:Problem_Config}. At this point,
the planner backs up to the parent node of that shown in Figure~\ref{fig:Problem_Config}
and thence escapes from the trapping loop. RHCP variants with longer
horizons are more severely affected by this problem because the number
of nodes on the horizon from the perspective of Figure~\ref{fig:Problem_Config}
grows exponentially with the horizon depth. 

In summary, the general trend observed in Figures~\ref{fig:Planning-Times}
and \ref{fig:Stance Changes} is that short horizon depths (\emph{i.e.,}
$\kmax\leq2$) typically result in faster performance than CVBFP at
the cost of less efficient planning outputs. Likewise, longer horizon
depths (\emph{i.e., }$\kmax\geq3$) typically correspond to paths
of better or equivalent quality to those of CVBFP, while requiring
more time to generate. In most scenarios, there is no value of $\kmax$
for which RHCP exceeds CVBFP in \emph{both} metrics, with the exception
of the Wall Walking scenario, in which this is true when $\kmax=4$.

\section{Conclusions\label{sec:Conclusions}}

This paper presented a novel multi-contact motion planning framework
for legged robots, consisting of a receding-horizon tree search methodology
and a posture generator utilising vector-field inequalities. The proposed
motion planner (RHCP) successfully generated several motions for the
hexapod robot Corin \citep{Corin-WebsiteEntry-UOM2020Corin} in four
challenging environments belonging to different classes: the Stepping
Stones scenario requires the robot to walk over a `river' using
sparse and irregularly placed surfaces; the Wall Walking scenario
features a narrow corridor that requires contacts to be formed on
a vertical wall; the Chimney Walking scenario requires a large gap
to be crossed using the vertical surfaces exclusively; and the Chimney
Climbing scenario sees the robot vertically climb up a chimney. Our
approach adds new capabilities not found in the state-of-the-art,
including rapid re-planing capabilities, reduced susceptibility to
contact-critical local minima, and simultaneously planning whole-body
linking trajectories alongside the stance sequence. 

A thorough experimental comparison of RHCP and CVBFP in virtual environments
found that, with short horizon depths ($\kmax\leq2$), RHCP consistently
generates plans as fast or faster than CVBFP across all scenarios
tested, though these plans are generally of lower quality. Likewise,
longer horizon depths ($\kmax\geq3$) typically result in motion plans
that are as efficient or more efficient than those of CVBFP across
all scenarios, while requiring more time to generate. The exception
to this rule was the Wall Walking scenario, in which RHCP with $\kmax=4$
generated stance sequences that were more efficient than those of
CVBFP while also requiring less time to plan.

Though RHCP sacrifices CVBFP's guarantee of completeness, we find
experimentally that only one of our 1400 motion planning attempts
using RHCP were affected by this.

The undesired emergent behaviours displayed by the heuristic intervention
mechanism in some scenarios (most notably Chimney Walking) represents
a major limitation of our current approach, as does the reliance on
heuristics to mitigate CLM issues more generally. Therefore, a major
focus of future work is to reformulate the posture generator such
that contacts can be placed in such a way that CLM are prevented from
forming. Another limitation of the approach is that it can be difficult
to predict which horizon depth is most applicable for a given situation,
which we plan to address through the development of an adaptive mechanism.
Furthermore, the current formulation relies on a precise model of
the environment using mathematical primitives such as planes. In order
to support real-time planning operations in unknown environments,
the planner will need to be reformulated in order to handle uncertainty
in the environment model, and to work with partially observed scenarios.
Finally, future work also aims to validate the motion plans presented
here by executing them with a physical robot.

\section*{CRediT Authorship Contribution Statement}

\textbf{Daniel S. J. Derwent:} Conceptualization, Methodology, Software,
Validation, Formal analysis, Investigation, Data Curation, Writing
- Original Draft, Visualization.\textbf{ Simon Watson:} Conceptualization,
Resources, Writing - Review \& Editing, Supervision, Project administration.
\textbf{Bruno V. Adorno:} Conceptualization, Methodology, Resources,
Writing - Review \& Editing, Supervision, Project administration,
Funding acquisition.

\section*{Declaration of Competing Interest}

The authors declare that they have no known competing financial interests
or personal relationships that could have appeared to influence the
work reported in this paper. 

\section*{Acknowledgements}

This work was supported by a grant from the University of Manchester
and by the Royal Academy of Engineering under the Research Chairs
and Senior Research Fellowships programme. Sponsoring organisations
did not have any involvement in the study design, the collection,
analysis or interpretation of data, the writing of this article or
decision to submit this article for publication.

Corin was developed by Hassan Hakim Khalili and Wei Cheah \citep{Corin-WebsiteEntry-UOM2020Corin}.
R code used for statistical analysis in Section~\ref{sec:Results}
was based on scripts released by John K. Kruschke, with adaptations
by Ana Christina Almada Campos and the authors.

\section*{Data Availability}

Research data related to this article is available through Figshare
\citep{Dataset-for-this-paper}.

\bibliographystyle{elsarticle-num}
\addcontentsline{toc}{section}{\refname}\bibliography{references}

\end{document}